\pdfoutput=1

\documentclass[11pt]{article}

\usepackage{acl}

\usepackage{times}
\usepackage{latexsym}
\usepackage[T1]{fontenc}
\usepackage{caption}
\usepackage{subcaption}
\usepackage[compact]{titlesec}
\usepackage{appendix}

\usepackage{wrapfig,lipsum,booktabs}
\usepackage{array,multirow,graphicx}
\usepackage{dingbat}
\usepackage{microtype}
\usepackage{linguex}
\usepackage{tipa}
\usepackage[british]{babel}
\usepackage{color,soul}
\usepackage{arydshln}
\usepackage[utf8]{inputenc}
\usepackage{adjustbox}

\usepackage{tikz}
\usepackage{float}
\def\checkmark{\tikz\fill[scale=0.4](0,.35) -- (.25,0) -- (1,.7) -- (.25,.15) -- cycle;} 
\newcommand{\bluecheck}{{\color{blue}\checkmark}}

\newcommand\blfootnote[1]{%
  \begingroup
  \renewcommand\thefootnote{}\footnote{#1}%
  \addtocounter{footnote}{-1}%
  \endgroup
}
\def \ourmodel{SERENGETI}

\title{SERENGETI: Massively Multilingual Language Models for Africa}

\author{\normalsize Ife Adebara$^{1,\star}$ ~ AbdelRahim Elmadany$^{1,\star}$ ~ Muhammad Abdul-Mageed$^{1,2}$ ~ Alcides Alcoba Inciarte$^{1}$ \\
\normalsize $^{1}$Deep Learning \& Natural Language Processing Group,
  The University of British Columbia\\\normalsize  $^{2}$Department of Natural Language Processing \& Department of Machine Learning, MBZUAI\\ %
  \texttt{\normalsize \{ife.adebara@,a.elmadany@,muhammad.mageed@,alcobaaj@mail.\}ubc.ca}}

\begin{document}
\maketitle

\begin{abstract}
Multilingual pretrained language models (mPLMs) acquire valuable, generalizable linguistic information during pretraining and have advanced the state of the art on task-specific finetuning. To date, only $\sim31$ out of $\sim2,000$ African languages are covered in existing language models. We ameliorate this limitation by developing~\ourmodel, a massively multilingual language model that covers $517$ African languages and language varieties. We evaluate our novel models on eight natural language understanding tasks across $20$ datasets, comparing to $4$ mPLMs that cover $4-23$ African languages. \ourmodel~outperforms other models on $11$ datasets across the eights tasks, achieving $82.27$ average F\textsubscript{1}. We also perform analyses of errors from our models, which allows us to investigate the influence of language genealogy and linguistic similarity when the models are applied under zero-shot settings. We will publicly release our models for research.\footnote{\href{https://github.com/UBC-NLP/serengeti}{https://github.com/UBC-NLP/serengeti}}
\blfootnote{ $^{\star}$ Authors contributed equally.}\\


\end{abstract}

	
\section{Introduction}\label{sec:motivation}
\begin{figure}[t]
  \centering
\includegraphics[width=\columnwidth]{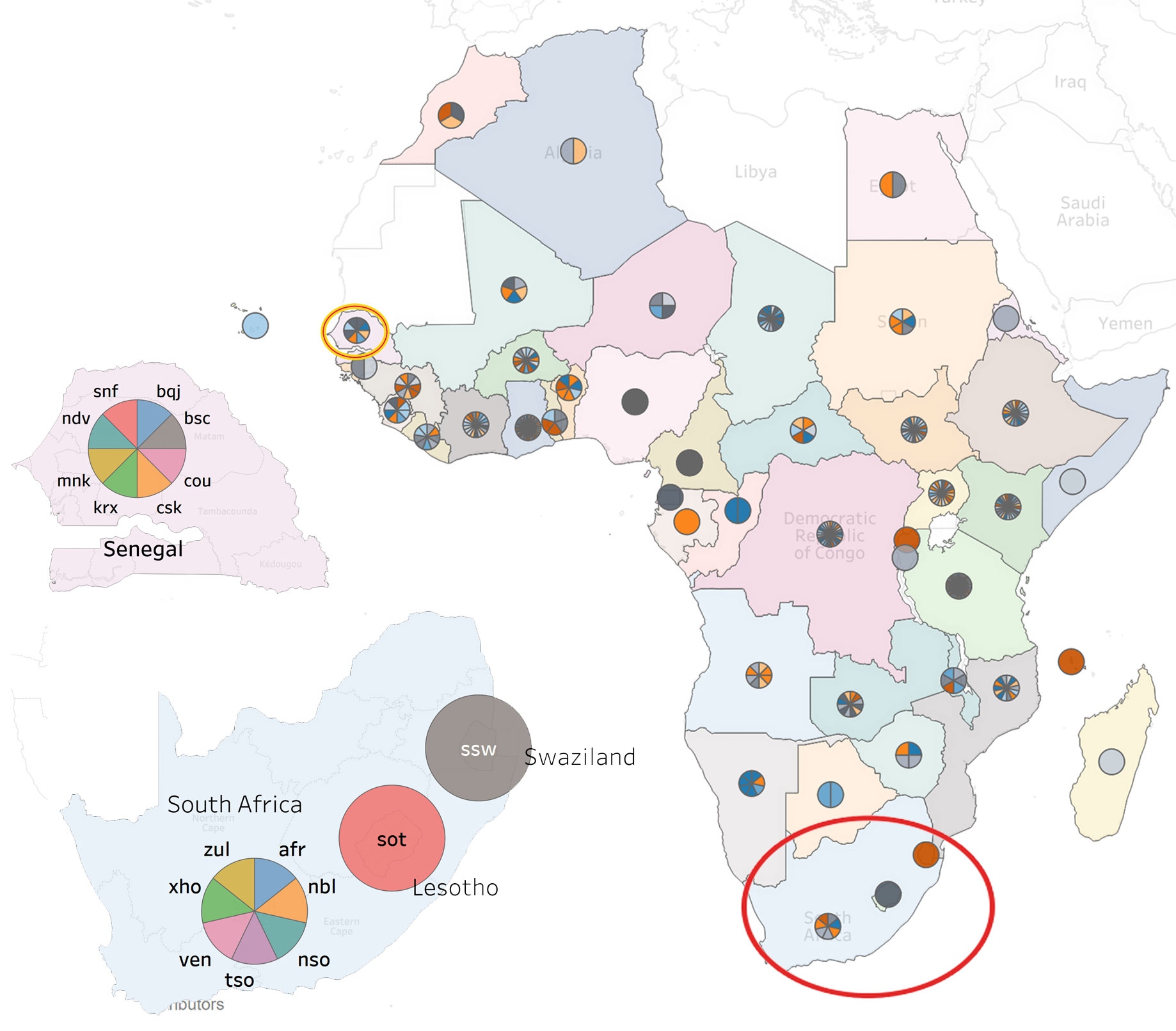}
\caption{\small All $517$ languages in our dataset across the $50$ African countries our data comes from. The language varieties are represented as colored pie shapes within each country. We zero in on South Africa, Lesotho, Swaziland, and Senegal to show detail. We provide a larger map in Appendix~\ref{appdx-fig:countries}.}
\label{fig:countries} 
\end{figure}
Pretraining NLP models with a language modeling objective has gained popularity as a precursor to task-specific finetuning \cite{10.1162/tacl_a_00298}. Pretrained models like BERT \cite{devlin-etal-2019-bert}, ELMo \cite{peters-etal-2018-deep}, Roberta \cite{liu2019roberta}, GPT \cite{radford2018improving, radford2019language, NEURIPS2020_1457c0d6}, and BART \cite{lewis-etal-2020-bart} have advanced the state of the art in a wide variety of tasks, demonstrating how these models acquire valuable, generalizable linguistic information during the pretraining process. However, training language-specific models is possible for only a few languages which have large amounts of data. A popular alternative has been pretrained multilingual language models (mPLM) such as mBERT \cite{devlin-etal-2019-bert} and XML-R \cite{conneau-etal-2020-unsupervised}. mPLMs are trained on large amounts of unlabelled data from multiple languages so that low resource languages may benefit from shared vocabulary and other linguistic information from high-resource and similar languages in the model. The vast majority of the world's $\sim7,000$ languages today remain uncovered by mPLMs, however.

African languages are no exception. Although there are few mPLMs that support a small number of African languages~\cite{devlin-etal-2019-bert, ogueji-etal-2021-small, nzeyimana-niyongabo-rubungo-2022-kinyabert, alabi-etal-2022-adapting, jude-ogundepo-etal-2022-afriteva, conneau-etal-2020-unsupervised}, these cover only a total of $31$ languages. This is grossly inadequate considering that Africa is believed to be home to $\sim2,000$ languages \cite{ethnologue}. Each of these languages encapsulates unique features that are essential in preserving linguistic diversity. The same way every species embodies essential value to the natural ecosystem, each language plays a crucial role in the linguistic ecosystem. That is, each language encodes knowledge about people, their traditions, wisdom, and environment, as well as how it is that they interact with the sum of the concepts in their own culture \cite{adebara-abdul-mageed-2022-towards}. This in turn allows people and communities to preserve and transmit their knowledge, values, unique modes of thinking, meaning and expression, history, culture, traditions, and memory to next generations, while participating in society and constructing their future~\cite{unescodoc}. 

Language technology plays an important role in building inclusive knowledge societies, providing access to education and information, supporting freedom of expression, cultural and linguistic diversity, and further stimulating innovation. This technology thus has great impact on multiple domains, including education, government, health, recreation, among others. This motivates  adequate representation of African languages in the ongoing technological revolution. This is also likely to connect Africa to the rest of the world. Building technologies for African languages may also aid languages that may be at risk of falling into a state of disuse at an alarming rate, thus hopefully preventing subsequent language death that may become inevitable \cite{adebara-abdul-mageed-2022-towards}. 

Developing LMs that represent a large number of African languages is therefore very crucial for achieving progress in Afrocentric NLP \cite{adebara-abdul-mageed-2022-towards} and indeed in addressing issues related to representation bias in artificial intelligence and linguistic diversity - two research themes of international relevance \cite{benderetal2021}. Motivated by this call for Afrocentric NLP, we introduce\textbf{~\ourmodel}. \ourmodel~is a massively multilingual language model exploiting a large manually-curated dataset for $517$ African languages and language varieties. These languages belong to $14$ \textit{language families} and are written in $5$ different \textit{scripts}. In addition to these African languages, \ourmodel~is also pretrained on the top $10$ most spoken languages globally. 

We also introduce \textbf{AfroNLU}, an extensive benchmark exploiting \textit{$20$ different datasets} across \textit{$28$ different languages and language varieties} for various NLP tasks. For even richer evaluation, we also apply our models to an African language identification task covering all the $517$ languages in our pretraining. To the best of our knowledge, AfroNLU is the most extensive and \textit{inclusive} evaluation benchmark proposed to date for African NLP. 

Our contributions in this work are as follows: \textbf{(1)} we collect a large dataset of $517$ African languages and language varieties and exploit it to develop \ourmodel. \textbf{(2)} we propose AfroNLU, a new extensive benchmark for African NLU that has the widest and most inclusive coverage for African NLP today. \textbf{(3)} we benchmark \ourmodel~on AfroNLU and show through meaningful comparisons how our model excels and acquire new SOTA. \textbf{(4)} we offer a linguistically motivated analysis of model performance substantiated in language genealogy, allowing us for the first time to derive insights across the widest range of African languages in the African NLP literature to date.

The rest of the paper is organized as follows: In Section~\ref{sec:litreview} we discuss related work. We describe genealogical information in Section~\ref{sec:genetic}. Next, we give a detailed description of \ourmodel~in Section~\ref{sec:models}. In Section \ref{sec:benchmark} we describe AfroNLU, the benchmark we create. We present performance of~\ourmodel~in Section~\ref{sec:eval} and compare it to other mPLMs. We conclude in Section~\ref{sec:conc}, and outline a number of limitations and use cases for our work in Section~\ref{sec:limits} and Section \ref{sec:ethics}.

\section{Related Work}\label{sec:litreview}
\noindent\textbf{\textit{Afrocentric NLP.}} An \textit{Afrocentric} approach to technology development is crucial for African languages. An afrocentric approach will mean that what technologies to build and how to build, evaluate, and deploy them arises from the needs of local African communities \cite{adebara-abdul-mageed-2022-towards}. We provide more details in Section \ref{app:literature_review} in the Appendix.
\begin{table*}[!h]
\centering
\resizebox{\textwidth}{!}{%
\begin{tabular}{ll}
\toprule
\textbf{Language Model} &\textbf{African languages represented} \\   \midrule
MBERT & Afrikaans, Malagasy, Swahili, Yoruba \\
XLM-R & Afrikaans, Amharic, Hausa, Oromo, Somali, Swahili, Xhosa. \\ \midrule
KinyarBERT & Kinyarwanda \\
AfriBERTA & Afaan Oromoo, Amharic, Gahuza, Hausa, Igbo, Nigerian Pidgin, Somali, Swahili, Tigrinya and Yoruba \\
Afro-XLMR &  Afrikaans, Amharic, Hausa, Igbo, Malagasy, Chichewa, Oromo, Nigerian Pidgin, Kinyarwanda, Kirundi, \\
& Shona, Somali, Sesotho, Swahili, isiXhosa, Yoruba, and isiZulu \\
AfroLM & Amharic, Afaan Oromoo, Bambara, Ghomala, Ewe, Fon, Hausa, Igbo, Kinyarwanda, Lingala, Luganada, \\ 
& Luo, Moore, Chewa, Nigerian Pidgin, Shona, Swahili, Setswana, Akan Twi, Wolof, Xhosa, Yoruba, IsiZulu \\
\midrule
\textbf{\ourmodel} & Includes 517 African languages.\\\bottomrule
\end{tabular}
}
\caption{Encoder-only models with African languages represented.}
\label{tab:resources}

\end{table*} 
\noindent\textbf{\textit{African Language Models.}}
Here, we briefly describe language models covering any number of African languages. Since we develop encoder-only models in this work, we will restrict our discussion to this category of models. We provide information about the African languages covered by these models in Table~\ref{tab:resources}.

\noindent{\textbf{\textit{AfriBERTa}}}~\cite{ogueji-etal-2021-small} is trained using a Transformer with the standard masked language modelling objective and covers $11$ African languages. The pretraining corpus for this model is small (only $108.8$ million tokens), when compared to many other models. 
\noindent{\textbf{\textit{AfroLM}}}~\cite{afroLM} supports $23$ African languages, the largest number of African languages before~\ourmodel. It is trained on a multi-domain dataset from various sources \cite{adelani-etal-2022-thousand, alabi_2022_adaptive, jude-ogundepo-etal-2022-afriteva, niyongabo-etal-2020-kinnews}. It uses a self-active learning framework and achieves SOTA on NER, sentiment analysis, and text classification. 
\noindent{\textbf{\textit{Afro-XLM-R}}}~\cite{alabi-etal-2022-adapting} uses language adaptation on the $17$ most-resourced African languages and three other high-resource foreign languages widely used in Africa (i.e., English, French, and Arabic) simultaneously to provide a single model for cross-lingual transfer learning. Authors show that Afro-XLM-R has competitive results with AfriBERTa and XLM-R on NER, topic classification, and news classification. 
\noindent{\textbf{\textit{KINYaBERT}}}~\cite{nzeyimana-niyongabo-rubungo-2022-kinyabert} uses a two-tier BERT architecture that involves a morphological analyzer and explicitly represents morphological information for Kinyawanda--a morphologically rich African language. Authors show that KINYaBERT achieves good convergence and accuracy, and is robust on multiple downstream tasks.
\noindent{\textbf{\textit{mBERT}}} \cite{devlin-etal-2019-bert} is a multilingual variant of BERT trained on $104$ languages including four African languages. \noindent{\textbf{\textit{XLM-R}}} \cite{conneau-etal-2020-unsupervised} uses a Transformer based architecture and obtains SOTA on cross-lingual classification, sequence labeling, and question answering on $100$ languages including eight African languages. 

\section{Genealogy of African Languages}\label{sec:genetic}
Genealogical or genetic classification groups languages based on their historical and evolutionary relationships. Genetically related languages are often classified into similar families in a hierarchical tree like structure that shows the level of similarity between the languages. Languages with a higher degree of similarity belong to the same class while languages with a lower degree of similarity are further subdivided into different classes and subclasses. Two closely related languages can therefore be viewed as sisters of the same parent language/ancestor--they are languages that evolved over time and/or space from an older parent language~\cite{ludwig_2020}. 
Typological classification differs from geneological classification in that the former is based on grammatical features or types \cite{rainer_2020}. For instance, a typological classification would group tone languages together, or split languages based on their morphological structure into, for instance, isolating or agglutinating languages. Despite this difference, languages that belong to the same family often share similar typological information \cite{ludwig_2020}. For example, most Benue-Congo languages are tone languages \cite{WILLIAMSON2006734}. 
\vspace{-1mm}
In the case of African languages, where typological information is scarcely available \cite{adebara-abdul-mageed-2022-towards, hock2011world}, utilizing genetic classes may be a useful way to determine typological information. If the typological information of one language in a group is known, we may make a sensible assumption that other languages in that group perhaps share similar features with minor variations. We use geneological classification information in evaluating \ourmodel's behaviour. Specifically, we investigate the relationship between language similarity and model performance in zero-shot scenarios for South African languages in some datasets in our benchmark. We use classification information from Ethnologue \cite{ethnologue} in all our analyses. We provide a broad overview of the families in our models under six broad ancestors in Section \ref{app:typology} in the Appendix. 

\section{\ourmodel}\label{sec:models}

\subsection{Pretraining Data}
\begin{table*}[h!]
\centering
\resizebox{\textwidth}{!}{%
\begin{tabular}{lccccccc}
\toprule
\multicolumn{1}{c}{}                                 & \multicolumn{2}{c}{\textbf{Vocabulary}}                                    & \multicolumn{1}{c}{} & \multicolumn{1}{c}{}    & \multicolumn{3}{c}{\textbf{Training Data}}                                                                          \\\cline{2-3}   \cline{6-8} 
\multicolumn{1}{l}{\multirow{-2}{*}{\textbf{Model}}} & \multicolumn{1}{c}{\textbf{Tok}} & \multicolumn{1}{c}{\textbf{Vocab Size}} & \multicolumn{1}{c}{\multirow{-2}{*}{\textbf{\#Params}}}      & \multicolumn{1}{c}{\multirow{-2}{*}{\textbf{\#Lang.} (afr/all)}} & \multicolumn{1}{c}{\textbf{Tokens} (afr/all)} & \multicolumn{1}{c}{\textbf{Size} (afr/all)} & \multicolumn{1}{c}{\textbf{Source}} \\
\midrule
xlmr                                            & SP                               & 250k                                    & 270M                                      & 8 / 100                                       & UNK/164B                                & UNK/2.4 GB                            & CC-100                                            \\
mbert                                           & WP                               & 110K                                    & 110M                                      & 4 / 100                                       & UNK/12.8B                               & UNK/100GB                             & Books, Wiki.                                      \\ \midrule
Afro-XLMR                                       & SP                               & 70.6K                                   &270M                                      & 17 / 20                                       & ---                                 & 21.6 GB                           & mC4, CC, BBC, VOA                                 \\
AfriBERTa                                       & WP                               & 70k                                     & 111M                                      & 11 / 11               & 108.8M                              & 0.94 GB                           & BBC, CC                   \\
AfroLM & SP & 250K & 264M & 23/23 & ---& 0.73GB& mC4, CC, BBC, VOA \\\midrule
\ourmodel-E110                                       & WP                               & 110K                                    & 170M                                        & 517 / 527                                     &  7.1B/8.6B                                &40/42GB                              & RT, News, GD, HD, EC                                        \\
\ourmodel-E250                                       & WP                               & 250K                                    & 277M                                        & 517/527                                      & 7.1B/8.6B                                &40/42GB                              & RT, News, GD, HD, EC                                        \\
\ourmodel                                            & SP                               & 250K                                    & 278M                                         & 517/527                                     & 7.1B/8.6B                                &40/42GB		                               & RT, News, GD, HD, EC   \\                                    

\bottomrule
\end{tabular}%
}
\caption{Models with African languages that we compare \ourmodel~with. SP: SentencePiece, WP: WordPiece. Data sources include - CC: CommonCrawl, EC: Existing corpora, GD: Government documents, HD: Health documents, RT: Religious text, UNK: Unknown.}
\label{tab:encoder-only}
\end{table*}

\ourmodel~is pretrained using $42$GB of data comprising a multi-domain, multi-script collection of texts that we manually curate. The pretraining data covers $517$ African languages and the $10$ most spoken languages globally (i.e., Arabic, English, French, German, Greek, Italian, Portuguese, Russian, Spanish, and Turkish). The multi-domain dataset comprises texts from religious, news, government documents, health documents, and existing corpora written in five scripts from the set \textit{\{Arabic, Coptic, Ethiopic, Latin, and Vai\}}. For the top ten foreign languages, we randomly select $1$M paragraphs from Wikipedia for each language to use in our overall pretraining data. We provide further details of the pretraining data in Section \ref{app:pretraing-data} in the Appendix. We also show all languages in our pretraining data in Tables \ref{tab:lang_listI}, \ref{tab:lang_listII}, and \ref{tab:lang_listIII}.

\subsection{Preprocessing}
To prepare the raw data for pretraining, we perform light preprocessing to retain a faithful representation of the naturally occurring text. Specifically, we ensure that images and non-text materials are not in our dataset by using regular expression and manual curation techniques. We do not perform any further preprocessing of the data before splitting the text off into tokens. For tokenization, we use a WordPiece tokenizer \cite{song-etal-2021-fast}. We experiment with two vocabulary sizes, $110$K and $250$K. 


\subsection{\ourmodel~Models}
We pretrain both Electra style~\cite{Clark2020ELECTRA, chi2021xlme} as well as XLM-R style \cite{conneau-etal-2020-unsupervised} models, as follows.

\noindent\textbf{\ourmodel-E110 and~\ourmodel-E250.}
We first pretrain Electra \cite{chi2021xlme} style models. Electra uses a multilingual replaced token detection (MRTD) objective for training. Unlike other training objectives, the goal of MRTD is to distinguish real input tokens from corrupted tokens. Models built with this objective are pretrained as discriminators rather than generators. We train the models with two vocabulary sizes, $110$K and $250$K, and hence refer to them as~\ourmodel-E110 and~\ourmodel-E250. Each of these models has 12 layers and $12$ attention heads. We pretrain each model for $40$ epochs with a sequence length of $512$, a learning rate of $2e-4$ and a batch size of $216$ and $104$ for the~\ourmodel-E110 and~\ourmodel-E250, respectively. We pre-train the models on $1$ Google Cloud TPU with $8$ cores (v$3.8$) from TensorFlow Research Cloud (TFRC). \footnote{\href{https://www.tensorflow.org/tfrc}{https://www.tensorflow.org/tfrc}.}



\noindent\textbf{\ourmodel~Model.} Apart form the Electra models, we also experiment with an XLM-R base architecture. We train the model with a $250$K vocabulary size for $20$ epochs. This model has $12$ layers and $12$ attention heads, a sequence length of $512$ and a batch size of $8$. We pre-train this model on $80$ M$50$ AMD Pod GPUs with $16$G ram. Our XLM-R model has better performance compared to the Electra models as we will show. We provide information about each model we build and compare with in Table \ref{tab:encoder-only}.



\section{AfroNLU Benchmark}\label{sec:benchmark}
\begin{table*}[]
\centering
\resizebox{\textwidth}{!}{%
\begin{tabular}{lllrrr}
\toprule
\textbf{Cluster}                    &\textbf{Dataset}& \textbf{Languages}&\textbf{TRAIN} & \textbf{DEV} & \textbf{TEST} \\ \midrule

\multirow{8}{*}{NER}                & masakaner-v1\textsuperscript{$\star$} &  amh, hau, ibo, kin, lug, luo, pcm, swh, wol, yor  & 443,692 &  60,515 & 134,126   \\ 
                & masakaner-v2\textsuperscript{$\star$} &  bam, bbj, ewe, fon, hau, ibo, kin, lug, mos, nya, \\
                & & pcm, sna, swa, tsn, twi, wol, xho, yor, zul & 2,537,792 & 362,837 & 726,830   \\ 
                
                & masakaner-east\textsuperscript{$\star$} & amh, kin, lug, luo, swh  & 162,388 & 21,206 & 46,407   \\ 
  & masakaner-eastwest\textsuperscript{$\star$} & amh, hau, ibo, kin, lug, luo, pcm, swh, wol, yor                 & 416,113 &  56,512 &  126,176    \\
  & masakaner-west\textsuperscript{$\star$} & hau, ibo, pcm, wol, yor                  &   253,725 & 35,306 & 79,769   \\
  & nchlt-ner \textsuperscript{$\star$}   & afr, nbl, nso, sot, ssw, tsn, tso, ven, xho, zul                       & 1,749,372 & 219,703 & 215,616    \\
  & yoruba-twi-ner\textsuperscript{$\star$} & yor                     & 20,237 & 2,811 & 5,471    \\
  & wikiann\textsuperscript{$\star$} & afr, amh, ibo, mlg, kin, som, swh, yor                        & 9,244 & 9,240 & 9,424    \\ \midrule
Phrase Chunking                    & phrase-chunk\textsuperscript{$\star$} & afr, nso, sot, ssw, tsn, tso, ven, zul                    &107,492& 12,972 & 13,389   \\ \midrule

POS  & igbo-pos\textsuperscript{$\star$} & ibo                         & 756,775 & 94,692 & 95,048    \\ \midrule
\multirow{4}{*}{News}               & amharic-news\textsuperscript{$\dagger$}& amh                       &41,185 & 5,148 & 5,149   \\
  & kinnews\textsuperscript{$\dagger$}& kir                          & 15,308 & 1,701 & 4,254    \\
  & kirnews\textsuperscript{$\dagger$} & run                           & 3,320 & 369 & 923    \\
  & swahili-news-v0.2\textsuperscript{$\dagger$} & swh                 & 19,986 & 2,221 & 7,338   \\ \midrule
\multirow{3}{*}{Sentiment Analysis} & bambara-v2\textsuperscript{$\dagger$} & bam                       &2,436 & 305 & 305    \\
  & pidgin-tweet\textsuperscript{$\dagger$}  & pcm                    & 11,200 & 1,400 & 1,400    \\
  & yosm\textsuperscript{$\dagger$}   & yor                            & 800 & 200 & 500    \\ \midrule
\multirow{2}{*}{Topic}              & hausa-topic\textsuperscript{$\dagger$}   & hau                    & 2,045 & 290 & 582    \\
  & yoruba-topic\textsuperscript{$\dagger$}   & yor                   &1,340 & 189 & 379    \\
  \midrule
QA                                  & qa-swahili\textsuperscript{$\dagger$}  & swh                       &49,881 & 5,077 & 499   \\ \hline
\multirow{2}{*}{LID}                                 & AfroLID\textsuperscript{$\dagger$}  & 517 African Languages                       &2,496,980 & 25,850 & 51,400  \\ 
&   Afri-Senti  & amh, hau, ibo, pcm, swh, yor & & & -\\
\bottomrule
\end{tabular}%
}
\caption{Distribution of AfroNLU datasets. \textsuperscript{$\star$} indicates that datasize is measured at token level. \textsuperscript{$\dagger$} indicates data size measured at sentence level.}
\label{tab:afronlu_data}
\end{table*}

\begin{table}[h!]
\centering
\resizebox{\columnwidth}{!}{%
\begin{tabular}{lcccc}
\toprule
\textbf{Tasks}                   & \textbf{AfriBERTa} & \textbf{Afro-XLMR} & \textbf{KinyaBERT} & \textbf{\ourmodel} \\\midrule
NER                     & \bluecheck                 & \bluecheck                 & \bluecheck                 & \bluecheck                 \\
PC           & ---                  & ---                  & ---                  & \bluecheck                 \\ 
POS                    & ---                  & ---                  & ---                  & \bluecheck                 \\ 
NC     & \bluecheck                 & \bluecheck                 & ---                  & \bluecheck                 \\ 
SA      & ---                  & \bluecheck                 & ---                  & \bluecheck                 \\ 
TC    & ---                  & \bluecheck                 & \bluecheck                 & \bluecheck                 \\ 
QA      & ---                  & ---                  & ---                  & \bluecheck                 \\ 
LID & ---                  & ---                  & ---                  & \bluecheck                 \\ 
GLUE                    & ---                  & ---                  & \bluecheck                 & ---                  \\ 
\bottomrule
\end{tabular}%
}
\caption{Tasks evaluation comparison across different African language MLMs. NER: named entity recognition, PC: phrase chunking, POS: part of speech, NC: news classification, SA: sentiment analysis, TC: topic classification, QA: question answering, LID: language identification.}\label{tab:model_comparison}
\vspace{-4mm}
\end{table}
Our goal is to evaluate our models extensively, and so we combine all available datasets we could acquire to create an evaluation benchmark that we refer to as \textbf{AfroNLU}. AfroNLU is composed of \textit{seven different tasks}, covering both token and sentence level tasks, across $18$ different datasets. The benchmark covers a total of \textit{$32$ different languages and language varieties}. In addition we evaluate our best model (\ourmodel) on an African language identification (LID) task covering all the $517$ languages in our pretraining collection. For LID, we use two datasets to test \ourmodel. This puts AfroNLU at a total of \textit{20 different datasets} and eight different tasks. To the best of our knowledge, our evaluation benchmark is the most extensive compared to previous published research. We provide detailed statistics of the datasets comprising AfroNLU in Table \ref{tab:afronlu_data}. We also provide a detailed comparison of our AfroNLU benchmark with evaluation data from other models in Table~\ref{tab:model_comparison}. We now describe each of the downstream tasks in AfroNLU.

\subsection{Named Entity Recognition (NER)}
We evaluate our models on NER datasets across multiple languages. We use MasakhaNER data \cite{ifeoluwa2021masakhaner}, WikiAnn \cite{pan2017cross, rahimi-etal-2019-massively}, Yoruba-Twi NER data \cite{alabi2020massive}, Distance Supervision NER (DS NER) Data \cite{hedderich-etal-2020-transfer} and multiple NER data from \href{https://repo.sadilar.org/handle/20.500.12185/7}{SADiLaR}. For our experiments, we use the region aggregates on MasakhaNER. Specifically, we use MasakhaNER-east, MasakhaNER-west, and~MasakhaNER-eastwest. MasakhaNER-east includes NER data for Amharic, Kinyawanda, Luganda, Luo, and Swahili. MasakhaNER-west includes NER data for Hausa, Igbo, Nigerian-Pidgin, Wolof, and Yoruba.~MasakhaNER-eastwest, on the other hand, includes a combination of MasakhaneNER-east and MasakhaneNER-west. Data from SADiLaR cover ten indigenous South African languages and is annotated for person, organisation, location, and miscellaneous named entities. Miscellaneous named entities refer to all rigid designators that do not fall into one of the other categories, including temporal expressions (dates and times), URLs, numerical expressions, publications, names of languages, nationalities, among others. More details about the datasets are in Table~\ref{tab:afronlu_data}. 
\hfill
\subsection{Part of Speech Tagging}
We test our models on POS tagging datasets for Igbo taken from IgboNLP~\cite{10.1145/3146387, 10.1145/3314942}. In Table~\ref{tab:afronlu_data}, we provide the statistical details for the dataset.
\hfill
\subsection{Phrase Chunks}
We evaluate our models on phrase chunks datasets for ten Indigenous languages of South Africa (see Table \ref{tab:afronlu_data}). The data has annotations for noun, verb, adjective, adverbial, and prepositional phrase chunks. Words not belonging to these phrase types are labelled with the tag \textit{O}.

\subsection{Sentiment Analysis}
We finetune our model on three sentiment analysis datasets, including Bambara Sentiment dataset \cite{diallo2021bambara}, YOSM--a new Yorùbá Sentiment Corpus for Movie Reviews~\cite{shode_africanlp}, and the Nigerian Pidgin sentiment dataset \cite{oyewusi2020semantic}, respectively. Some details of these datasets is in Table \ref{tab:afronlu_data}. 

\subsection{News classification}
We use news classification datasets for Amharic \cite{azime2021amharic}, Kinyarwanda \cite{niyongabo-etal-2020-kinnews}, Kirundi \cite{niyongabo-etal-2020-kinnews}, and Swahili \cite{davis_david_2020_4300294, davis_david_2020_5514203}. The Amharic dataset contains six classes--news, sport, politics, international news, business, and entertainment. The Swahili dataset also has six categories including local news, international, finance, health, sports, and entertainment. The datasets for Kinyarwanda and Kirundi have $14$ and $12$ categories each, respectively. 
Again, data statistics are in Table~\ref{tab:afronlu_data}.

\subsection{Topic classification}
We include topic classification datasets for Y{o}r\`{u}b\'{a} and Hausa \cite{hedderich-etal-2020-transfer}.  The Y{o}r\`{u}b\'{a} and Hausa datasets contain news titles collected from VOA Hausa and BBC Y{o}r\`{u}b\'{a} news sites. The Y{o}r\`{u}b\'{a} dataset has seven topics--Nigeria, Africa, world, entertainment, health, sports, and politics, while the Hausa dataset is categorized into five topics - Nigeria, Africa, world, health, and politics. In Table~\ref{tab:afronlu_data}, we provide details about the data split sizes.

\subsection{Question Answering}
We use \href{https://github.com/google-research-datasets/tydiqa}{TYDIA} question answering dataset \cite{clark-etal-2020-tydi}. The dataset has a primary task and a gold passage task. The primary task has two subtasks, one for passage selection and another that is a minimal answer span. For the passage selection subtask, a list of passages is given and the required response is either the index of the passage where the answer to the question is or null (if no answer exists in the passage). The minimal answer span subtask on the other hand gives a full article and the expected answer is either the start and end byte indices of the minimal span that answers the question, yes or no response, or null (if no minimal answer exists). For the gold passage task, a correct answer is predicted from a passage containing one answer. This is similar to existing reading comprehension. We use the Kiswahili dataset alone, since it is the only African language in the dataset. Details about the data splits can be found in Table~\ref{tab:afronlu_data}. 






\subsection{Language Identification}
We also evaluate~\ourmodel~on the task of language identification (LID). 
LID focuses on identifying the human language a piece of text or speech segment belongs to, making automatic LID an important first step in processing human language appropriately~\cite{tjandraetal2021improved, thara_etal_2021}. We use datasets from AfroLID \cite{afrolidpaper} for this task. AfroLID data is a multi-genre, multi-script dataset for $517$ African languages. We compare the performance of AfroLID data on our models with performance on AfroLID tool. To ensure a fair comparison, the data used for AfroLID is completely different from the data used for \ourmodel. We also evaluate our LID model on AfriSenti dataset \cite{muhammad2022naijasenti, yimam-etal-2020-exploring}.
\section{Experimental Setup and Evaluation}\label{sec:eval}
\begin{table*}[h]
\resizebox{\textwidth}{!}{%
\begin{tabular}{llcllllccc}
\toprule
\textbf{Cluster}                             & \textbf{Dataset}                & \textbf{SOTA} &\textbf{XLMR} & \textbf{mBERT}               & \textbf{Afro-XLMR} & \textbf{AfriBERTa} & \textbf{\ourmodel-E110}               & \textbf{\ourmodel-E250}      & \textbf{\ourmodel}              \\
\midrule
%
\multirow{8}{*}{NER}                & masakaner-v1     & 84.80\textsuperscript{$\pm0.3$}$\ddagger\ddagger\ddagger$               & 81.41 \textsuperscript{$\pm$0.26}  & 78.57 \textsuperscript{$\pm$0.53}  & 84.16 \textsuperscript{$\pm$0.45} &81.42 \textsuperscript{$\pm$0.30} & 81.23 \textsuperscript{$\pm$0.32} & 81.54 \textsuperscript{$\pm$0.68} & \textbf{84.53} \textsuperscript{$\pm$0.56} \\
                & masakaner-v2     & 87.00
\textsuperscript{$\pm1.2$}$\ddagger\ddagger\ddagger$              & 87.17 \textsuperscript{$\pm$0.18}  & 84.82\textsuperscript{$\pm$0.96}  & 88.69 \textsuperscript{$\pm$0.12} & 86.22 \textsuperscript{$\pm$0.06} & 86.57 \textsuperscript{$\pm$0.27} & 86.69 \textsuperscript{$\pm$0.29} & \textbf{88.86} \textsuperscript{$\pm$0.25} \\
														
               & masakaner-east     & 80.62\textsuperscript{$\star$}                 & 80.38 \textsuperscript{$\pm$0.56}  & 78.33 \textsuperscript{$\pm$1.25}  & 83.02 \textsuperscript{$\pm$0.31} & 79.31 \textsuperscript{$\pm$0.92} & 80.53 \textsuperscript{$\pm$0.71} & 81.26 \textsuperscript{$\pm$0.68} & \textbf{83.75} \textsuperscript{$\pm$0.26} \\
                                    & masakaner-eastwest & 82.34\textsuperscript{$\star$}                  &82.85 \textsuperscript{$\pm$0.38}  & 82.37 \textsuperscript{$\pm$0.90} & \textbf{86.31} \textsuperscript{$\pm$0.30} & 82.98 \textsuperscript{$\pm$0.44} & 82.90 \textsuperscript{$\pm$0.49} & 83.67 \textsuperscript{$\pm$0.44} & 85.94 \textsuperscript{$\pm$0.27} \\
                                    & masakaner-west     & 83.11\textsuperscript{$\star$}                  & 82.85\textsuperscript{$\pm$0.79}  & 83.99 \textsuperscript{$\pm$0.39} & \textbf{86.78} \textsuperscript{$\pm$0.44} & 84.08 \textsuperscript{$\pm$0.32} & 82.06 \textsuperscript{$\pm$0.67} & 83.45 \textsuperscript{$\pm$0.81} & 86.27 \textsuperscript{$\pm$0.94} \\
                                    & nchlt-ner          & --- & 71.41 \textsuperscript{$\pm$0.07}  & 70.58 \textsuperscript{$\pm$0.26} & 72.27 \textsuperscript{$\pm$0.14} & 68.74 \textsuperscript{$\pm$0.29} & 64.46 \textsuperscript{$\pm$0.37} & 64.42 \textsuperscript{$\pm$0.24} & \textbf{73.18} \textsuperscript{$\pm$0.24} \\
                                    & yoruba-twi-ner     & --- & 61.18 \textsuperscript{$\pm$2.19}  & 70.37 \textsuperscript{$\pm$0.61} & 58.48 \textsuperscript{$\pm$1.85} & 69.24 \textsuperscript{$\pm$3.05} & 61.77 \textsuperscript{$\pm$1.24} & 57.99 \textsuperscript{$\pm$2.61} &\textbf{ 71.25} \textsuperscript{$\pm$1.73} \\
                                    & wikiann            & \multicolumn{1}{l}{}   & 83.82 \textsuperscript{$\pm$0.39}  & 82.65 \textsuperscript{$\pm$0.77} &\textbf{ 86.01} \textsuperscript{$\pm$0.83} & 83.05 \textsuperscript{$\pm$0.20}  & 83.17 \textsuperscript{$\pm$0.54} & 84.85 \textsuperscript{$\pm$0.53} & 85.83 \textsuperscript{$\pm$0.94} \\ \midrule
Phrase Chunking                    & phrase-chunk      & --- & 88.86 \textsuperscript{$\pm$0.18}  & 88.65 \textsuperscript{$\pm$0.06} & 90.12 \textsuperscript{$\pm$0.12} & 87.86 \textsuperscript{$\pm$0.20}  & 90.39 \textsuperscript{$\pm$0.21} & 89.93 \textsuperscript{$\pm$0.33} & \textbf{90.51} \textsuperscript{$\pm$0.04} \\\midrule
POS                                 & igbo-pos           & --- & 85.50 \textsuperscript{$\pm$0.08}   & 85.42 \textsuperscript{$\pm$0.13} & 85.39 \textsuperscript{$\pm$0.21} & 85.43 \textsuperscript{$\pm$0.05} & 85.50 \textsuperscript{$\pm$0.16}  & \textbf{85.61} \textsuperscript{$\pm$0.13} & 85.54 \textsuperscript{$\pm$0.08} \\\midrule
\multirow{4}{*}{News Classification}               & amharic-news       & ---                  & 84.97 \textsuperscript{$\pm$0.55}  & 59.01 \textsuperscript{$\pm$1.47} & 86.18 \textsuperscript{$\pm$0.85} & 86.54 \textsuperscript{$\pm$1.20}  & 86.50 \textsuperscript{$\pm$0.71}  & 86.34 \textsuperscript{$\pm$0.30}  & \textbf{86.82} \textsuperscript{$\pm$0.72} \\
                                    & kinnews            &                  & 76.58 \textsuperscript{$\pm$0.70}   & 77.45 \textsuperscript{$\pm$0.43} & 79.13 \textsuperscript{$\pm$0.53} & 80.40 \textsuperscript{$\pm$1.50}   & \textbf{81.43} \textsuperscript{$\pm$1.02} & 80.38 \textsuperscript{$\pm$1.36} & 79.80 \textsuperscript{$\pm$0.68}  \\
                                    & kirnews            & ---                  & 57.18 \textsuperscript{$\pm$3.44}  & 74.71 \textsuperscript{$\pm$2.56} & 87.67 \textsuperscript{$\pm$0.92} & \textbf{89.59} \textsuperscript{$\pm$0.27} & 78.75 \textsuperscript{$\pm$3.24} & 86.60 \textsuperscript{$\pm$1.28}  & 87.53 \textsuperscript{$\pm$2.31} \\
                                    & swahili-news-v0.2  & --- & 87.50 \textsuperscript{$\pm$0.91}   & 85.12 \textsuperscript{$\pm$0.93} & 87.49 \textsuperscript{$\pm$1.26} & 87.91 \textsuperscript{$\pm$0.36} & 87.33 \textsuperscript{$\pm$0.28} & 86.12 \textsuperscript{$\pm$1.30}  & \textbf{88.24} \textsuperscript{$\pm$0.99} \\\midrule
\multirow{3}{*}{Sentiment Analysis} & bambara-v2         & 64.00\textsuperscript{$\dagger$}                  & 47.17 \textsuperscript{$\pm$1.83} & 64.56 \textsuperscript{$\pm$1.71} & 59.40 \textsuperscript{$\pm$0.56}  & 65.06 \textsuperscript{$\pm$2.08} & 65.07 \textsuperscript{$\pm$2.59} &\textbf{ 65.76} \textsuperscript{$\pm$2.02} & 63.36 \textsuperscript{$\pm$3.31} \\
                                    & pidgin-tweet       & --- & 70.42 \textsuperscript{$\pm$0.68}  & 68.59 \textsuperscript{$\pm$0.47} & \textbf{71.40} \textsuperscript{$\pm$0.51}  & 69.19 \textsuperscript{$\pm$0.97} & 71.06 \textsuperscript{$\pm$0.39} & 70.46 \textsuperscript{$\pm$1.02} & 69.74 \textsuperscript{$\pm$0.92} \\
                                    & yosm               & 87.20\textsuperscript{$\ddagger$}                   & 85.57 \textsuperscript{$\pm$1.09}  & 85.25 \textsuperscript{$\pm$0.25} & 87.46 \textsuperscript{$\pm$0.42} & \textbf{88.66} \textsuperscript{$\pm$0.23} & 86.86 \textsuperscript{$\pm$0.95} & 85.58 \textsuperscript{$\pm$1.51} & 87.86 \textsuperscript{$\pm$0.81} \\\midrule
\multirow{2}{*}{Topic}              & hausa-topic        & 48.52 \textsuperscript{$\dagger\dagger$}                  & 85.80 \textsuperscript{$\pm$1.45}   & 81.38 \textsuperscript{$\pm$0.42} & 88.67 \textsuperscript{$\pm$0.30}  & \textbf{92.59} \textsuperscript{$\pm$0.69} & 88.52 \textsuperscript{$\pm$1.31} & 89.07 \textsuperscript{$\pm$0.95} & 89.93 \textsuperscript{$\pm$0.49} \\
                                    & yoruba-topic       & 54.93 \textsuperscript{$\dagger\dagger$}                  & 54.69 \textsuperscript{$\pm$2.89}  & 71.79 \textsuperscript{$\pm$1.43} & 75.13 \textsuperscript{$\pm$1.40}  & \textbf{81.79} \textsuperscript{$\pm$0.66} & 65.22 \textsuperscript{$\pm$4.72} & 66.34 \textsuperscript{$\pm$4.09} & 79.87 \textsuperscript{$\pm$1.61} \\\midrule
QA                                  & qa-swahili         & 81.90 \textsuperscript{$\ddagger\ddagger$}                 & 82.79 \textsuperscript{$\pm$1.93}  &\textbf{ 83.40} \textsuperscript{$\pm$0.78}  & 79.94 \textsuperscript{$\pm$0.39} & 57.3 \textsuperscript{$\pm$1.8}   & 79.76 \textsuperscript{$\pm$0.52} & 81.25 \textsuperscript{$\pm$1.33} & 80.01 \textsuperscript{$\pm$0.78}\\

\midrule

\multicolumn{3}{c}{ \textbf{AfroNLU Score}  }      & 
76.91	&	77.85	&	81.09	&	80.37	&	79.45	&	79.87	&	\textbf{82.44} \\
\bottomrule
\end{tabular}%
}
\caption{Performance of models on seven AfroNLU benchmark TEST datasets. ($F_1$) score is the evaluation metric. Our model (\ourmodel) significantly outperforms AfriBERTa (the 2\textsuperscript{nd} in row) on 13/18 datasets and achieve SOTA on 9/18 datasets.  \textbf{SOTA} as reported on \textsuperscript{$\star$}\cite{ifeoluwa2021masakhaner}, \textsuperscript{$\dagger$}\cite{diallo2021bambara}, \textsuperscript{$\ddagger$}\cite{shode_africanlp}, \textsuperscript{$\dagger\dagger$}\cite{hedderich-etal-2020-transfer} and \textsuperscript{$\ddagger\ddagger$}\cite{clark-etal-2020-tydi}, \textsuperscript{$\ddagger\ddagger\ddagger$}\cite{masakaner2.0}. We use a dash (-) to represent tasks without a known SOTA.} 
\label{tab:afronlu_test_results}
\end{table*}
We evaluate \ourmodel~on eight task clusters in the benchmark, and report results on our Test set in Table~\ref{tab:afronlu_test_results}. We also report performance on our Dev set in Table~\ref{tab:afronlu_dev_results} (Appendix). For each task cluster, we finetune for a maximum of $25$ epochs with a patience value of five. We compare results from~\ourmodel, \ourmodel-E110, and \ourmodel-E250 to encoder-only models covering any number of African languages. Specifically, we compare with XLMR, mBERT, Afro-XLMR, and AfriBERTa. We report the results of each experiment as an average of three runs, showing the standard deviation. We also evaluate \ourmodel~on language identification and show results on Afrolid in Table \ref{tab:afrolid_results} and on Afrisenti in Table \ref{tab:afrolid_vs_serengeti}. For multilingual datasets in each task, we show evaluation results per language, comparing the performance of various models in Table \ref{tab:langs_test_results} in the Appendix. 


\begin{table}[h!]
\centering

\resizebox{0.6\columnwidth}{!}{%
\begin{tabular}{llclllllll}
\toprule
\textbf{Task}                & \textbf{AfroLID}  & \textbf{\ourmodel}              \\
\midrule

Dev   & 96.14\textsuperscript{$\star$}                 & \textbf{97.64} \textsuperscript{$\pm$0.02}  \\ 
Test   & 95.95\textsuperscript{$\star$}                 & \textbf{97.41} \textsuperscript{$\pm$0.02}  \\ 
\bottomrule
\end{tabular}%
}
\caption{Performance of~\ourmodel~on African LID ($F_1$). \textsuperscript{$\star$} Results as reported in \newcite{afrolidpaper}.}
\label{tab:afrolid_results}
\end{table}

\begin{table}[h!]
\small
\centering
\begin{tabular}{lcc}\toprule
                         & \textbf{AfroLID} & \textbf{\ourmodel} \\ \midrule
Amharic (amh)         & 97.00               &     \textbf{99.50 }\textsuperscript{$\pm$0.01}             \\
Hausa  (hau)         & 89.00               &     \textbf{98.09}\textsuperscript{$\pm$0.02}                \\
Igbo  (ibo)          & 46.00               &     \textbf{95.28}\textsuperscript{$\pm$0.00}               \\
Nigerian Pidgin (pcm) & 56.00               &     \textbf{77.73}\textsuperscript{$\pm$0.01}               \\
Swahili   (swh)       & 96.00               &    \textbf{98.66}\textsuperscript{$\pm$0.02}         \\
Yoruba     (yor)      & 82.00               &       \textbf{98.96}\textsuperscript{$\pm$0.00}           \\\bottomrule
\end{tabular}
\caption{Comparison between AfroLID \cite{afrolidpaper} and ~\ourmodel on AfriSenti Dev dataset.}
\label{tab:afrolid_vs_serengeti}
\end{table}
\subsection{Performance Analysis}
We report the results for seven of our eight tasks in Table \ref{tab:afronlu_test_results}. 

\noindent{\textbf{Named Entity Recognition (NER).}}
\ourmodel~sets a new SOTA on six out of eight datasets on the NER cluster. The lowest $F_1$ across all models are on NCHLT and Yoruba-Twi datasets (on both Dev and Test). \ourmodel~achieves best performance on both of these datasets on Test (with $73.18$ $F_1$ on the first and $71.25$ on the second). 


\noindent{\textbf{Phrase Chunking.}}
\ourmodel~outperforms all models on the phrase chunking task on both Dev and Test data, reaching $90.51$ $F_1$ on Test. 

\noindent{\textbf{Part of Speech (POS) Tagging.}}
In the POS tagging task, \ourmodel~outperformed all other models in the Dev. and Test sets. 

\noindent{\textbf{News Classification.}}
Our~\ourmodel~outperforms other models on three out of four datasets on Test data (and on two datasets on Dev).\footnote{Our \ourmodel-E110 outperforms \ourmodel~on one dataset in Dev and Test sets}. We do not report SOTA results for Amharic, Kirnews, and Kinnews datasets because their authors report performance in accuracy (and so are not comparable to our results). We show performance of \ourmodel~on each category in the news classification cluster in Figure \ref{fig:classification} in the Appendix.



\noindent{\textbf{Sentiment Analysis.}}
\ourmodel-E250 outperforms other models on one out of three tasks in our sentiment analysis task cluster. Afro-XMLR and AfriBERTa outperform other models on one each. To further investigate performance, we conduct an error analysis on the three sentiment datasets (see Figure~\ref{fig:confmatrix} in the Appendix).




\noindent{\textbf{Topic Classification.}}
AfriBERTa outperforms other models on both tasks in our topic classification cluster, followed by~\ourmodel. We show confusion matrices for Hausa and Yoruba topic classification in Figure~\ref{fig:topicclassification} in the Appendix.

\noindent{\textbf{Language Identification.}}
\ourmodel~outperforms AfroLID on AfroLID and AfriSenti data (see Table \ref{tab:afrolid_results} and \ref{tab:afrolid_vs_serengeti} for details). We also compare the performance of \ourmodel~to AfroLID, and  \href{https://github.com/wooorm/franc}{Franc}\footnote{A publicly available LID tool covering $88$ African languages.}, on the $88$ African languages represented in Franc in Table \ref{tab:serengeti_franc_vs_afrolid_test} (Appendix). \ourmodel~outperforms AfroLID and Franc with an average $F_1$ score of $96.29$. \ourmodel~outperforms both models on $59$ languages and has similar results with AfroLID on $19$ languages. 
 Next, we evaluate the performance of \ourmodel~on Creole languages. Again, we record improvement in results for Creole languages when compared with AfroLID. \ourmodel~outperforms AfroLID in $7$ out of $9$ languages and acquires similar scores on $2$ languages. We assume that the addition of the ten most spoken languages to the pretraining data for \ourmodel~may have helped the model learn the Creoles better. This is because Creoles share some features including vocabularies and syntax with some of those top ten languages. 

\subsection{Error Analysis}
In the sentiment analysis cluster, best performance is recorded for positive categories while negative categories have the worst performance. A fine-grained analysis of the Yoruba sentiment dataset found that \ourmodel~failed to correctly categorize sentiment if the polarity item(s) were not seen in training, can be associated with both positive and negative sentiments, the polarity item(s) is a negation, or if ambivalent markers are present in the sentence. We provide a table showing examples of each type of error we found in Table \ref{tab:yor_senti_error} in the Appendix. For the news classification task, politics and tourism are the best performing classes while education and relationships have the worst performance on kirnews and kinnews respectively. It is important to mention that the worst performing categories do not have the smallest data sizes. For the topic classification, the best performance is on the world class for Hausa topic modelling while entertainment and sport have best performance for Yoruba. The worst performance is on Nigeria and health for Hausa and Yoruba topic datasets respectively. 

\subsection{Imbalanced Distribution}
We find imbalances in the class distributions for all datasets except YOSM. We find a positive correlation between the size of each category in a dataset and the model accuracy. We also find a positive correlation with the number of examples in a specific class and the accuracy we acquire. We provide confusion matrices that represents the sizes of each category and the performance of \ourmodel~in Figures \ref{fig:num_classification}, \ref{fig:confmatrix_sent}, and \ref{fig:topicclassification_num} in the Appendix.

\begin{table*}[h!]
\small
\begin{tabular}{ccccccccccc||cccccc}
\toprule
             & \textbf{afr}  & \textbf{nbl}  & \textbf{nso}  & \textbf{sot}  & \textbf{ssw}  & \textbf{tsn} & \textbf{tso}  & \textbf{ven}  & \textbf{xho}  & \textbf{zul}  & \textbf{kin} & \textbf{lug}  & \textbf{nya}  & \textbf{run}  & \textbf{sna}  & \textbf{som}  \\ \toprule
\textbf{afr} & 1             & 0.28          & 0.35          & 0.26          & 0.27          & 0.36         & 0.29          & 0.22          & \textbf{0.42} & 0.38          & 0.34          & 0.38          & 0.26          & 0.25          & 0.25          & \textbf{0.43} \\
\textbf{nbl} & 0.28          & 1             & \textbf{0.47} & \textbf{0.41} & \textbf{0.62} & 0.26         & \textbf{0.48} & \textbf{0.42} & \textbf{0.41} & \textbf{0.55} & 0.37          & 0.35          & \textbf{0.48} & \textbf{0.43}          & \textbf{0.46} & 0.35          \\
\textbf{nso} & 0.35          & \textbf{0.47} & 1             & \textbf{0.55} & \textbf{0.47} & 0.38         & \textbf{0.51} & \textbf{0.40} & \textbf{0.42} & \textbf{0.50} & \textbf{0.40} & 0.38          & \textbf{0.42} & 0.39          & 0.39          & \textbf{0.42} \\
\textbf{sot} & 0.26          & \textbf{0.41} & \textbf{0.55} & 1             & \textbf{0.43} & 0.27         & \textbf{0.52} & \textbf{0.46} & 0.31          & \textbf{0.41} & 0.33          & 0.29          & \textbf{0.45} & \textbf{0.40} & 0.39          & 0.34          \\
\textbf{ssw} & 0.27          & \textbf{0.62} & \textbf{0.47} & \textbf{0.43} & 1             & 0.25         & \textbf{0.50} & \textbf{0.44} & 0.38          & \textbf{0.52} & 0.36          & 0.33          & \textbf{0.48} & \textbf{0.43} & \textbf{0.43} & 0.34          \\
\textbf{tsn} & 0.36          & 0.26          & 0.38          & 0.27          & 0.25          & 1            & 0.28          & 0.21          & 0.39          & 0.36          & 0.31          & 0.36          & 0.25          & 0.24          & 0.23          & 0.37          \\
\textbf{tso} & 0.29          & \textbf{0.48} & \textbf{0.48} & \textbf{0.52} & \textbf{0.50} & 0.28         & 1             & \textbf{0.47} & 0.37          & \textbf{0.48} & 0.38          & 0.34          & \textbf{0.51} & \textbf{0.44} & \textbf{0.44} & 0.37          \\
\textbf{ven} & 0.22          & \textbf{0.42} & \textbf{0.40} & \textbf{0.46} & \textbf{0.44} & 0.21         & 0.47          & 1             & 0.27          & 0.35          & 0.29          & 0.26          & \textbf{0.44} & 0.38          & \textbf{0.41} & 0.29          \\
\textbf{xho} & \textbf{0.42} & \textbf{0.41} & \textbf{0.42} & 0.31          & 0.38          & 0.39         & 0.37            & 0.27          & 1             & \textbf{0.56} & \textbf{0.41} & \textbf{0.47} & 0.35          & 0.33          & 0.32          & \textbf{0.45} \\
\textbf{zul} & 0.38          & \textbf{0.55} & \textbf{0.50} & \textbf{0.41} & \textbf{0.52} & 0.36         & \textbf{0.48} & 0.35          &      \textbf{0.56}    & 1             & \textbf{0.44} & \textbf{0.44} & \textbf{0.44} & \textbf{0.40} & 0.39          & \textbf{0.45} \\ \bottomrule
\end{tabular}
\caption{Jaccard Similarity for South African languages and some languages that are genealogically similar to them. Each of the $10$ South African languages are represented on each row. The genealogically similar languages we explore are after the horizontal lines. Specifically, we have: Kinyarwanda (kin), Luganda (lug), Chichewa (nya), Rundi (run), Shona (sna) and Somali (som). We highlight similarity scores of 0.4 and above in bold face.} \label{tab:sa_jaccard}
\end{table*}
\begin{table*}[htb!]
\centering
\small
\begin{tabular}{cccccccc}\toprule
\textbf{Dataset}                       & \textbf{Lang} & \textbf{XLMR} & \textbf{BERT} & \textbf{mBERT} & \textbf{Affro-XLMR} & \textbf{AfriBERTa} & \textbf{SERENGETI} \\\toprule 
\multirow{8}{*}{\textbf{NCHLT-NER}} & \textbf{afr}  &
80.68\textsuperscript{$\pm$0.75}  & \colorbox{red!10}{71.47} & 80.08\textsuperscript{$\pm$0.29}   & 80.55\textsuperscript{$\pm$0.11}      & \colorbox{red!10}{74.5}\textsuperscript{$\pm$0.64}        & \textbf{81.57}\textsuperscript{$\pm$0.59}                     \\
                                       & \textbf{nbl}  & \colorbox{red!10}{74.64}\textsuperscript{$\pm$0.66}  &  \colorbox{red!10}{61.02}& \colorbox{red!10}{73.48}\textsuperscript{$\pm$0.18}   & 75.26\textsuperscript{$\pm$0.28}      & \colorbox{red!10}{72.28}\textsuperscript{$\pm$0.67}       & \textbf{77.13}\textsuperscript{$\pm$0.67} \\
                                       & \textbf{nso}  &               \colorbox{red!10}{77.0}\textsuperscript{$\pm$1.23}   &  \colorbox{red!10}{64.27}& \colorbox{red!10}{78.75}\textsuperscript{$\pm$0.45}   & 80.13\textsuperscript{$\pm$0.51}      & \colorbox{red!10}{75.45}\textsuperscript{$\pm$1.09}       & \textbf{80.69}\textsuperscript{$\pm$0.64}\\
                                       & \textbf{sot}  &             \colorbox{red!10}{54.71}\textsuperscript{$\pm$1.51} & \colorbox{red!10}{49.75} & \colorbox{red!10}{54.68}\textsuperscript{$\pm$0.49}   & 55.57\textsuperscript{$\pm$0.2}       & \colorbox{red!10}{54.09}\textsuperscript{$\pm$0.98}       & \textbf{56.26}\textsuperscript{$\pm$1.52} \\
                                       & \textbf{ssw}  &              \colorbox{red!10}{71.75}\textsuperscript{$\pm$0.65}  &  \colorbox{red!10}{65.18}& \colorbox{red!10}{71.24}\textsuperscript{$\pm$0.75}   & 72.35\textsuperscript{$\pm$1.02}      & \colorbox{red!10}{69.38}\textsuperscript{$\pm$0.58}       & \textbf{73.37}\textsuperscript{$\pm$0.82} \\
                                       & \textbf{tsn}  &               \colorbox{red!10}{77.02}\textsuperscript{$\pm$0.22}  & \colorbox{red!10}{70.96} & \colorbox{red!10}{76.35}\textsuperscript{$\pm$0.47}   & 77.68\textsuperscript{$\pm$0.96}      & \colorbox{red!10}{73.89}\textsuperscript{$\pm$1.41}       & \textbf{79.05}\textsuperscript{$\pm$0.75}\\
                                       & \textbf{tso}  &               \colorbox{red!10}{74.24}\textsuperscript{$\pm$0.08} & \colorbox{red!10}{65.09} & \colorbox{red!10}{72.95}\textsuperscript{$\pm$0.67}   & \colorbox{red!10}{74.85}\textsuperscript{$\pm$0.43}      & \colorbox{red!10}{71.05}\textsuperscript{$\pm$0.9}        & \textbf{75.13}\textsuperscript{$\pm$0.31}\\
                                       & \textbf{ven}  &              \colorbox{red!10}{64.06}\textsuperscript{$\pm$0.31} & \colorbox{red!10}{61.51} & \colorbox{red!10}{63.11}\textsuperscript{$\pm$1.27}   & \colorbox{red!10}{64.39}\textsuperscript{$\pm$0.36}      & \colorbox{red!10}{63.24}\textsuperscript{$\pm$1.26}       & \textbf{65.42}\textsuperscript{$\pm$0.76} \\
                                       & \textbf{xho}  &               70.77\textsuperscript{$\pm$2.45} & \colorbox{red!10}{58.17} & \colorbox{red!10}{68.54}\textsuperscript{$\pm$1.44}   & 72.37\textsuperscript{$\pm$0.39}      & \colorbox{red!10}{67.00}\textsuperscript{$\pm$1.27}        & \textbf{72.92}\textsuperscript{$\pm$0.29} \\
                                       & \textbf{zul}  &               \colorbox{red!10}{69.44}\textsuperscript{$\pm$0.62} & \colorbox{red!10}{54.27} & \colorbox{red!10}{67.74}\textsuperscript{$\pm$1.46}   & 70.28\textsuperscript{$\pm$0.49}      & \colorbox{red!10}{67.17}\textsuperscript{$\pm$0.15}       & \textbf{71.20}\textsuperscript{$\pm$0.44}  \\\hline
\multirow{8}{*}{\textbf{Phrase Chunk}} & \textbf{afr}  & 95.34\textsuperscript{$\pm$0.16} &   \colorbox{red!10}{89.92}   & 95.68\textsuperscript{$\pm$0.30}          & 95.13\textsuperscript{$\pm$0.06}            & \colorbox{red!10}{90.22}\textsuperscript{$\pm$0.81}             &\textbf{ 96.01}\textsuperscript{$\pm$0.14}               \\
                                       & \textbf{nso}  &               \colorbox{red!10}{96.57}\textsuperscript{$\pm$0.61}       & \colorbox{red!10}{95.26} & \colorbox{red!10}{96.85}\textsuperscript{$\pm$0.55}         & \textbf{98.36}\textsuperscript{$\pm$0.2}             & \colorbox{red!10}{96.47}\textsuperscript{$\pm$0.14}             & 98.28\textsuperscript{$\pm$0.1}                    \\
                                       & \textbf{sot}  &               \colorbox{red!10}{82.93}\textsuperscript{$\pm$0.38}       &\colorbox{red!10}{80.59}  & \colorbox{red!10}{83.08}\textsuperscript{$\pm$0.78}         & 85.28\textsuperscript{$\pm$0.61}            & \colorbox{red!10}{82.18}\textsuperscript{$\pm$0.93}             & \textbf{85.69}\textsuperscript{$\pm$0.76}\\
                                       & \textbf{ssw}  &\colorbox{red!10}{82.9}\textsuperscript{$\pm$1.03}         &\colorbox{red!10}{82.09} &  \colorbox{red!10}{81.91}\textsuperscript{$\pm$0.47}         & \textbf{84.73}\textsuperscript{$\pm$0.18}            & \colorbox{red!10}{83.24}\textsuperscript{$\pm$0.11}             & 83.45\textsuperscript{$\pm$0.12}\\
                                       & \textbf{tsn}  &  \colorbox{red!10}{92.77}\textsuperscript{$\pm$0.16}       & \colorbox{red!10}{92.09}& \colorbox{red!10}{92.64}\textsuperscript{$\pm$0.66}         & 94.11\textsuperscript{$\pm$0.49}            & \colorbox{red!10}{92.71}\textsuperscript{$\pm$0.42}             & \textbf{94.03}\textsuperscript{$\pm$0.19}\\
                                       & \textbf{tso}  &\colorbox{red!10}{86.42}\textsuperscript{$\pm$0.46}       & \colorbox{red!10}{86.75} & \colorbox{red!10}{86.90}\textsuperscript{$\pm$0.31}          & \colorbox{red!10}{87.39}\textsuperscript{$\pm$0.18}            & \colorbox{red!10}{86.73}\textsuperscript{$\pm$0.95}             & \textbf{89.32}\textsuperscript{$\pm$0.43}\\
                                       & \textbf{ven}  &        \colorbox{red!10}{92.31}\textsuperscript{$\pm$0.45}       & \colorbox{red!10}{92.32} & \colorbox{red!10}{90.47}\textsuperscript{$\pm$0.32}         & \colorbox{red!10}{92.42}\textsuperscript{$\pm$0.68}            & \colorbox{red!10}{92.02}\textsuperscript{$\pm$0.33}             & \textbf{92.54}\textsuperscript{$\pm$0.21}\\
                                       & \textbf{zul}  &               \colorbox{red!10}{87.30}\textsuperscript{$\pm$0.26}      & \colorbox{red!10}{84.93}  & \colorbox{red!10}{87.29}\textsuperscript{$\pm$1.04}         & 88.67\textsuperscript{$\pm$0.66}            & \colorbox{red!10}{85.74}\textsuperscript{$\pm$0.55}             & \textbf{90.05}\textsuperscript{$\pm$0.81}\\\bottomrule               
\end{tabular}
\caption{Performance of mPLMs and BERT on each language in NCHLT-NER and Phrase-Chunk datasets we use for the genealogy analysis. ($F_1$) score is the evaluation metric. We use \colorbox{red!10}{\textbf{Red}} highlights to indicate languages in zero-shot setting. We evaluate BERT, a monolingual model as a sanity check for our evaluation.}
\label{tab:genealogy_analysis}
\end{table*}
\subsection{Genealogy \& Language Contact}
Our preliminary analyses show that language similarity may improve model performance in zero-shot settings. This we believe is due to high cross-lingual transfer information \cite{conneau-etal-2020-unsupervised} from similar languages. Similar languages often share many features (e.g., vocabulary, syntax, and script) sometimes up to a point of mutual intelligibility \cite{Nassenstein_2019, Arndt_2015, roy2006state}. Languages in contact may also have such similarities. By \textit{language in contact}, we mean all languages that speakers of a specific language interact with and influence. A language can be in contact with another due to trade, geographic proximity, migration, or even colonization. Languages in contact can influence each other in multiple ways, such as borrowing words, grammatical structures, phonology, or orthographic conventions \cite{matras_2009}. To illustrate our hypothesis, we select two datasets with South African (SA) languages in AfroNLU - NCHLT-ner and phrase-chunk. We select SA languages because they are contact languages (see Figure \ref{fig:nigercongo} in Appendix for a genealogical classification tree that highlights the SA languages.) \cite{Nassenstein_2019, Arndt_2015, roy2006state}.

To determine the significance of language similarity and language contact in our own zero-shot settings, we measure the Jaccard similarity between the pretraining data for the SA languages (see Table \ref{tab:sa_jaccard}). We find strong similarities between some of these languages (see bolded examples in Table \ref{tab:sa_jaccard}). We also finetune a BERT model and compare the performance of BERT with MBERT. We do this because BERT does not include any similar language in its representation. 

XLM-R, mBERT, and AfriBERTa are not trained on most SA languages but have high scores in zero-shot settings see Table \ref{tab:genealogy_analysis} and Table~\ref{tab:langs_test_results} in Appendix. We argue that XLM-R in addition to cross-lingual transfers from other languages acquires representation from afr and xho where xho alone shares more than 0.4 similarity with afr, nbl, nso, and zul. mBERT also learns representation from afr while AfriBERTa learns representations from Gahuza which is a code-mixed variety of KIN and RUN. BERT on the other hand significantly performs lower than MBERT in all languages except on ssw, and ven (Phrase chunk).~\ourmodel, however, outperforms other models on these languages which demonstrates the impact of pretraining on each of these languages.

These analyses are in no way conclusive, but do provide  insights on how linguistic information may impact model performance in zero-shot settings. Future work can further probe the influence of similar languages in a more in-depth fashion. (See Appendix \ref{app:lang_contact} for detailed analysis).

\section{Conclusion}\label{sec:conc}
We reported our efforts to develop~\ourmodel, a suite of three massively multilingual language models for African NLP. \ourmodel~outperforms $4$ mPLMs on $11$ datasets across $8$ tasks. We provide extensive evaluations of model outputs, including zero-shot performance of the mPLMs. We also offer broad linguistically-motivated analyses of model performance. 
 \section{Limitations}\label{sec:limits}

We identify the following limitations for our work:
\begin{enumerate}
\item 
Due to limited access to a wide network of native speakers from the majority of languages, we were able to manually inspect only a subset of languages present in our pretraining data. Specifically, we could only manually evaluate Afrikaans, Yorùbá, Igbo, Hausa, Luganda, Kinyarwanda, Chichewa, Shona, Somali, Swahili, Xhosa, Bemba, and Zulu. Future work should focus on increasing the subset of languages evaluated manually in order to ensure quality. We believe automatic analyses are not sufficient before development of models that get deployed in particular applications.

\item Another limitation is related to our inability to perform extensive analysis of biases and hateful speech present in our pretraining data. Again, this is due to relatively restricted access to native speakers (and even automated tools) to perform this analysis. As a result, we cannot fully ensure that our models is free from biases and socially undesirable effects. Therefore, it is important that these models be used with care and caution, and be analyzed for biases and socially undesirable effects before use.

\item Additionally, due to unavailability of sufficient computing resources, we were unable to evaluate large language models such as BLOOM, even though it covers $22$ African languages. 

\item Finally, even though AfroNLU has diverse tasks at the word and sentence level, these tasks only cover few African languages. We therefore encourage the creation of more datasets for downstream NLU tasks in more (and more diverse) African languages. We believe broader benchmarks will continue to be important for future progress in African NLP.
\end{enumerate}

\section{Ethics Statement and Wider Impacts}\label{sec:ethics}
\ourmodel~aligns with Afrocentric NLP where the needs of African people is put into consideration when developing technology. We believe \ourmodel~will not only be useful to speakers of the languages supported, but also researchers of African languages such as anthropologists and linguists. We discuss below some use cases for~\ourmodel~and offer a number of broad impacts.
\begin{enumerate}
    \item \ourmodel~aims to address the lack of access to technology in about $90\%$ of the world's languages, which automatically discriminates against native speakers of those languages. More precisely, it does so by focusing on Africa. To the best of our knowledge,~\ourmodel~is the first massively multilingual PLM developed for African languages and language varieties. A model with knowledge of $517$ African languages, is by far the largest to date for African NLP. 
    \item \ourmodel~enables improved access of important information to the African community in Indigenous African languages. This is especially beneficial for people who may not be fluent in other languages. This will potentially connect more people globally. 
    \item \ourmodel~affords opportunities for language preservation for many African languages. To the best of our knowledge, \ourmodel~consists of languages that have not been used for any NLP task until now. We believe that it can help encourage  continued use of these languages in several domains, as well as trigger future development of language technologies for many of these languages.
    \item To mitigate discrimination and bias, we adopt a manual curation of our datasets. Native speakers of Afrikaans, Yorùbá, Igbo, Hausa, Luganda, Kinyarwanda, Chichewa, Shona, Somali, Swahili, Xhosa, Bemba, and Zulu also manually evaluated a subset of the data to ensure its quality. The data collected for this work is taken from various domains to further ensure a better representation of the language usage of native speakers.
    \item Although LMs are useful for a wide range of applications, they can also be misused.~\ourmodel~is developed using publicly available datasets that may carry biases. Although we strive to perform analyses and diagnostic case studies to probe performance of our models, our investigations are by no means comprehensive nor guarantee absence of bias in the data. In particular, we do not have access to native speakers of most of the languages covered. This hinders our ability to investigate samples from each (or at least the majority) of the languages.
\end{enumerate}

\section*{Acknowledgements}\label{sec:acknow}
MAM gratefully acknowledges support from Canada Research Chairs (CRC), the Natural Sciences and Engineering Research Council of Canada (NSERC; RGPIN-2018-04267), the Social Sciences and Humanities Research Council of Canada (SSHRC; 435-2018-0576; 895-2020-1004; 895-2021-1008), Canadian Foundation for Innovation (CFI; 37771), Digital Research Alliance of Canada,\footnote{\href{https://alliancecan.ca}{https://alliancecan.ca}} UBC ARC-Sockeye,\footnote{\href{https://arc.ubc.ca/ubc-arc-sockeye}{https://arc.ubc.ca/ubc-arc-sockeye}} Advanced Micro Devices, Inc. (AMD), and Google. Any opinions, conclusions or recommendations expressed in this material are those of the author(s) and do not necessarily reflect the views of CRC, NSERC, SSHRC, CFI, the Alliance, AMD, Google, or UBC ARC-Sockeye.
\bibliography{anthology,custom}
\bibliographystyle{acl_natbib}

\appendix
\clearpage
\appendixpage
\addappheadtotoc
\counterwithin{figure}{section}
We provide an overview of the Appendix. 

\begin{figure*}[h!]
  \centering
  \includegraphics[width=\textwidth]{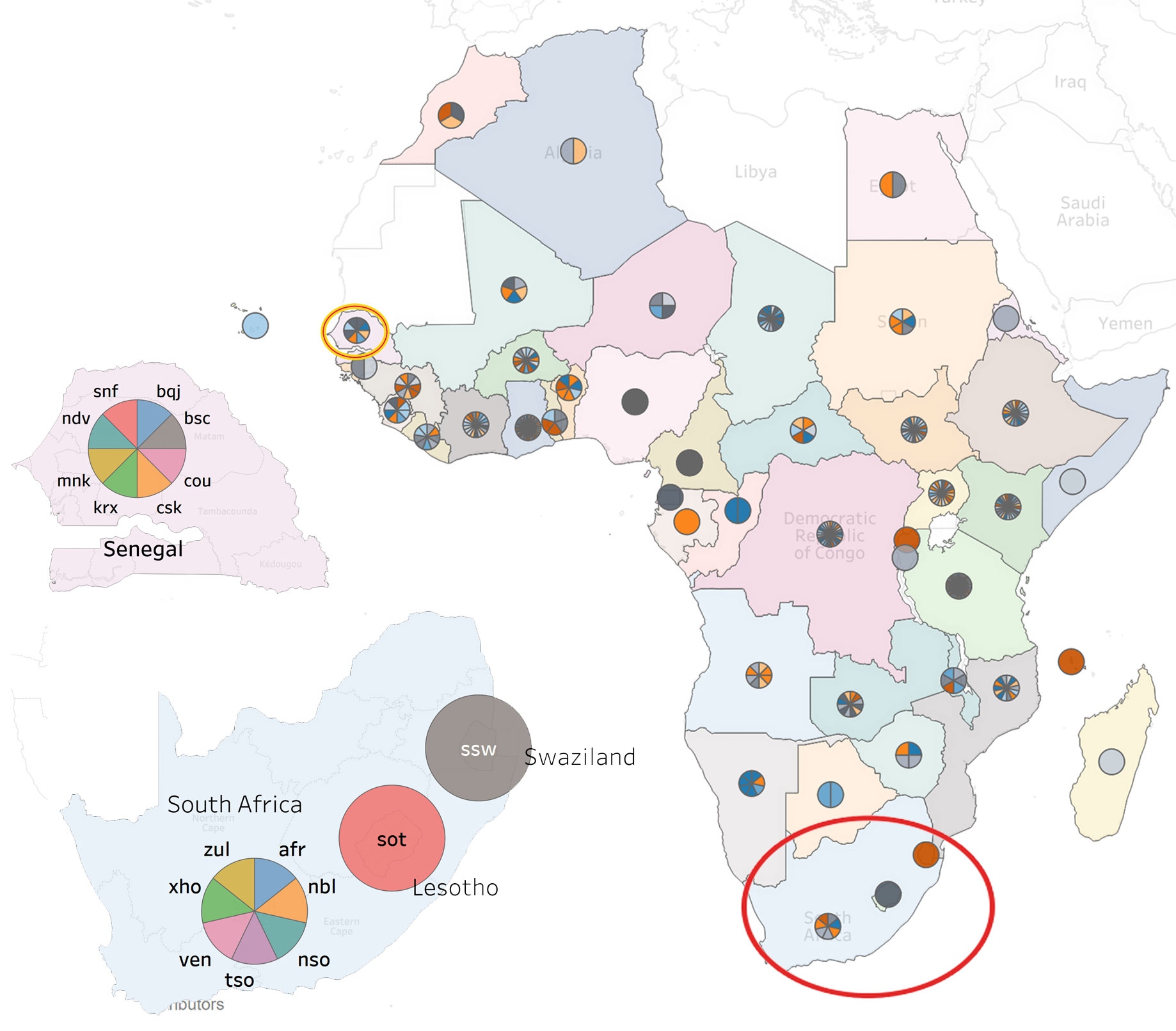}
\caption{\small All $517$ languages in our dataset across the $50$ African countries our data comes from. The language varieties are represented as colored pie shapes within each country. We zero in on South Africa, Lesotho, Swaziland, and Senegal to show detail.}
\label{appdx-fig:countries} 
\end{figure*}

\noindent\textbf{Introduction}
\begin{itemize}
\item We share a large map of Africa showing the $517$ Languages covered in our pretraining data in Figure \ref{appdx-fig:countries}. 
\item We also share the scripts represented in our pretraining data in Table \ref{tab:scripts}. 
\end{itemize}
\noindent\textbf{Literature Review}
\begin{itemize}
    \item We provide a more extensive literature review in \ref{app:literature_review}. We discuss Afrocentric NLP, multilingualism in NLP, diversity and inclusion in NLP and multilingual language models.
\end{itemize} 

\noindent\textbf{Pretraining Data}
We discuss the pretraining data in more detain in Section \ref{app:pretraing-data}. 

\noindent\textbf{Typology Information for AfroNLU}
In Section \ref{app:typology} we discuss $6$ families that cover the languages in $18$ datasets in AfroNLU. For each family, we provide visualizations that cover any number of languages in the $18$ datasets. We provide visualizations for:
\begin{itemize}
\item Afro-Asiatic in Figure \ref{fig:afroasiatic}, 
\item Austronesian in Figure \ref{fig:austronesean}, 
\item Creole in Figure \ref{fig:creole}, 
\item Indo-European in Figure \ref{fig:indoeuropean}, 
\item Niger-Congo in Figure \ref{fig:nigercongo}, and 
\item Nilo-Saharan in Figure \ref{fig:nilosaharan}.
\end{itemize}

\noindent\textbf{Evaluation}
We provide more information about the evaluations. We do the following:
\begin{itemize}
    \item We show \ourmodel's performance on the Dev. set in Table \ref{tab:afronlu_dev_results}.
    \item We show \ourmodel's performance on each language in our multilingual datasets in Table \ref{tab:langs_test_results}.
    \item We perform error analysis and show examples of errors in Table \ref{tab:yor_senti_error}. We also show confusion matrices for the news classification, sentiment analysis, and topic classification clusters in in Figure \ref{fig:classification}, Figure \ref{fig:confmatrix}, and Figure \ref {fig:topicclassification}.
    \item We discuss the implications of imbalanced distribution and show confusion matrices for the news classification, sentiment analysis, and topic classification clusters in  Figure \ref{fig:num_classification}, Figure \ref{fig:confmatrix_sent}, and Figure \ref{fig:topicclassification_num}.
    \item We show results from comparing \ourmodel~ with AfroLID and Franc on AfroLID test set in Table \ref{tab:afrolid_vs_serengeti}.
    \item Information about the languages in our pretraining data is provided in Table \ref{tab:lang_listI}, Table \ref{tab:lang_listII} and Table \ref{tab:lang_listIII}.
    \item We share statistics of the top ten languages with the largest data in \ourmodel~ and the ten languages with the least dataset in Table \ref{tab:freq}.
    \end{itemize}
\noindent\textbf{Genealogy /Language Contact Analysis}
We further analyize our claim on the interaction of similar languages and zero-shot settings in Section \ref{app:lang_contact}. 
\begin{itemize}
\item We create a Figure highlighting the languages er perform analysis on in Figure \ref{fig:geneaology_highlight}. 
\item We show the Jaccard similarity scores in Table \ref{tab:sa_jaccard}. \item Next we show the results of each language in zero-shot settings and results for finetuning on BERT in Table \ref{tab:genealogy_analysis}.
\end{itemize}
\counterwithin{table}{section}
\section{Introduction}\label{app:intro}

\begin{table}[!ht]
\begin{center}
\small 
\centering
\setlength{\tabcolsep}{5pt}
\begin{tabular}{ll}
 \toprule
\multicolumn{1}{l}{\textbf{Script}} & \multicolumn{1}{l}{\textbf{Languages }} \\ 
 \midrule
 Ethiopic & Amharic, Basketo, Maale, \\
 & \textsuperscript{$\star$}Oromo, Sebat Bet Gurage \\
& Tigrinya, Xamtanga\\\hdashline
 Arabic & Fulfude Adamawa, Fulfude Caka \\ 
 & Tarifit \\\hdashline
 Vai & Vai   \\\hdashline
 Coptic & Coptic \\
 \bottomrule
\end{tabular}
\end{center}\caption{Scripts represented in \ourmodel.}
\label{tab:scripts}
\end{table}
\section{Literature Review}\label{app:literature_review}
Representation learning is an integral part of modern NLP systems. It has significantly improved the state of the art in natural language understanding (NLU) and natural language generation (NLG). We now discuss Afrocentric NLP, Multilingualism in NLP, Diversity and Inclusion in NLP, MLMs, and LMs for African languages.
\subsection{Afrocentric NLP}
More than $2,000$ Indigenous languages are spoken in Africa, which is about a third of all languages spoken in the world \cite{ethnologue}. Unfortunately, the majority of these languages have not received any NLP attention to date. Rather, most NLP research has focused on higher resource languages. Most of these resourceful languages are typologically very different from Indigenous African languages. Methods used to develop technologies for these  languages remain \textit{Western-centric}, and may not be directly extensible to Indigenous African languages \cite{adebara-abdul-mageed-2022-towards}. Existing NLP technologies also mostly function within the contexts of values and beliefs that reflect western societies and pose unique challenges if the technologies are applied within African communities. 


Afrocentric NLP adopts a holistic approach to NLP throughout the life cycle of NLP policy making to model development and deployment. It discourages the current language data gas flaring policies that have led to the low resource status of many Indigenous African languages. Afrocentric NLP entails an understanding of the need for multi-dimensional policies that influence the language policy in education, media, government, and other domains to create ever-increasing, multi-domain, big data sources for NLP. During the archival and collection of language data, Afrocentric NLP necessitates respect of user consent, data sovereignty, wishes of local communities, and privacy \cite{sutherland2018digital, daigle2021data, 10.1093/idpl/ips014}. For model development, approaches tailored to the unique typological features of African languages are of utmost priority. This also means development of models that understand simple to complex tones--a common feature in about $80\%$ of African languages--serial verb constructions, and many other features ~\cite{hyman2003african, creissels2008africa}. Afrocentric NLP also prioritizes deploying models in formats that people without programming experience can easily use. Furthermore, from an Afrocentric approach, development of certain NLP applications such as language models, language identification tools, spelling checkers, language specific keyboards, and machine translation systems is crucial to advance NLP for African languages. 

\subsection{Multilingualism in NLP}
Multilingualism, the ability to handle multiple languages within a single system or model, is becoming increasingly important as the amount of text and speech data in many languages increase. NLP systems capable of handling multiple languages can provide greater access to information and communication for people who speak languages other than those most commonly used or supported by NLP. 

Multilingualism in NLP \cite{ruder2022statemultilingualai} is mainly achieved through building \texttt{(1)} a single model trained on several languages \cite{devlin-etal-2019-bert, conneau-etal-2020-unsupervised} and \texttt(2) transfer learning \cite{JMLR:v21:20-074, he2022towards, ruder-etal-2019-transfer}. In the former, large transformer models have achieved state-of-the-art on many tasks while the latter has enabled the use of low-resource languages through finetuned on various NLP tasks. Due to lack of adequate (or good quality) pretraining data \cite{caswell2021quality}, transfer learning is often the most accessible method for a few low resource languages. Unfortunately, about $94\%$ of the world's languages are either \textit{left-behinds}, in that it is probably impossible to build NLP resources for them, or \textit{scraping-bys} with no labelled datasets \cite{joshi-etal-2020-state}. For the left-behinds, labelled and unlabelled data is unavailable and even transfer learning approaches are beyond reach. So far, to the best of our knowledge, the largest multilingual model for African languages is pretrained on only $28$ African languages \cite{afroLM}. 

Most multilingual models are often trained with no more than $100$ languages because increasing the number of language would mean decreasing its capacity to learn representations of each language~\cite{conneau-etal-2020-unsupervised}. Nevertheless, increasing model size was shown to ameliorate this problem \cite{goyal-etal-2021-larger}. In some cases, these benchmarks are translations from English \cite{artetxe-etal-2020-cross, nzeyimana-niyongabo-rubungo-2022-kinyabert, ponti-etal-2020-xcopa} and may not necessarily be a good evaluation for the languages. This is because translating from a source language may mask concept gaps and differences in linguistic constituents \cite{Guillaume} in the target language. That is, translations are at best approximations of the target language \cite{adebara-abdul-mageed-2022-towards, joshi-etal-2020-state}. For example, when translating into English (which marks (in)definiteness morphologically) from Yor{\`u}b{\'a} (which uses bare nouns but marks these features contextually), ambiguities arise \cite{adebara-etal-2022-linguistically}.

For evaluation of multilingual models, several benchmarks have been created\cite{artetxe-etal-2020-cross} with most of these supporting English and other high-resource languages. More recently, a few evaluation sets were introduced for African languages~\cite{ifeoluwa2021masakhaner, shode_africanlp, niyongabo-etal-2020-kinnews}.We include these evaluation sets in our benchmark, which we henceforth refer to as AfroNLU. 


When evaluating multilingual models, reporting model performance for each language in the benchmark is preferred because reporting the results as a single value on all languages may mask the model's performance on individual languages~\cite{ruder2022statemultilingualai}. Large pre-training data, fine-tuning data, and evaluation benchmarks remain open challenging questions for achieving progress in multilingual NLP. For \ourmodel, we report results for each language in each benchmark across the $9$ tasks we evaluate on. 

\subsection{Diversity and Inclusion in NLP}
Diversity relates to the level of variety within a system. It is the measure of distinctiveness between the various individuals within a group. Inclusion on the other hand relates to the level of representation or alignment of an individual within a group and the ability for that individual to function to its fullest ability \cite{Fosch-Villaronga2022, Mitchell_2020}. Diversity and inclusion in NLP has gained increasing attention in recent years. In general, there is an acknowledgement that over-representation (and under-representation) of certain groups in the data used to train models \cite{Mitchell_2020} can be amplified by resulting technologies. This raises concerns about the technology and how it is that it can further existing biases and societal inequalities. But these biases can be exhibited in various ways beyond training data, including the algorithms implemented, the diversity of researchers and engineers developing the models, and the societal and cultural context in which they are used. 

Although this is starting to change, often times most of the data exploited in NLP models come from closely related Western languages. Most of these languages are Indo-European~\cite{aji-etal-2022-one, joshi-etal-2020-state}, and many of them share close geographic proximity and typology. In addition, the people who speak these languages have similar cultures. The implication is that several linguistic phenomena and typologies are under-represented in NLP data while those prevalent in Indo-European languages are over-represented \cite{chakravarthi-muralidaran-2021-findings}. About $88.38\%$ of the $2,679$ languages whose typology is described in WALS~\cite{dryer2013order} have not been used in NLP~\cite{joshi-etal-2020-state}. Many ideas and topics, alien to Western cultures have also never been seen~\cite{adebara-abdul-mageed-2022-towards, Bender_2011} in NLP data. African languages--and indeed many low resource languages--have rich linguistic typology, probably not seen in any other language in the world \cite{Bender_2011}. An obvious problem with the current lack of diversity in NLP data is that the methods and models developed have overfit to these Indo-European typologies and cannot generalize to other typologies. Similarly, machine translation systems have been found to exhibit gender, racial~\cite{NIPS2016_a486cd07, caliskan2017, chakravarthi-muralidaran-2021-findings} and stylistic biases~\cite{hovy-etal-2020-sound} in their outputs perpetuated through the data used for training.

A number of studies have also found that algorithms could exhibit biases~\cite{HOOKER2021100241, pmlr-v81-buolamwini18a, dwork_2011}. For example, a recent study that investigated performance of Amazon Transcribe and Google Speech-To-Text on British English reported notably higher error rates for second language speakers of different varieties of British English~\cite{markl_2022}. In another study, an evaluation of automatic speech recognition systems show substantial performance differences between 'standard' US English and African American English (AAE) varieties~\cite{koenecke_2020}. In this study, commercial ASR systems developed by Amazon, Apple, Google, IBM, and Microsoft were evaluated and higher rates of errors were recorded for speakers of AAE than speakers of standard US varieties. Similar studies have also recorded higher errors in non-white users of English \cite{WASSINK202250, martin2020understanding}. Other studies also reported differences in the performance of Youtube's automatic caption in different settings. One study reported higher accuracy in the transcriptions of US English compared with Indian English \cite{meyer-etal-2020-artie}. Another reported lower accuracy scores for women and speakers of Scottish English \cite{tatman-2017-gender} and non-white speakers of English \cite{tatman17_interspeech}.

Apart from data and algorithmic biases, the diversity crises in AI research is also argued to perpetuate historical biases~\cite{pmlr-v142-freire21a}. A more inclusive and diverse workforce could promote the exploration of questions and solutions beyond currently investigated research questions \cite{Fosch-Villaronga2022}. Several initiatives have been adopted to increase diversity in AI, including providing travel grants to marginalized communities to attend conferences, creating mentoring opportunities, special workshops, and community diversity chairs. A number of organizations have also been developed to promote diversity and inclusion in AI and NLP, such as \href{https://www.masakhane.io/}{Masakhane}, \href{https://blackinai.github.io/}{Black in AI}, \href{https://latinxinai.github.io/}{LatinX in AI}.

The impact of using biased systems in decision making have been extensively studied. Algorithmic decision-making using biased systems have been shown to have significant discriminatory effects in health~\cite{ziad_2019, eubanks_2018}, employment~\cite{solon_2016}, housing~\cite{pmlr-v81-buolamwini18a, solon_2016}, government benefit allocation~\cite{eubanks_2018}, policing~\cite{pmlr-v81-buolamwini18a, solon_2016, angwin_2018}, and freedom~\cite{angwin_2018}. Lack of diversity also has implication on access to technology. Currently, due to the use of a few high resource languages in NLP, there is limited global access to important applications such as machine translation, speech processing, information retrieval, and sentiment analysis. These technologies play an important role in ensuring a language thrives and offer major contributions to ongoing communication, literacy, education, and translation efforts in communities worldwide. These languages which have barely been used for NLP, usually referred to as low-resource languages, represent more than 90\% of the world's $7,000$ languages~\cite{joshi-etal-2020-state}. The current focus of NLP on resource-rich languages does also have aggravating effects on the language endangerment problem which has been of serious concern for linguistics and language policy around the world. An alarming $50-90\%$ of languages have been envisaged to go extinct by the end of the century due to the domination by some of these resource-rich languages~\cite{BESACIER201485}.




Overall, diversity and inclusion in NLP remain active areas of research and comprise pressing issues of international significance. \ourmodel~contributes to diversity and inclusion in NLP as follows: \textbf{(1)} We develop \ourmodel, a suite of massively, multilingual language models that support $517$ African languages and language varieties. To the best of our knowledge, more than $400$ of these languages have never been represented in any language model to date. \textbf{(2)} The languages we support belong to $14$ language families. \textbf{(3)} We provide a massive benchmark covering $28$ languages across eight different tasks.

\subsection{Multilingual Language Models}

MLMs have proven effective for cross-lingual NLU and NLG, often outperforming monolingual language models~\cite{conneau-etal-2020-unsupervised}. Different objectives have been adopted for training \cite{doddapaneni2021primer}, using Transformer architectures. These LMs use one of the three different variants of Transformer architectures--encoder-decoder, encoder-only and decoder-only \cite{cai_2022}. 

In the encoder-decoder models, input is encoded by the encoder side and the decoder conducts the operation to predict the sequence one token at a time or just reconstruct it by denoising. MBART \cite{liu-etal-2020-multilingual-denoising}, AfriTeva \cite{jude-ogundepo-etal-2022-afriteva}, M2M100 \cite{m2m100_2020}, and MT5 \cite{xue-etal-2021-mt5} are representatives for this architecture. Encoder-only models use only the encoder part of the transformer architecture, while decoder-only models use its decoder only. Some examples of encoder-only models are BERT \cite{devlin-etal-2019-bert}, XLMR \cite{conneau-etal-2020-unsupervised}, and Electra \cite{chi2021xlme}, while BLOOM \cite{Scao2022BLOOMA1}, GPT \cite{radford2018improving, radford2019language, brown_2020}, OPT \cite{opt_2022} are examples of decoder-only models. Most LMs developed for African languages use an encoder-only architecture, except AfriTEVA and AfroT5 which use encoder-decoder architectures.   


These models are further finetuned on specific tasks. Finetuning has demonstrated its effectiveness on various NLU and NLG downstream tasks including part of speech tagging~\cite{conneau-etal-2020-unsupervised}, named entity recognition~\cite{ushio-camacho-collados-2021-ner, conneau-etal-2020-unsupervised}, and question answering~\cite{conneau-etal-2020-unsupervised}. Finetuning follows a transfer learning approach which attempts to transfer knowledge from other sources to benefit a current task. This is based on the premise that previous knowledge may improve solutions for a current task \cite{5288526, JMLR:v21:20-074, he2022towards, ruder-etal-2019-transfer}. Transfer learning allows the domains, tasks, and distributions used in training and testing to be different thereby enabling a new task to leverage previously acquired domain knowledge. Potential benefits include faster learning, better generalization, and a more robust system. In the real world, we find many examples of transfer learning where humans transfer previous knowledge while learning or performing a task. For instance, knowing how to play the piano may facilitate learning to play the guitar and knowing how to ride a bicycle may facilitate learning to ride a motorbike. Finetuning is thus done by reusing the LM's parameters as a starting point, while adding one task-specific layer trained from scratch. Finetuning can be done on an individual or joint basis~\cite{kitaev-etal-2019-multilingual}. In the former, a model is finetuned on single language for a specific downstream task. In the later, training data from a combination of multiple languages can be jointly finetuned in a single model.

\section{Pretraining Data}\label{app:pretraing-data}
We provide details of our pretraining data below:
\noindent{\textbf{Religious Domain.}}
Our religious data is taken from online Bibles, Qurans, and data crawled from the Jehovah's witness website. We also include religious texts from the book of Mormon. 

\noindent{\textbf{News Domain.}}
We collect data from online newspapers~\cite{adebara-abdul-mageed-2022-towards} and news sites such as Voice of America, Voice of Nigeria, BBC, Global voices, and DW news sites. We collect local newspapers from $27$ languages from across Africa. 

\noindent{\textbf{Government Documents.}}
We collect government documents \href{https://www.sadilar.org/index.php/en/}{South African Centre for Digital Language Resources} (SADiLaR), and the \href{https://www.ohchr.org/en/udhr/pages/searchbylang.aspx}{Universal Declaration of human rights} (UDHR) in multiple languages.

\noindent{\textbf{Health Documents.}}
We collect multiple health documents from the Department of Health, State Government of Victoria, Australia. We collect documents in Amharic, Dinka, Harari, Oromo, Somali, Swahili, and Tigrinya.

\noindent{\textbf{Existing Corpora.}}
We collect corpora available on the web for different African languages, including from \href{https://www.gutenberg.org/browse/languages/af}{Project Gutenberg} for Afrikaans, \href{https://zenodo.org/record/3668495#.YcTXu2DMJyy}{South African News data.} for Sepedi and Setswana, OSCAR \cite{AbadjiOrtizSuarezRomaryetal.2021} for Afrikaans, Amharic, Somali, Swahili, Oromo, Malagasy, and Yoruba. We also used \href{https://opus.nlpl.eu/Tatoeba.php}{Tatoeba} for Afrikaans, Amharic, Bemba, Igbo, Kanuri, Kongo, Luganda, Malagasy, Sepedi, Ndebele, Kinyarwanda, Somali, Swahili, Tsonga, Xhosa, Yoruba, and Zulu; \href{https://zenodo.org/record/3553423#.YcTXkWDMJyx}{Swahili Language Modelling Data} for Swahili; \href{https://github.com/ijdutse/hausa-corpus/blob/master/data/all-merged-hausa-datasets.txt}{Ijdutse corpus} for  Hausa; \href{https://github.com/AI-Lab-Makerere/Data4Good}{Data4Good corpora} for Luganda, CC-100 for Amharic, Fulah, Igbo, Yoruba, Hausa, Tswana, Lingala, Luganada, Afrikaans, Somali, Swahili, Swati, North Sotho, Oromo, Wolof, Xhosa, and Zulu; \href{https://huggingface.co/datasets/castorini/afriberta-corpus}{Afriberta-Corpus} for Afaan / Oromo, Amharic, Gahuza, Hausa, Igbo, Pidgin, Somali, Swahili, Tigrinya and Yoruba; \href{https://huggingface.co/datasets/mc4}{mC4} for Afrikaans, Amharic, Hausa, Igbo, Malagasy, Chichewa, Shona, Somali, Sepedi, Swahili, Xhosa, Yoruba and Zulu. 

\section{Typology Information for AfroNLU}\label{app:typology}
\ourmodel~consists of languages from $14$ families including: Afro-Asiatic, Austronesean, Creole-English, Creole-French, Creole-Kongo, Creole-Ngbandi, Creole-Portuguese, khoe-kwadi-Hainum, khoe-kwadi-Nama khoe-kwadi-Southwest, Indo-European, Niger-Congo, and Nilo Saharan.  We discuss the classes from AfroNLU which includes Afro-Asiatic, Austronesian, Creole-English, Niger-Congo, and Nilo-Saharan. 
\subsection{Afro-Asiatic}
\begin{figure*}[ht]
  \centering
  \includegraphics[width=\linewidth]{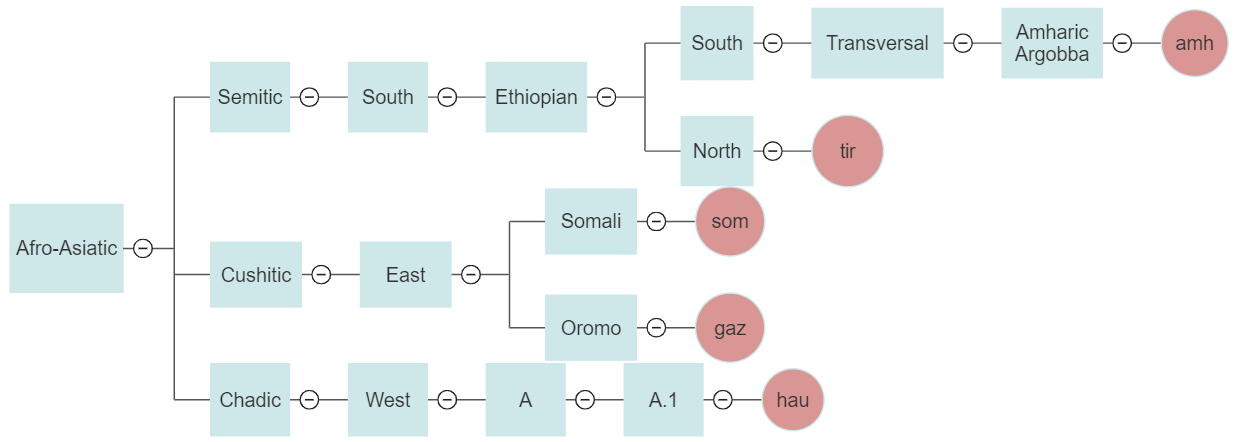}
\caption{\small Afro-Asiatic languages in \ourmodel~pretraining data. Amharic (amh), Hausa (hau), Oromo (gaz), Somali (som) and Tigrinya (tir) are presented in red circles.}\label{fig:afroasiatic} 
\end{figure*}

Afro-Asiatic (\textit{aka} Hamito-Semitic) is one of the language families of Africa. It consists of five or six branches: Berber, Chadic, Cushitic, Egyptian,  Omotic (or a single Cush-Omotic), and Semitic\cite{handbook2020_afroasiatic, bernard_classification}. Many Afro-Asiatic languages are spoken in Central, East, North, and West Africa. They are also spoken in the Middle East and in scattered communities in Europe, the United States, and the Caucasus \cite{AfroasiaticLanguages}. In Figure \ref{fig:afroasiatic}, we show relationship between the Afro-asiatic languages in AfroNLU. 

\subsection{Austronesian}
\begin{figure}[ht]
  \centering
  \includegraphics[width=\linewidth]{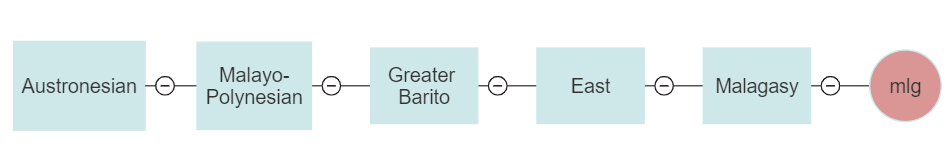}
\caption{\small Austroneasean language family consisting of Malagasy (mlg).}\label{fig:austronesean} 
\end{figure}

Austronesian languages are found along Mainland Southeast Asia, through Indonesia, Western New Guniea, and the Madagascar area in Africa \cite{ethnologue}. Many of them have been shown to exhibit an isolating word structure. This means that the words in these languages are of minimal morphological complexity~\cite{gildavid2020}. In Figure \ref{fig:austronesean}, we show the geneology for Malagasy, the only Austronesian language in our benchmark. 
\subsection{Creole}
\begin{figure}[ht]
  \centering
  \includegraphics[width=\linewidth]{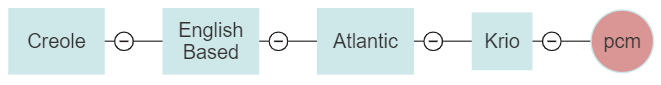}
\caption{\small \ourmodel~pretraining data has one creole language, Nigerian Pidgin, indicated with ISO-639-3 code pcm.}\label{fig:creole} 
\end{figure}

A creole language is one spoken initially only in situations of contact between speakers of two or more mutually unintelligible languages, and not as a language within an ethnic group \cite{sommer_2020}. Historically, creoles have evolved along trade routes or in colonized communities particularly when several groups of people without a common lingua franca are forced to communicate in the presence of a dominant language. Creole languages therefore often include lexical items and grammatical features from multiple contact languages. Usually, one dominant language that is also referred to as the \textit{lexifier} language contributes a majority of the vocabulary. Creole languages are classified based on their geographical location and are further grouped according to their main lexifier languages, their presumed origins, and the major languages with which they are in contact (i.e., \textit{contact} languages). Figure \ref{fig:creole} shows the geneology for Nigerian Pidgin, the only Creole in our pretraining collection.

\subsection{Indo-European}
\begin{figure}[ht]
  \centering
  \includegraphics[width=\linewidth]{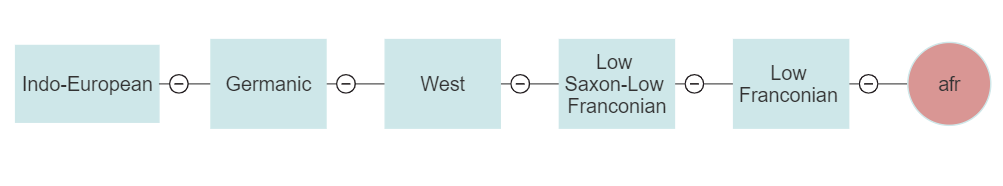}
\caption{\small Indo-European language family consisting of Afrikaans (afr).}\label{fig:indoeuropean} 
\end{figure}

Afrikaans is the only "Indigenous" Indo-European language spoken in Africa. Although it may also be viewed as not being truly Indigenous to Africa \cite{Kirsten2018}. Indo-European languages were originally domiciled in Europe, Iran, Turkey, Western Asia and India \cite{clackson_2007, ethnologue, bernard_classification, Kirsten2018}. However, due to migration, Indo-European languages are spoken around the world. In 2003, over 2.5 billion people spoke an Indo-European language \cite{clackson_2007}. In Figure \ref{fig:indoeuropean}, we show the geneology for Afrikaans.

\subsection{Niger-Congo}

\begin{figure*}[ht]
  \centering
  \includegraphics[width=\linewidth]{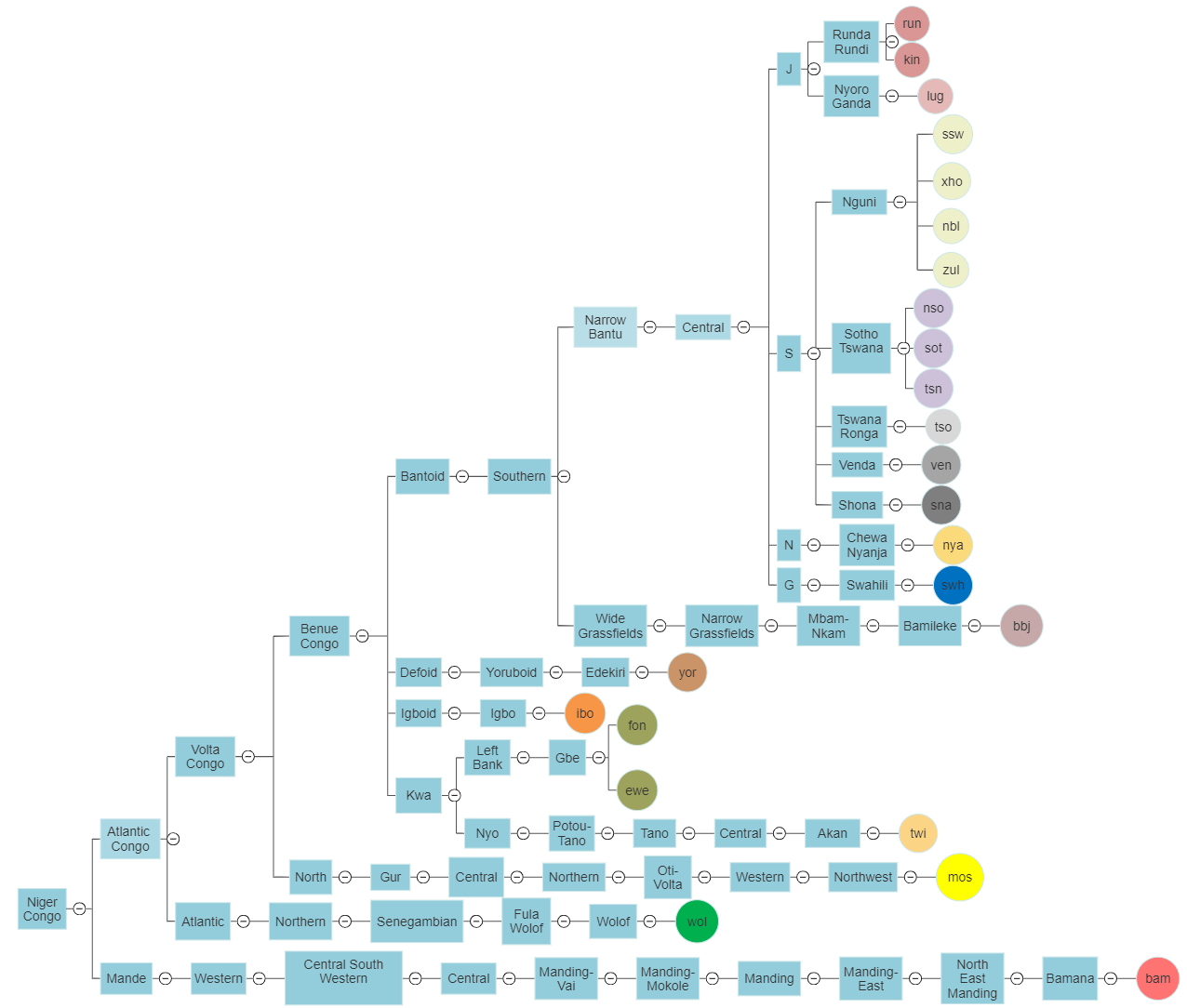}
\caption{\small Niger Congo Languages in AfroNLU benchmark. Languages which are siblings of the same parent are presented in similar colours.}\label{fig:nigercongo} 
\end{figure*}

Niger-Congo, also referred to as Niger-Kordofanian, is the largest language family in Africa \cite{jeff_2020, bernard_classification}. It consists of the highest number of languages and speakers in Africa. Niger-Congo languages spread across sub-Saharan Africa, with Benue-Congo, including Bantu languages dominating the southern part of the continent. Figure \ref{fig:nigercongo} shows the Niger-congo languages in our collection. Although we use similar colours for languages which are sisters of the same parent, only some of those languages are mutually intelligible. That is speakers of each individual language understand each other's language without learning it. Specifically, Kinyawanda (kin) and Kirundi (run) are mutually intelligible \cite{Nassenstein_2019}. Ndebele, Siswati, Xhosa, and Zulu also share various levels of intelligibility mutually intelligible \cite{Arndt_2015, roy2006state}. Sepedi, Sotho, and Tswana also share some levels of mutual intelligibility \cite{roy2006state}. 

\subsection{Nilo-Saharan}
\begin{figure*}[!]
  \centering
  \includegraphics[width=\linewidth]{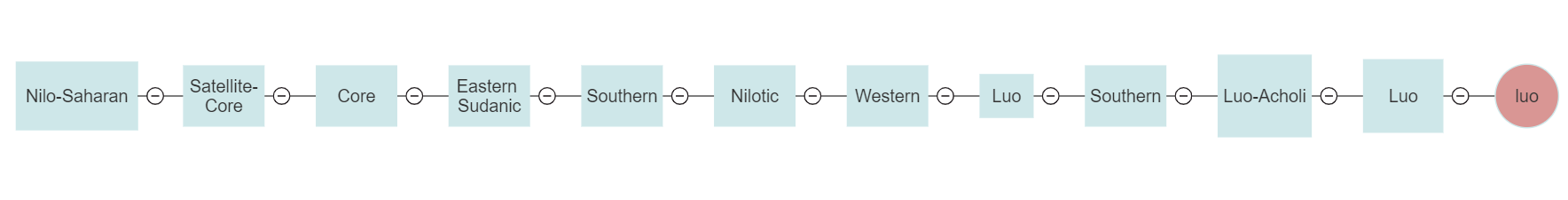}
\caption{\small Nilo Saharan language family with Luo (luo)}\label{fig:nilosaharan} 
\end{figure*}

Nilo-Saharan is subdivided into four branches that include North Eastern, Central Sudanic and two disputed branches--Songhay and Koman \cite{dimmendaal2019linguistic, handbook2020, bernard_classification}. These branches are further divided into other subgroups, languages, and dialects. Nilo-Saharan languages are spoken predominantly by eastern and central African pastoralists, and includes in its main Chari-Nile branch the Central Sudanic and Eastern Sudanic (also called Nilotic) languages. Figure \ref{fig:nilosaharan} shows the Nilo-saharan languages in our pretraining data. 

\section{Evaluation}\label{app:label}
\subsection{Performance Analysis}
\begin{figure*}[h!]
\centering
\begin{subfigure}[b]{0.49\textwidth}
\includegraphics[width=\textwidth]{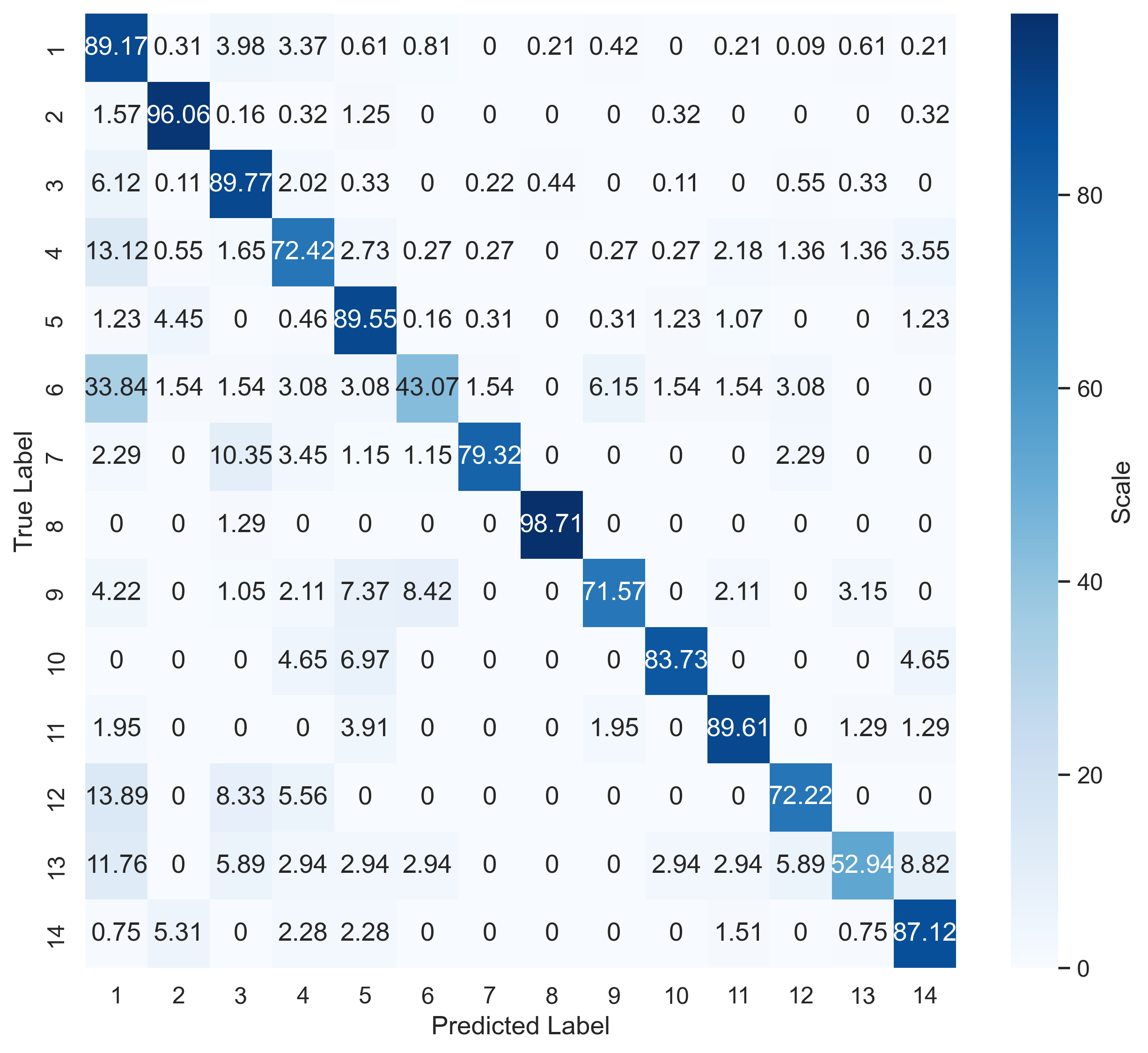}   
\caption{Kinnews }    
\end{subfigure}    
\hspace{\fill}
\begin{subfigure}[b]{0.49\textwidth}
\includegraphics[width=\textwidth]{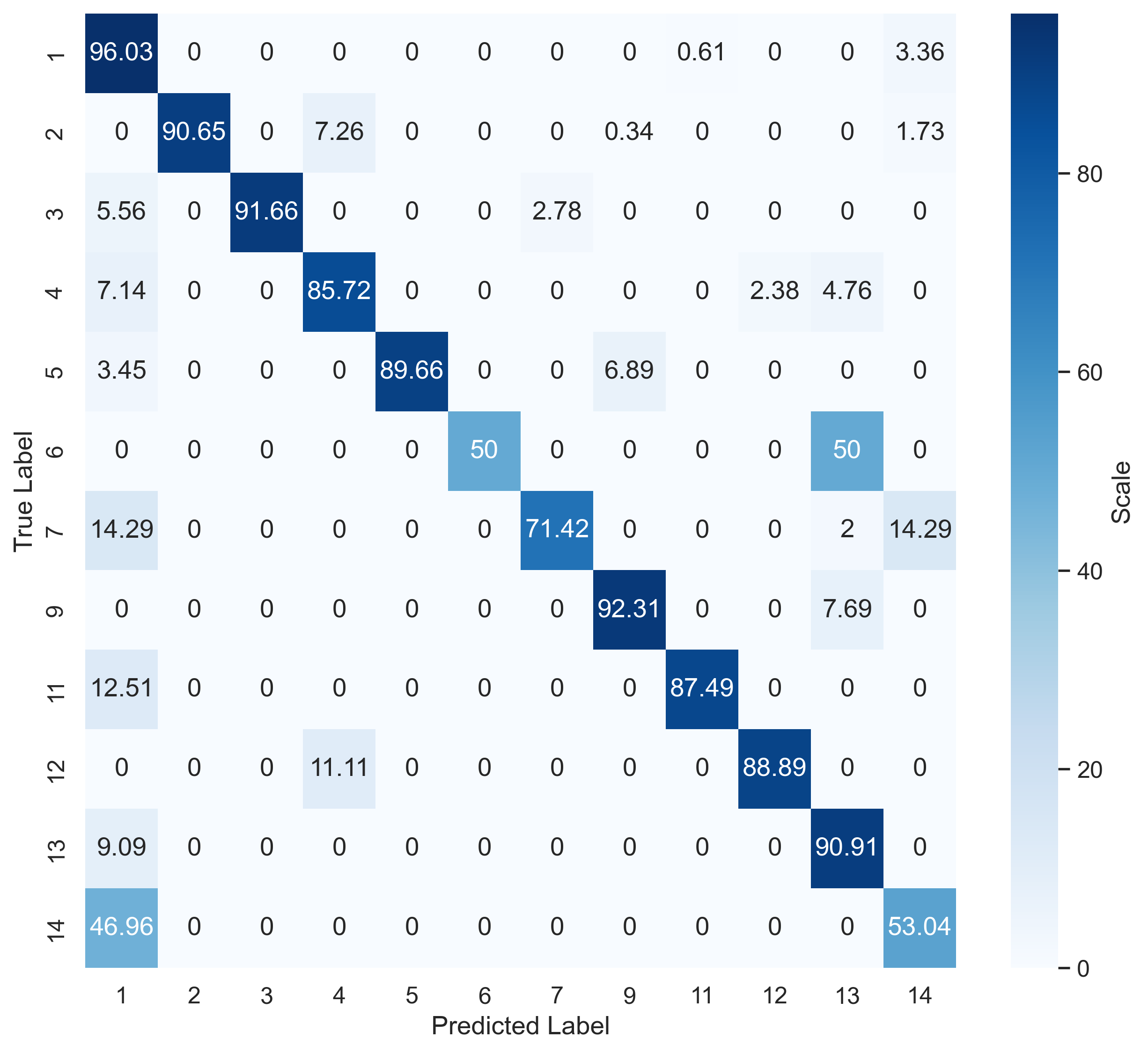}  
\caption{Kirnews}    
\end{subfigure}
\caption{Confusion matrices showing the performance of \ourmodel~for each categories in Kirnews and Kinnews classification datasets. The categories are (1) politics, (2) sports, (3) economy, (4) health, (5) entertainment, (6) history, (7) technology, (8) tourism, (9) culture, (10) fashion (11) religion, (12) environment, (13) education, and (14) relationship. Kirnews does not have Class 8 and 10.} \label{fig:classification}
\end{figure*}

\begin{figure*}[h!]
\centering
\begin{subfigure}[b]{0.3\textwidth}
\includegraphics[width=\textwidth]{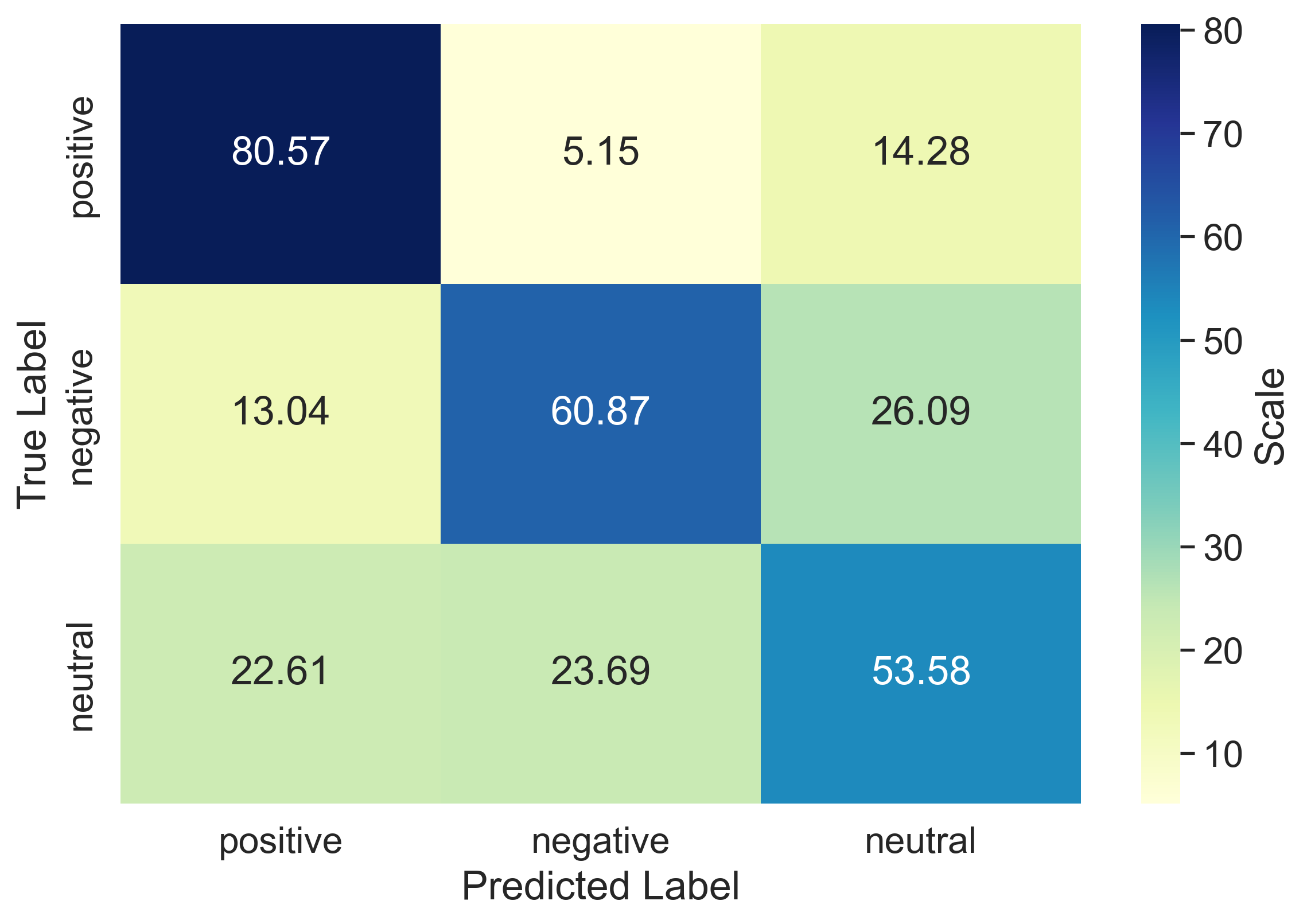}   
\caption{Bambara Sentiment Analysis}    
\end{subfigure}    
\hspace{\fill}  
\begin{subfigure}[b]{0.3\textwidth}
\includegraphics[width=\textwidth]{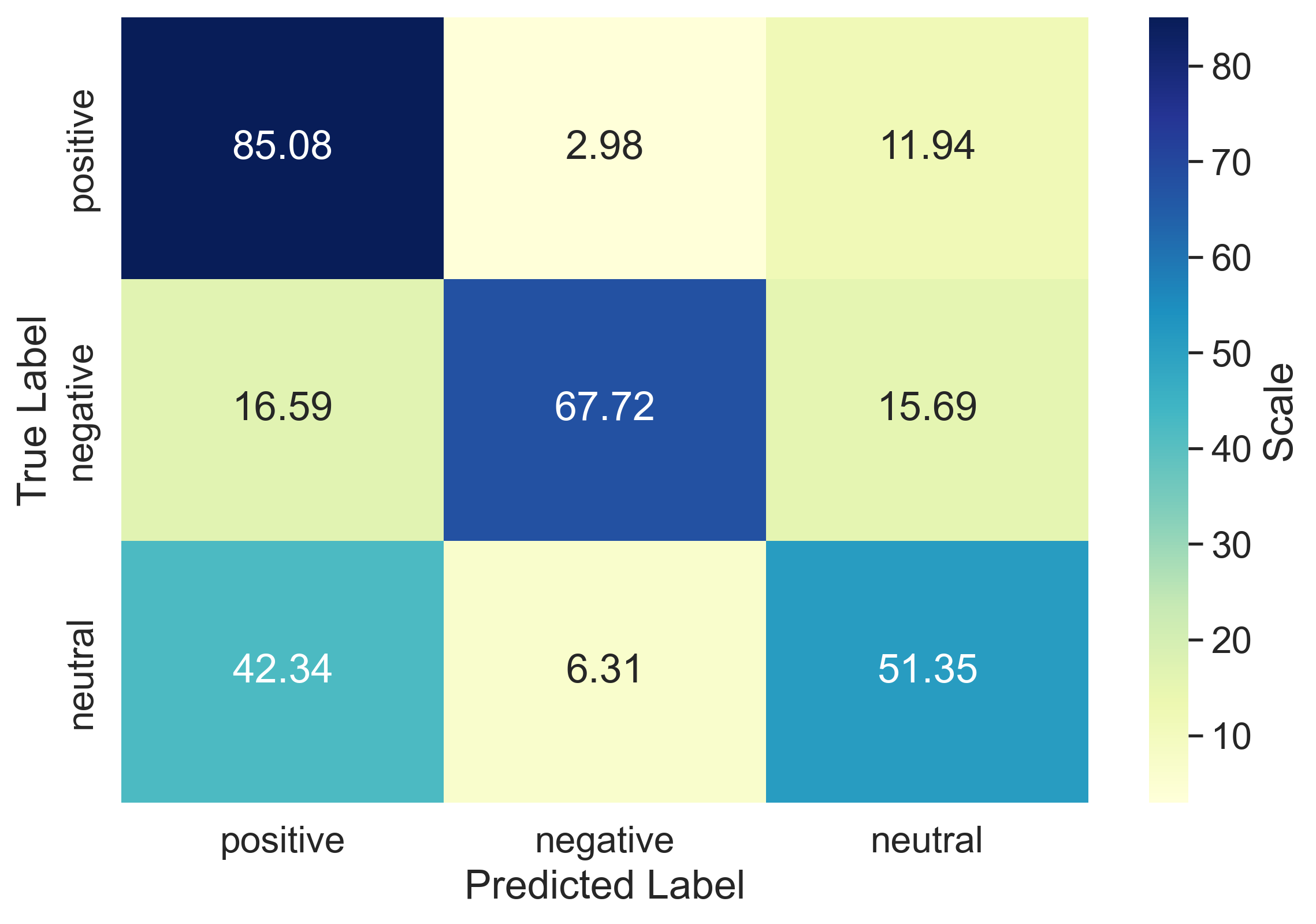}    
\caption{Pidgin Tweets}    
\end{subfigure}
\hspace{\fill}
\begin{subfigure}[b]{0.3\textwidth}
\includegraphics[width=\textwidth]{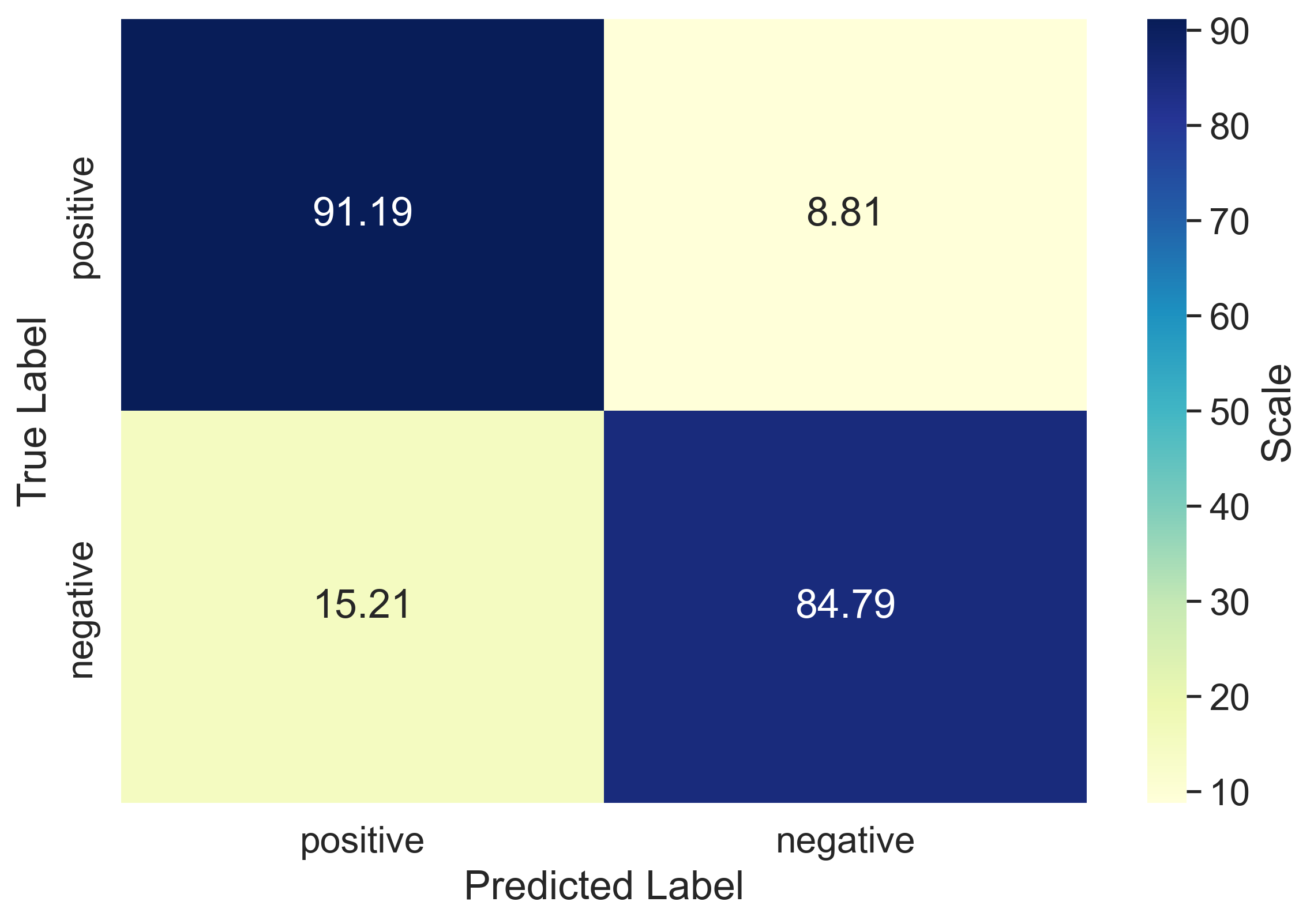}  
\caption{YOSM Sentiment}    
\end{subfigure}
\caption{Confusion matrices showing the performance of \ourmodel~for each category in Bambara, Pidgin tweets, and YOSM datasets.} \label{fig:confmatrix}
\end{figure*}

\begin{figure*}[h!]
\centering
\begin{subfigure}[b]{0.4\textwidth}
\includegraphics[width=\textwidth]{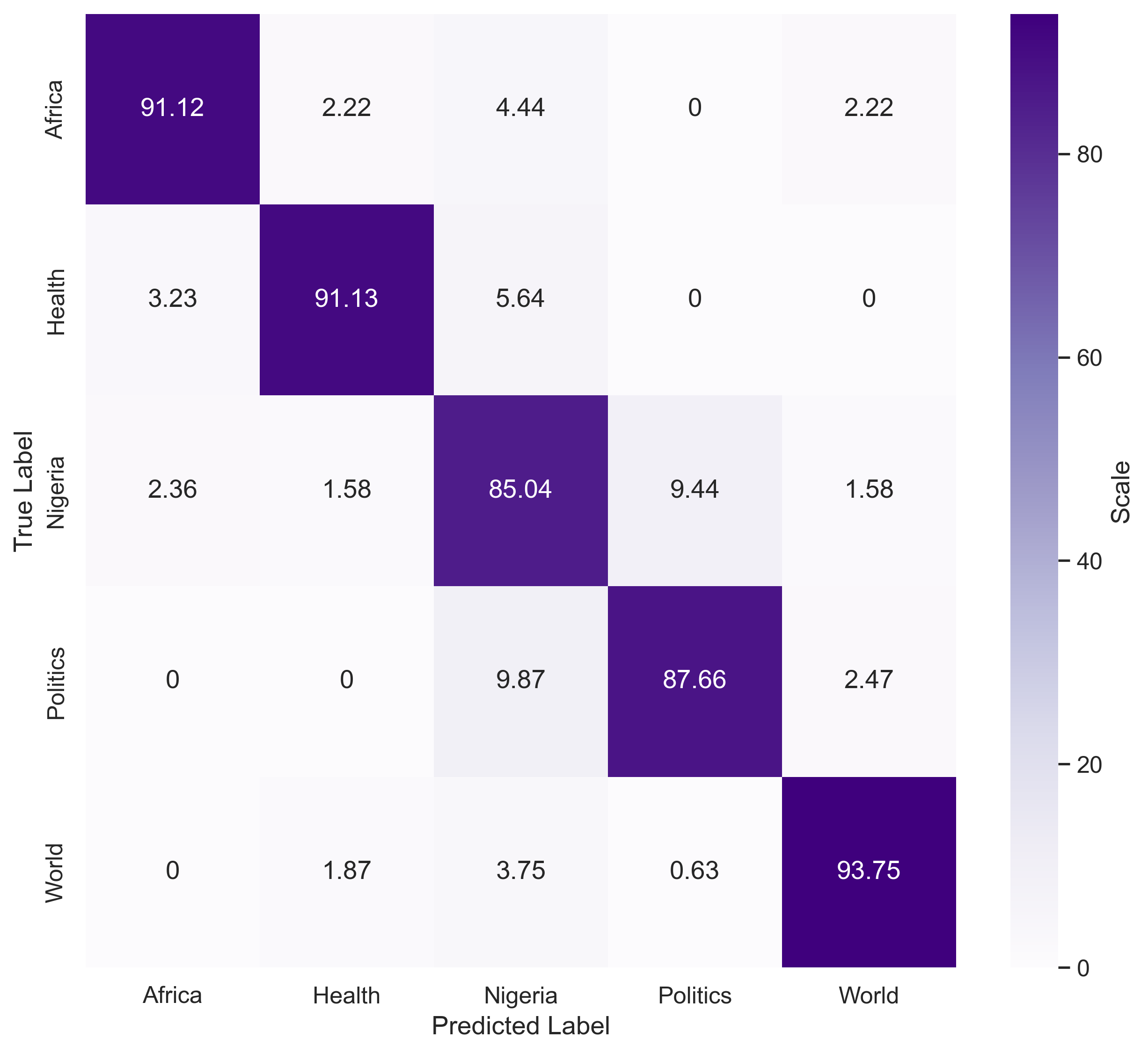}   
\caption{Hausa Topic Classification.}    
\end{subfigure}    
\hspace{\fill}
\begin{subfigure}[b]{0.4\textwidth}
\includegraphics[width=\textwidth]{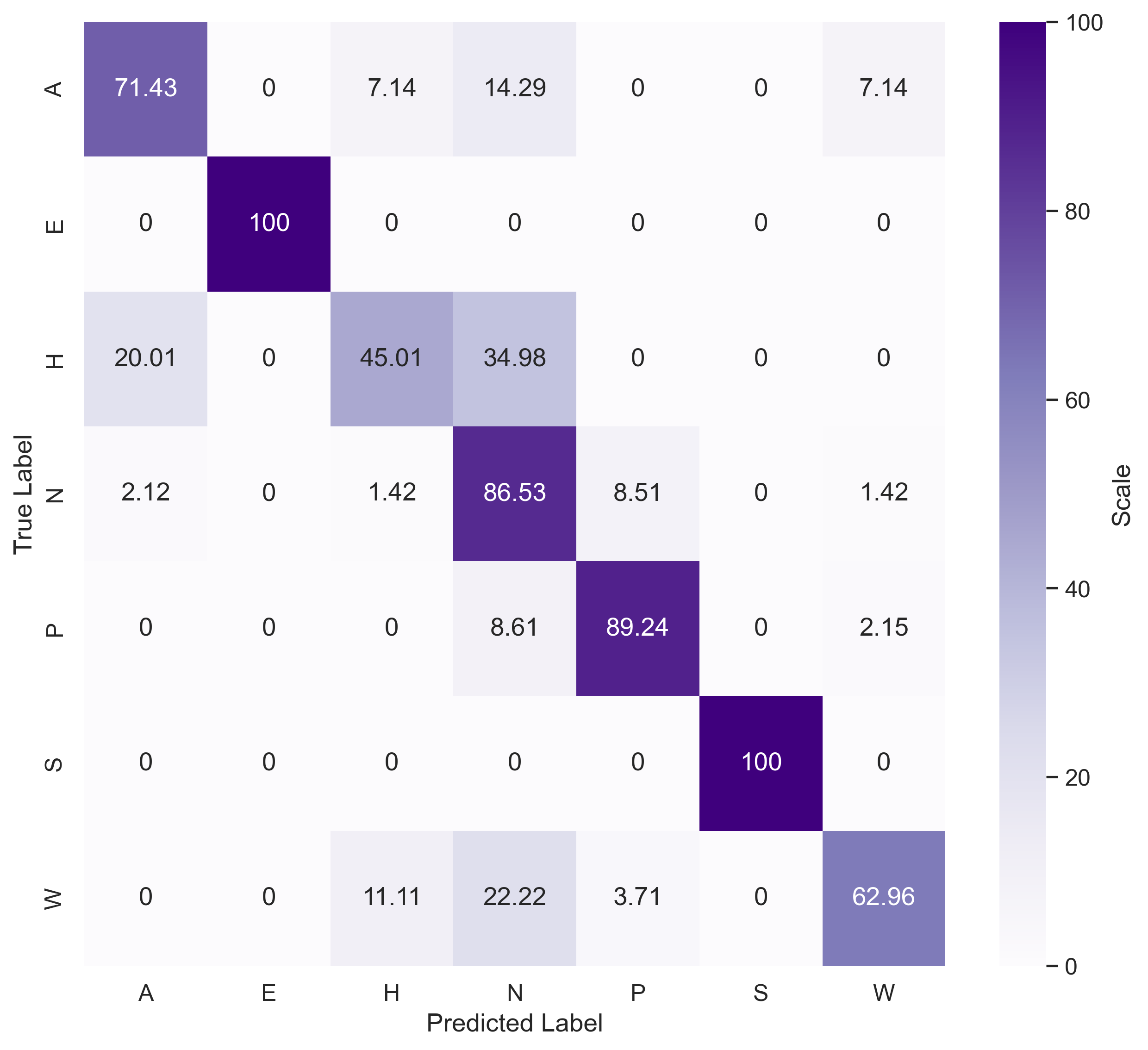}  
\caption{Yoruba Topic Classification.}    
\end{subfigure}
\caption{Confusion matrices showing the performance of \ourmodel~for each categories in Hausa and Yoruba topic classification datasets. A="Africa", E="Entertainment", H="Health", N="Nigeria", P="Politics", S="Sport", W="World"} \label{fig:topicclassification}
\end{figure*}

In this section, we provide more information about our evaluation procedure and results using visualizations and tables. Figure \ref{fig:classification} shows the confusion matrix for the news classification cluster. Figure \ref{fig:confmatrix} shows the performance of \ourmodel~on the sentiment analysis cluster. Each confusion matrix represents each dataset in the sentiment analysis cluster. In Figure \ref{fig:topicclassification}, we show \ourmodel~performance on each category in the topic classification datasets.

\begin{table*}[]
\resizebox{\textwidth}{!}{%
\begin{tabular}{llclllllll}
\toprule
\textbf{Cluster}                             & \textbf{Task}                & \textbf{SOTA} &\textbf{XLMR} & \textbf{mBERT}               & \textbf{Afro-XLMR} & \textbf{AfriBERTa} & \textbf{Serengeti-E110}               & \textbf{Serengeti-E250}      & \textbf{Serengeti}              \\
\midrule
%
\multirow{8}{*}{NER}                & masakaner-v1     & 84.8\textsuperscript{$\pm$0.3}                 & 85.59 \textsuperscript{$\pm$0.20}  & 82.82 \textsuperscript{$\pm$0.10}  & 87.79 \textsuperscript{$\pm$0.33} &85.19 \textsuperscript{$\pm$0.08} & 86.11\textsuperscript{$\pm$0.27} & 86.42 \textsuperscript{$\pm$0.26} & \textbf{88.82} \textsuperscript{$\pm$0.18} \\
                & masakaner-v2     & 85.7\textsuperscript{$\pm$0.1}\textsuperscript{$\star$}                 & 87.00 \textsuperscript{$\pm$0.12}  & 85.07\textsuperscript{$\pm$0.83}  & 87.46 \textsuperscript{$\pm$0.06} & 86.19 \textsuperscript{$\pm$0.11} & 86.51 \textsuperscript{$\pm$0.22} & 86.81 \textsuperscript{$\pm$0.24} & \textbf{88.98} \textsuperscript{$\pm$0.20} \\														
                              & masakaner-east     & --- & 83.52 \textsuperscript{$\pm$1.03}  & 82.85 \textsuperscript{$\pm$0.42} & 87.28 \textsuperscript{$\pm$0.68}  & 83.33 \textsuperscript{$\pm$0.56} & 85.64 \textsuperscript{$\pm$0.50} & 87.12 \textsuperscript{$\pm$0.62} & \textbf{88.09} \textsuperscript{$\pm$0.57} \\
                              
                                    & masakaner-eastwest & --- & 87.70 \textsuperscript{$\pm$0.30}  & 87.29 \textsuperscript{$\pm$0.33} & 89.34 \textsuperscript{$\pm$0.07} & 87.77 \textsuperscript{$\pm$0.34} & 88.14 \textsuperscript{$\pm$0.26} & 88.96\textsuperscript{$\pm$0.15} & \textbf{90.38} \textsuperscript{$\pm$0.17} \\
                                    
                                    & masakaner-west     & --- & 89.77 \textsuperscript{$\pm$0.53}  & 90.28 \textsuperscript{$\pm$0.46} & 89.97 \textsuperscript{$\pm$0.23}  & 89.36\textsuperscript{$\pm$0.46} & 88.24 \textsuperscript{$\pm$0.52} & 89.44 \textsuperscript{$\pm$0.56} & \textbf{91.58} \textsuperscript{$\pm$0.08} \\
                                    & nchlt-ner          & --- & 72.19 \textsuperscript{$\pm$0.13}  & 71.44 \textsuperscript{$\pm$0.07} & 73.22 \textsuperscript{$\pm$0.2}  & 69.25 \textsuperscript{$\pm$0.25} & 65.67 \textsuperscript{$\pm$0.07} & 65.86 \textsuperscript{$\pm$0.16} & \textbf{73.81} \textsuperscript{$\pm$0.18} \\
                                    & yoruba-twi-ner     & --- & 57.40 \textsuperscript{$\pm$2.51}   & 75.35 \textsuperscript{$\pm$0.78} & 68.02 \textsuperscript{$\pm$2.01} &\textbf{ 82.40} \textsuperscript{$\pm$0.04}  & 65.6 \textsuperscript{$\pm$2.87}  & 62.45 \textsuperscript{$\pm$1.04} & 79.68 \textsuperscript{$\pm$1.42} \\
                                    & wikiann            & --- & 84.82 \textsuperscript{$\pm$0.24}  & 84.68 \textsuperscript{$\pm$0.85} & \textbf{87.00} \textsuperscript{$\pm$1.12}    & 84.58 \textsuperscript{$\pm$0.46} & 84.21 \textsuperscript{$\pm$0.12} & 85.64 \textsuperscript{$\pm$0.36} & 86.91 \textsuperscript{$\pm$0.31} \\ \midrule
Phrase Chunking                    & phrase-chunk      & --- & 90.41 \textsuperscript{$\pm$0.10}   & 89.62 \textsuperscript{$\pm$0.24} & 91.54 \textsuperscript{$\pm$0.24} & 89.47 \textsuperscript{$\pm$0.22} & 91.99 \textsuperscript{$\pm$0.02} & 91.70 \textsuperscript{$\pm$0.27}  & \textbf{92.01} \textsuperscript{$\pm$0.18} \\\midrule
POS                                 & igbo-pos           & --- & 85.40 \textsuperscript{$\pm$0.04}   & 85.31 \textsuperscript{$\pm$0.16} & 85.23 \textsuperscript{$\pm$0.26} & 85.35 \textsuperscript{$\pm$0.07} & 85.39 \textsuperscript{$\pm$0.14} & \textbf{85.54} \textsuperscript{$\pm$0.12} & 85.36 \textsuperscript{$\pm$0.18} \\\midrule
\multirow{4}{*}{News}               & amharic-news       & --- & 85.83 \textsuperscript{$\pm$0.56}  & 60.83 \textsuperscript{$\pm$0.91} & 85.97 \textsuperscript{$\pm$0.34} & \textbf{87.03} \textsuperscript{$\pm$0.35} & 86.37 \textsuperscript{$\pm$0.42} & 86.13 \textsuperscript{$\pm$0.20}  & 86.84 \textsuperscript{$\pm$0.32} \\
                                    & kinnews            & --- & 76.5 \textsuperscript{$\pm$0.91}   & 77.98 \textsuperscript{$\pm$0.41} & 79.15 \textsuperscript{$\pm$0.57} & 78.21 \textsuperscript{$\pm$0.41} & \textbf{80.09} \textsuperscript{$\pm$0.68} & 79.54 \textsuperscript{$\pm$1.00}    & 79.32 \textsuperscript{$\pm$1.49} \\
                                    & kirnews            & --- & 53.77 \textsuperscript{$\pm$2.54}  & 66.87 \textsuperscript{$\pm$1.48} & 86.77 \textsuperscript{$\pm$1.49} & \textbf{86.72} \textsuperscript{$\pm$0.21} & 73.63 \textsuperscript{$\pm$6.66} & 83.18 \textsuperscript{$\pm$1.3}  & 85.39 \textsuperscript{$\pm$2.73} \\
                                    & swahili-news-v0.2  & --- & 88.43 \textsuperscript{$\pm$0.31}  & 85.28 \textsuperscript{$\pm$0.21} & 88.89 \textsuperscript{$\pm$0.58} & 88.76 \textsuperscript{$\pm$0.82} & 88.09 \textsuperscript{$\pm$1.02} & 86.97 \textsuperscript{$\pm$1.31} & \textbf{89.29} \textsuperscript{$\pm$0.74} \\\midrule
\multirow{3}{*}{Sentiment Analysis} & bambara-v2         & --- & 46.22 \textsuperscript{$\pm$1.94} & \textbf{65.00} \textsuperscript{$\pm$2.00}       & 62.81 \textsuperscript{$\pm$1.35} & 60.19 \textsuperscript{$\pm$1.61} & 60.50 \textsuperscript{$\pm$0.94}  & 63.90 \textsuperscript{$\pm$3.5}   & 63.17 \textsuperscript{$\pm$0.51} \\
                                    & pidgin-tweet       & --- & 69.99 \textsuperscript{$\pm$0.41}  & 69.00 \textsuperscript{$\pm$0.44}    & \textbf{71.41} \textsuperscript{$\pm$0.16} & 69.47 \textsuperscript{$\pm$0.84} & 69.98 \textsuperscript{$\pm$0.35} & 69.64 \textsuperscript{$\pm$0.23} & 68.27 \textsuperscript{$\pm$1.11} \\
                                    & yosm               & --- & 81.18 \textsuperscript{$\pm$1.63}  & 83.99 \textsuperscript{$\pm$0.49} & 85.50 \textsuperscript{$\pm$0.87}  & \textbf{87.47} \textsuperscript{$\pm$0.53} & 85.33 \textsuperscript{$\pm$0.76} & 83.00 \textsuperscript{$\pm$1.32}    & 84.83 \textsuperscript{$\pm$2.93} \\\midrule
\multirow{2}{*}{Topic}              & hausa-topic        & --- & 84.75 \textsuperscript{$\pm$1.88}  & 83.48 \textsuperscript{$\pm$1.52} & 87.83 \textsuperscript{$\pm$0.53} & 88.41 \textsuperscript{$\pm$0.49} & 87.50 \textsuperscript{$\pm$0.11}  & 88.21 \textsuperscript{$\pm$0.61} & \textbf{89.52} \textsuperscript{$\pm$1.11} \\
                                    & yoruba-topic       & --- & 64.37 \textsuperscript{$\pm$3.15}  & 82.81 \textsuperscript{$\pm$1.56} & \textbf{86.60} \textsuperscript{$\pm$1.21}  & 85.74 \textsuperscript{$\pm$2.23} & 78.11 \textsuperscript{$\pm$4.55} & 73.07 \textsuperscript{$\pm$3.38} & 83.58 \textsuperscript{$\pm$1.68} \\\midrule
\multicolumn{3}{c}{ \textbf{AfroNLU Score}  }      & 
77.77	&	79.54	&	82.96	&	80.92	&	80.03	&	80.43	&	\textbf{83.04}
\\
\bottomrule
\end{tabular}%
}
\caption{Performance of models on seven AfroNLU benchmark DEV datasets. ($F_1$) score is the evaluation metric. In QA task, we train the models on English squad TRAIN and DEV datasets. We exclude the QA from AfroNLU DEV datasets. We use a dash (-) for tasks without a known SOTA.}
\label{tab:afronlu_dev_results}
\end{table*}

\subsection{Error Analysis}
In the sentiment analysis cluster, best performance is recorded for positive categories while negative categories have the worst performance. A fine-grained analysis of the Yoruba sentiment dataset found that \ourmodel~failed to correctly categorize sentiment if the polarity item(s) were not seen in training, can be associated with both positive and negative sentiments, the polarity item(s) is a negation, or if ambivalent markers are present in the sentence. We provide a table showing examples of each type of error we found in Table \ref{tab:yor_senti_error} in the Appendix. For the news classification task, politics and tourism are the best performing classes while education and relationships have the worst performance on kirnews and kinnews respectively. It is important to mention that the worst performing categories do not have the smallest data sizes. For the topic classification, the best performance is on the world class for Hausa topic modelling while entertainment and sport have best performance for Yoruba. The worst performance is on Nigeria and health for Hausa and Yoruba topic datasets respectively. 
\begin{table*}[]
\begin{tabular}{llll}
\toprule
\multicolumn{1}{c}{\textbf{Category}}        & \multicolumn{1}{c}{\textbf{Sentence}}                                                      & \multicolumn{1}{c}{\textbf{Gold}} & \multicolumn{1}{c}{\textbf{Prediction}} \\ \toprule
\multirow{2}{*}{Ambivalence Markers}         & K\`{o} bur\'{u} \textcolor{red}{\textsubdot{s}\`{u}gb\textsubdot{\`{o}}n} \'{o} ti p\textsubdot{\'{e}}j\`{u}                                                                 & positive                          & negative                                \\ 
                                             & Sinim\'{a} t\`{i} a l\`{e} p\`{e} n\`{i} \`{i}r\`{a}w\textsubdot{\'{o}} sinim\'{a} t\`{i} \`{o} \`{n} \\ & k\textsubdot{o} m\textsubdot{\'{o}}n\`{a} m\textsubdot{\'{o}}n\`{a} \textcolor{red}{\textsubdot{s}\`{u}gb\textsubdot{\`{o}}n} n t\`{i} kò n\`{i} ohun \'{a}m\'{u}y\textsubdot{e} ni.  & negative                          & positive                                \\ \hline
\multirow{2}{*}{Negation}                    & Er\'{e} s\'{i}se naa \textcolor{red}{ko} d\'{a}ra to, ìt\`{a}n naa \textcolor{red}{k\`{o}} yeni, \\
& n\'{i} \`{e}r\`{o} t\`{e}mi \`{o}\textsubdot{s}\`{e}r\'{e} t\'{o} daa j\`{u} ni \`{i}y\'{a} n\'{a}à         & negative                          & positive                                \\
                                             & \textsubdot{S}e oun t\'{o} \textcolor{red}{o} f\textsubdot{\'{e}}.                                                            & negative                          & positive                                \\ \hline
\multirow{2}{*}{Not seen in training}      & W\textsubdot{o}n r\'{i} sinima y\`{i}\'{i} \textsubdot{s}e, \textcolor{red}{\`{a}gb\textsubdot{\'{o}}d\textsubdot{\`{o}}} w\`{o} ni                                                       & positive                          & negative                                \\
                                             & Ir\'{u} y\'{a}di f\'{i}\'{i}m\`{u}. Mo \textcolor{red}{k\'{o}r\`{i}r\'{a}} gbogbo d\'{i}d\'{a}gb\'{e} mi \\ & n\`{i}kan kej\`{i} t\`{i} o. \textcolor{red}{\`{I}d\textsubdot{o}t\'{i}} \'{n}l\'{a}! & negative                          & positive                                \\ \hline
\multirow{2}{*}{Polarity item can be either} & \textcolor{red}{\`{I}k\`{i}l\textsubdot{\`{o}}}. O n\'{i} l\'{a}ti wo \textcolor{red}{\`{i}par\'{i}} er\'{e} y\`{i}\'{i} n\'{i}tor\'{i} w\'{i}p\'{e} \'{n}kan \\ positive or negative & \textsubdot{s}\textsubdot{e}l\textsubdot{\`{e}} n\'{i} \textcolor{red}{\`{i}par\'{i}} er\'{e} n\'{a}\`{a}.      & positive                          & negative                                \\
                                             & N\`{i}kan n\'{i} \`{i}p\`{o} \textcolor{red}{\`{a}w\`{a}d\`{a}} Nollywood \textcolor{red}{gb\`{o}\`{o}r\`{o}}. \textsubdot{S}\'{e} \'{o} n\'{i} \\ & \textcolor{red}{\`{i}d\'{a}n\'{i}l\'{o}j\'{u} n\'{i}t\`{o}\'{o}t\textsubdot{\'{o}}}.      & negative                          & positive  \\ \bottomrule                            
\end{tabular}
\caption{Error analysis of Yoruba Sentiment analysis dataset. The polarity items are highlighted in red.}
\label{tab:yor_senti_error}
\end{table*}
\subsection{Imbalanced Distribution}
\begin{figure*}[h!]
\centering
\begin{subfigure}[b]{0.49\textwidth}
\includegraphics[width=\textwidth]{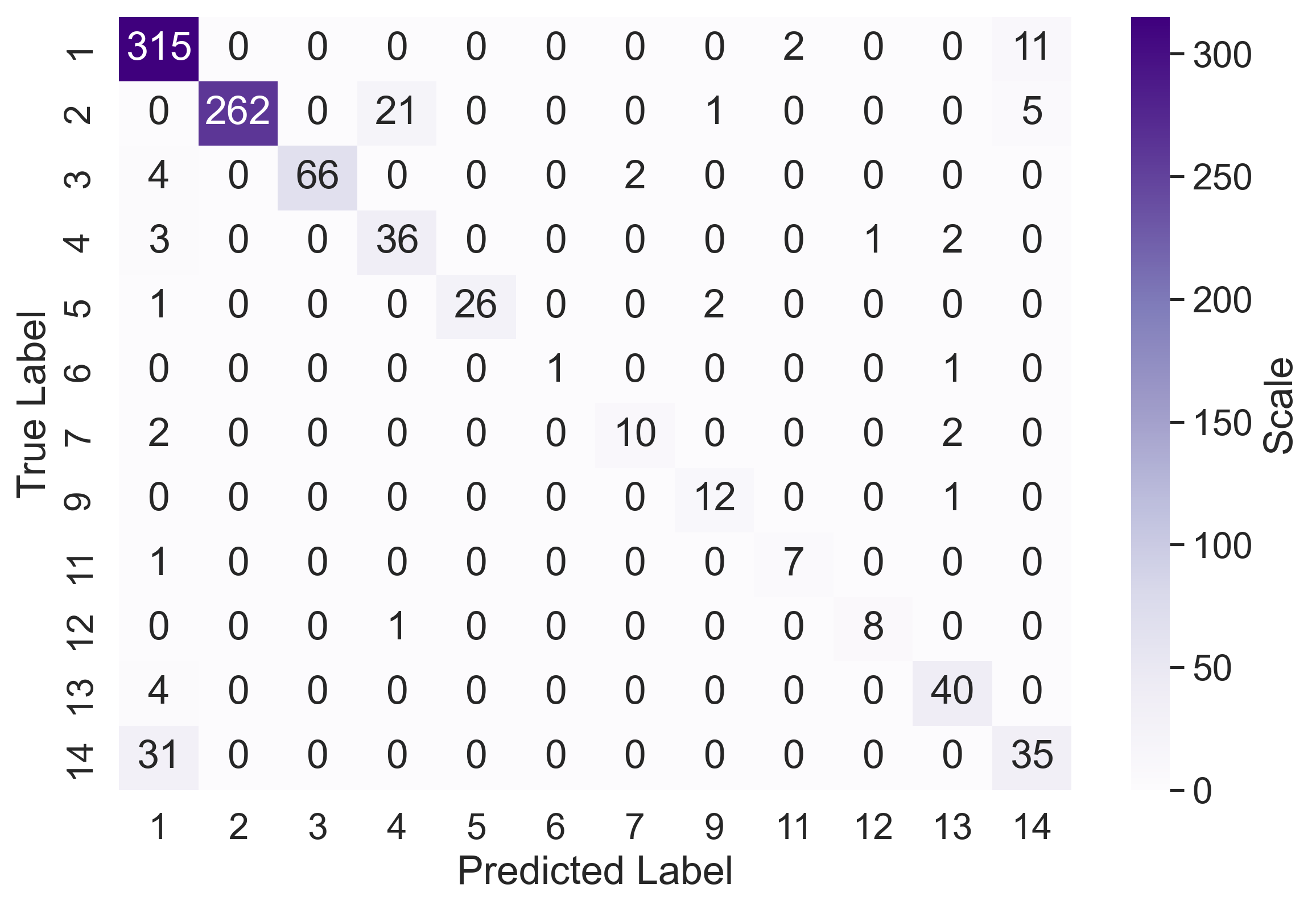}   
\caption{Kinnews }    
\end{subfigure}    
\hspace{\fill}
\begin{subfigure}[b]{0.49\textwidth}
\includegraphics[width=\textwidth]{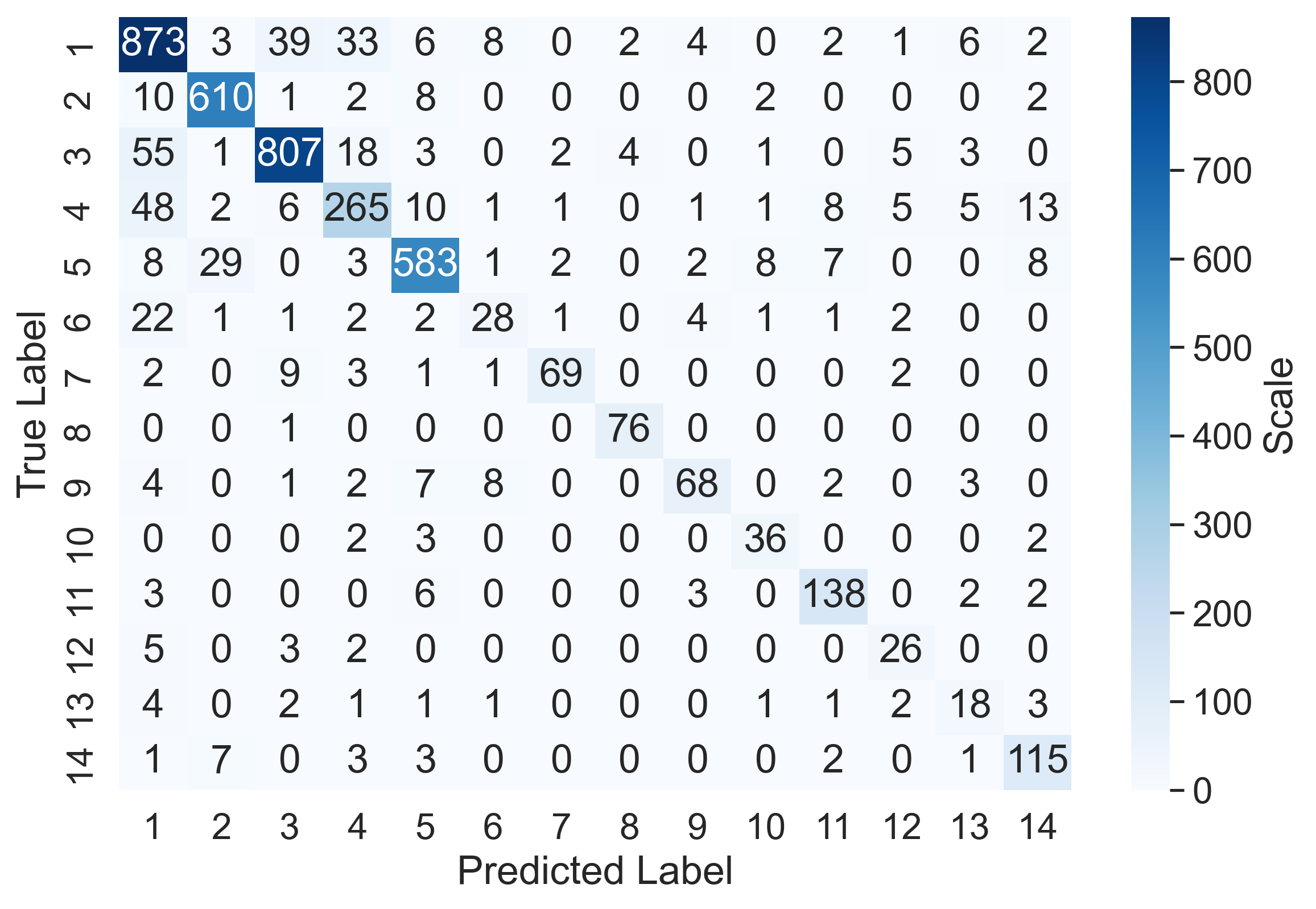}  
\caption{Kirnews}    
\end{subfigure}
\caption{Confusion matrices showing the performance of \ourmodel~for each categories in Kirnews and Kinnews classification datasets.} \label{fig:num_classification}
\end{figure*}

\begin{figure*}[h!]
\centering
\begin{subfigure}[b]{0.3\textwidth}
\includegraphics[width=\textwidth]{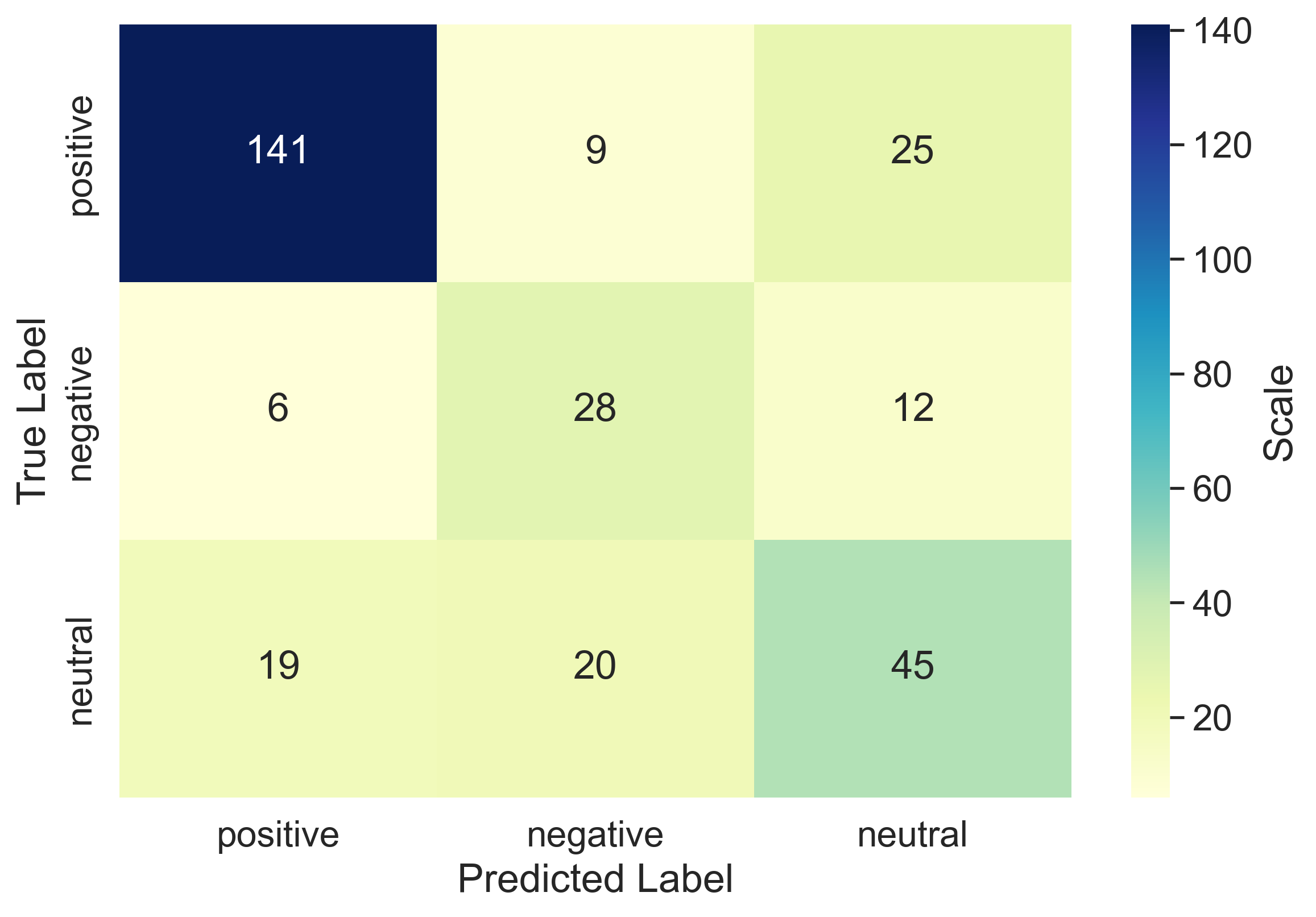}   
\caption{Bambara Sentiment Analysis}    
\end{subfigure}    
\hspace{\fill}  
\begin{subfigure}[b]{0.3\textwidth}
\includegraphics[width=\textwidth]{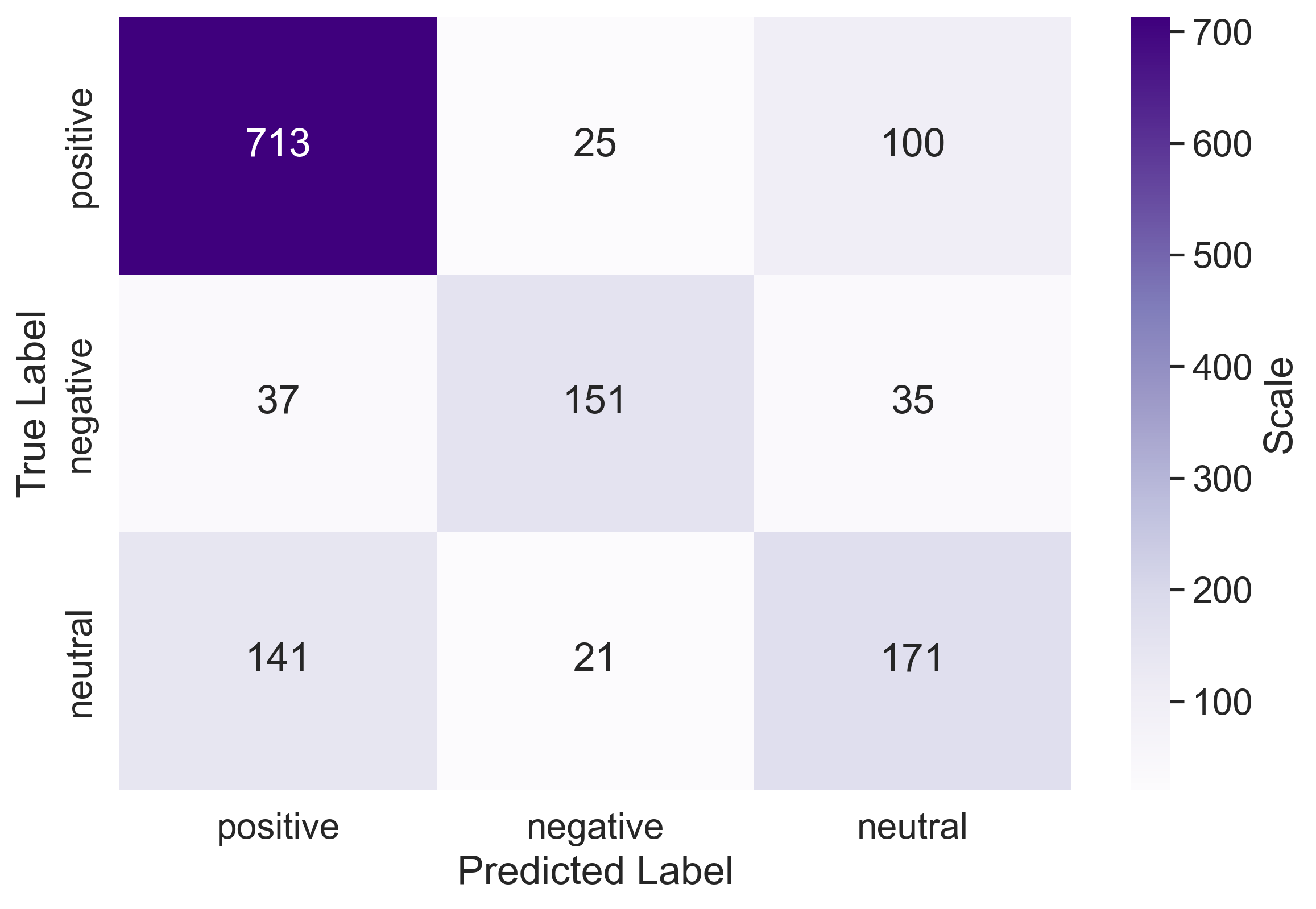}    
\caption{Pidgin Tweets}    
\end{subfigure}
\hspace{\fill}
\begin{subfigure}[b]{0.3\textwidth}
\includegraphics[width=\textwidth]{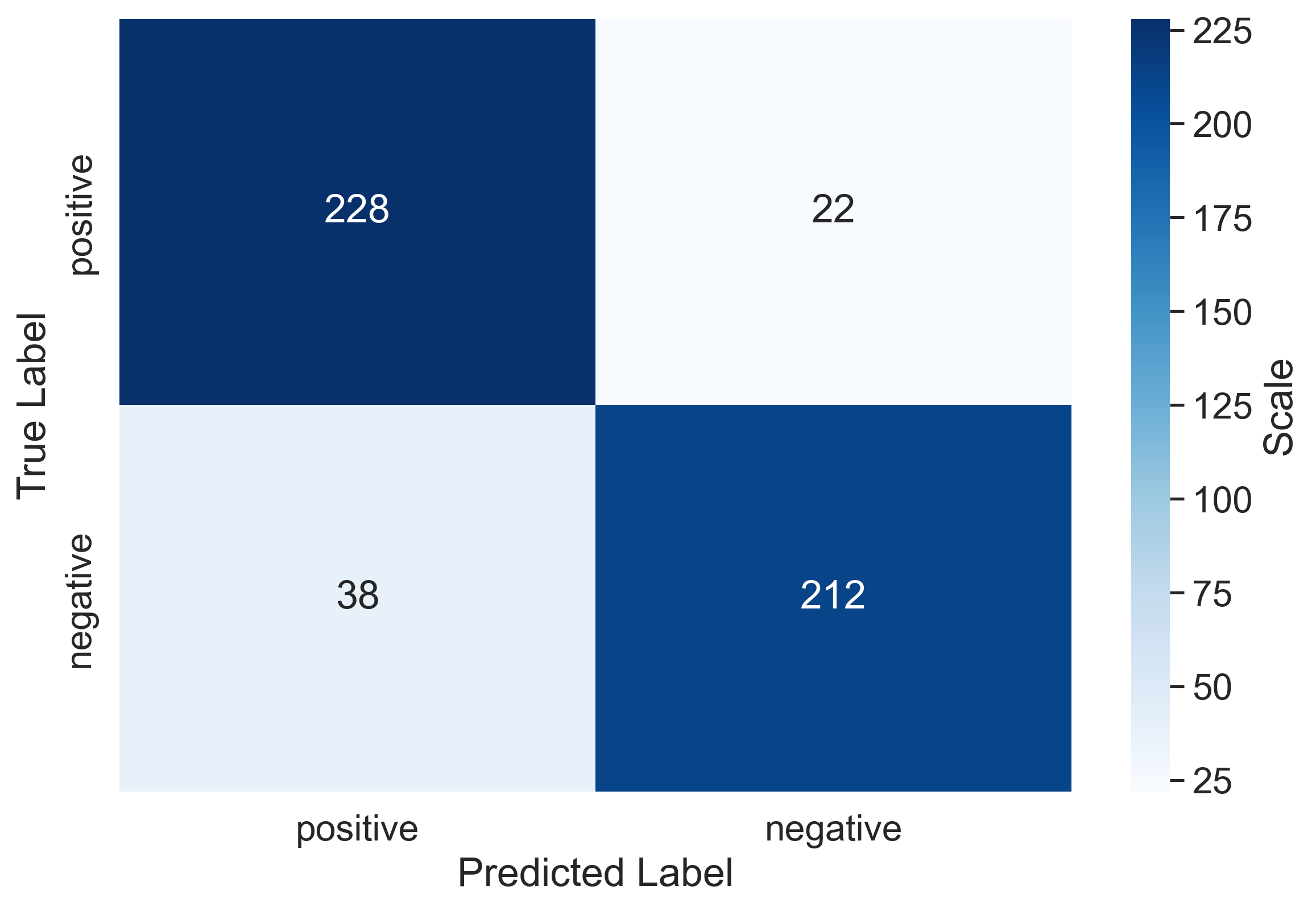}  
\caption{YOSM Sentiment}    
\end{subfigure}
\caption{Confusion matrices showing the performance of \ourmodel~for each category in Bambara, Pidgin tweets, and YOSM datasets.} \label{fig:confmatrix_sent}
\end{figure*}

\begin{figure*}[h!]
\centering
\begin{subfigure}[b]{0.49\textwidth}
\includegraphics[width=\textwidth]{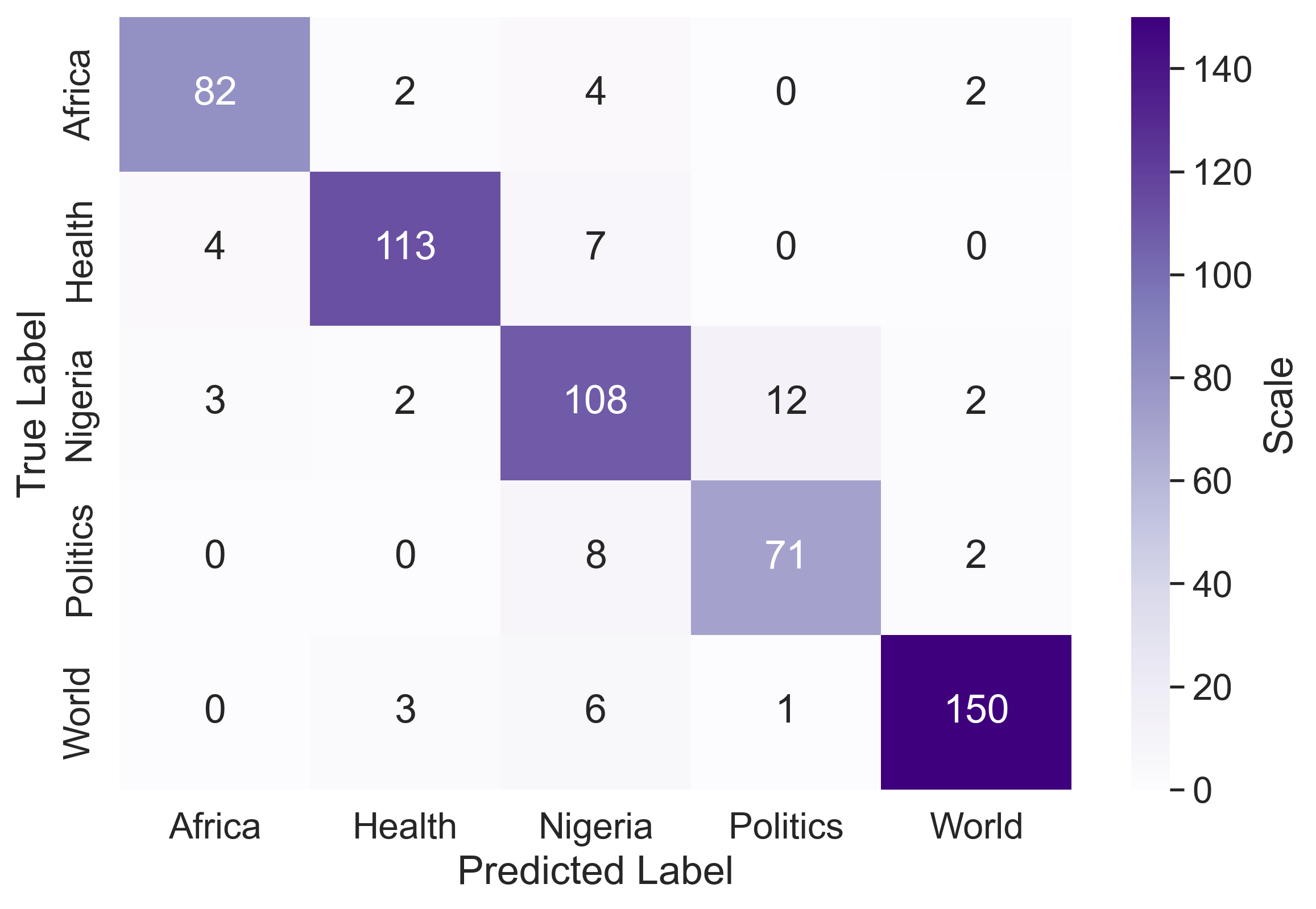}   
\caption{Hausa Topic Classification.}    
\end{subfigure}    
\hspace{\fill}
\begin{subfigure}[b]{0.49\textwidth}
\includegraphics[width=\textwidth]{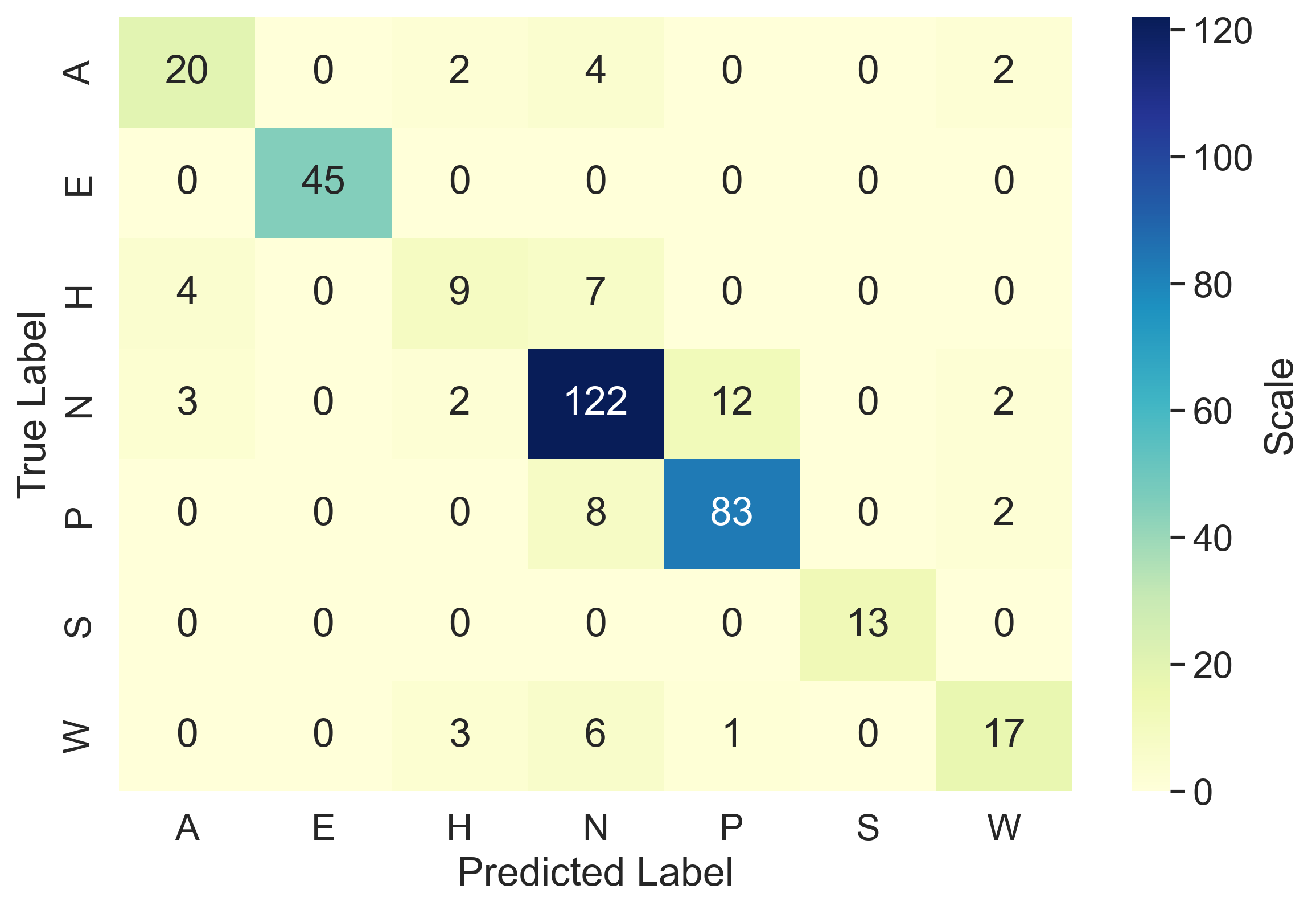}  
\caption{Yoruba Topic Classification.}    
\end{subfigure}
\caption{Confusion matrices showing the performance of \ourmodel~for each categories in Hausa and Yoruba topic classification datasets. A="Africa", E="Entertainment", H="Health", N="Nigeria", P="Politics", S="Sport", W="World"} \label{fig:topicclassification_num}
\end{figure*}

We find imbalances in the class distributions for all datasets except YOSM. We find a positive correlation between the size of each category in a dataset and the model accuracy. The larger the number of examples in a specific class, the better the accuracy, although we find a few exceptions. We provide confusion matrices that represents the sizes of each category and the performance of \ourmodel~in Figures \ref{fig:num_classification}, \ref{fig:confmatrix_sent}, and \ref{fig:topicclassification_num}.

\begin{table*}[]
\small
\centering
\resizebox{\textwidth}{!}{%
\begin{tabular}{cccccccccccc}
\toprule
\textbf{ISO-639-3} & \textbf{SERENGETI} & \textbf{AfroLID} & \textbf{Franc} & \textbf{ISO-639-3} & \textbf{SERENGETI} & \textbf{AfroLID} & \textbf{Franc} & \textbf{ISO-639-3} & \textbf{SERENGETI} & \textbf{AfroLID} & \textbf{Franc} \\ \toprule
aar            & \textbf{100.00}             & 96.00            & 74.00          & kde            & \textbf{99.00}              & 95.00            & 60.00          & pov            & \textbf{98.00}              & 93.00            & 82.00          \\
ada            & \textbf{100.00 }            & \textbf{100.00}          & 98.00          & kdh            & \textbf{100.00}             & 99.00            & 95.00          & run            & \textbf{97.00 }             & 91.00            & 68.00          \\
afr            & \textbf{100.00}             & 97.00            & 81.00          & kea            & \textbf{98.00}              & 96.07            & 0.00           & sag            & \textbf{100.00}             & \textbf{100.00}           & 30.00          \\
amh            & \textbf{99.00}              & 97.00            & 36.00          & kin            & \textbf{94.00}              & 89.00            & 47.00          & shk            & \textbf{100.00}             & \textbf{100.00 }          & 93.00          \\
bam            & \textbf{92.00}              & 70.00            & 30.00          & kmb            & \textbf{98.00}              & 94.00            & 71.00          & sna            & \textbf{98.00}              & 97.00            & 91.00          \\
bba            & \textbf{100.00}             & \textbf{100.00}           & 83.00          & kng            & \textbf{99.00}              & 98.00            & 58.00          & som            & \textbf{98.00}              & 95.00            & 89.00          \\
bci            & 97.00              & \textbf{98.00 }           & 92.00          & koo            & \textbf{96.00}              & \textbf{96.00}            & \textbf{96.00}          & sot            & 92.00              & 88.00            & \textbf{93.00}          \\
bem            & \textbf{98.00}              & 94.00            & 90.00          & kqn            & \textbf{99.00}              & 98.00            & 84.00          & ssw            & \textbf{92.00 }             & 86.00            & 68.00          \\
bfa            & \textbf{100.00}             & 99.00            & 91.00          & kqs            &\textbf{ 99.00}              & 95.00            & 73.00          & suk            & \textbf{100.00}             & 99.00            & 34.00          \\
bin            & \textbf{100.00}             & 99.00            & 97.00          & ktu            & \textbf{98.00}              & 93.00            & 19.00          & sus            & \textbf{99.00}              & \textbf{99.00}            & 96.00          \\
bum            & \textbf{98.00}              & 97.00            & 72.00          & lia            & 98.00              & 97.00            & \textbf{100.00}         & swh            &\textbf{ 95.00}              & 77.00            & 70.00          \\
cjk            & \textbf{98.00}              & 96.00            & 56.00          & lin            & 98.00              &\textbf{ 99.00}            & 98.00          & tem            & \textbf{99.00}              &\textbf{ 99.00}            & 88.00          \\
crs            & \textbf{97.00}              & 96.00            & 83.00          & lot            & \textbf{100.00 }            & 99.00            & 93.00          & tir            & \textbf{100.00}             & 99.00            & 97.00          \\
dag            & \textbf{100.00}             & \textbf{100.00}           & \textbf{100.00}         & loz            & \textbf{100.00}             & 95.00            & 92.00          & tiv            & \textbf{100.00 }            & \textbf{100.00}           & 99.00          \\
dga            & 98.00              & \textbf{100.00}           & 78.00          & lua            & 98.00              & \textbf{99.00}            & 87.00          & toi            & \textbf{98.00}              &\textbf{ 98.00}            & 80.00          \\
dip            & \textbf{98.00}              & 93.00            & 86.00          & lue            & \textbf{98.00}              & 95.00            & 68.00          & tsn            & \textbf{81.00}              & 76.00            & 33.00          \\
dyu            & 95.00              & \textbf{96.00}            & 0.00           & lug            & \textbf{96.00}              & 87.00            & 64.00          & tso            & 97.00              & \textbf{99.00}            & 94.00          \\
ewe            & 93.00              & \textbf{97.00}            & \textbf{97.00}          & lun            &\textbf{ 97.00 }             &\textbf{ 97.00 }           & 86.00          & twi            & \textbf{100.00 }            & \textbf{100.00}           & 87.00          \\
fat            & \textbf{98.00}              & \textbf{98.00}            & 94.00          & men            & 98.00              & 98.00            & \textbf{99.00}          & umb            & \textbf{100.00}             & 99.00            & 76.00          \\
fon            & \textbf{98.00}              & 97.00            & 92.00          & mfq            & 92.00              & \textbf{95.00}            & 88.00          & vai            & \textbf{100.00}             & \textbf{100.00 }          & \textbf{100.00 }        \\
fuf            & \textbf{96.00}              & 93.00            & 52.00          & mos            &\textbf{ 99.00}              & 97.00            & 90.00          & ven            & \textbf{98.00}              & 95.00            & 85.00          \\
fuv            & \textbf{95.00 }             & 94.00            & 61.00          & nba            &\textbf{ 100.00 }            & 99.00            & 61.00          & vmw            & \textbf{98.00}              & 97.00            & 95.00          \\
gaa            & \textbf{98.00}              & 95.00            & 97.00          & nbl            & \textbf{79.00}              & 74.00            & 47.00          & wol            & \textbf{87.00}              & 81.00            & 21.00          \\
gaz            & 94.00              & 94.00           & \textbf{96.00}          & ndo            & \textbf{97.00 }             & 96.00            & 76.00          & xho            & \textbf{75.00}              & 67.00            & 30.00          \\
gjn            & \textbf{100.00}                & 98.00            & 99.00          & nso            & \textbf{89.00}              & 83.00            & 59.00          & xsm            & \textbf{99.00}              &\textbf{ 99.00}            & 53.00          \\
gkp            & 68.00              & 63.00            & \textbf{69.00}          & nya            & \textbf{99.00}              & 92.00            & 75.00          & yor            & \textbf{99.00}              & 98.00            & 66.00          \\
hau            & \textbf{95.00}              & 88.00            & 77.00          & nym            & 98.00              & \textbf{99.00}            & 54.00          & zdj            & \textbf{98.00}              & 96.00            & 63.00          \\
ibb            & \textbf{99.00}              & 98.00            & 84.00          & nyn            & \textbf{95.00}              & 92.00            & 92.00          & zul            & \textbf{68.00}              & 50.00            & 40.00          \\
ibo            & \textbf{97.00 }             & \textbf{97.00}            & 88.00          & nzi            & \textbf{100.00 }            & 97.00            & 98.00          &                &                    &                  &                \\
kbp            & \textbf{100.00}             & \textbf{100.00}           & 98.00          & pcm            & \textbf{96.00}              & \textbf{96.00}            & 82.00          &                &                    &                  &               
	\\\midrule
\multicolumn{3}{l}{\textbf{SERENGETI Average f1\_score: $96.29$}}       & \multicolumn{1}{l}{} & \multicolumn{3}{l}{AfroLID Average f1\_score: $91.63$}          & \multicolumn{1}{l}{} & \multicolumn{3}{l}{Franc Average: f1\_score $74.81$}            & \multicolumn{1}{l}{}
\\ \bottomrule
\end{tabular}%
}
\caption{$F_1$-scores for \ourmodel, AfroLID, and Franc on AfroLID's dataset for $88$ languages.}
\label{tab:serengeti_franc_vs_afrolid_test}
\end{table*}
\clearpage
\begin{figure*}[h!]
 \centering
\includegraphics[width=\linewidth]{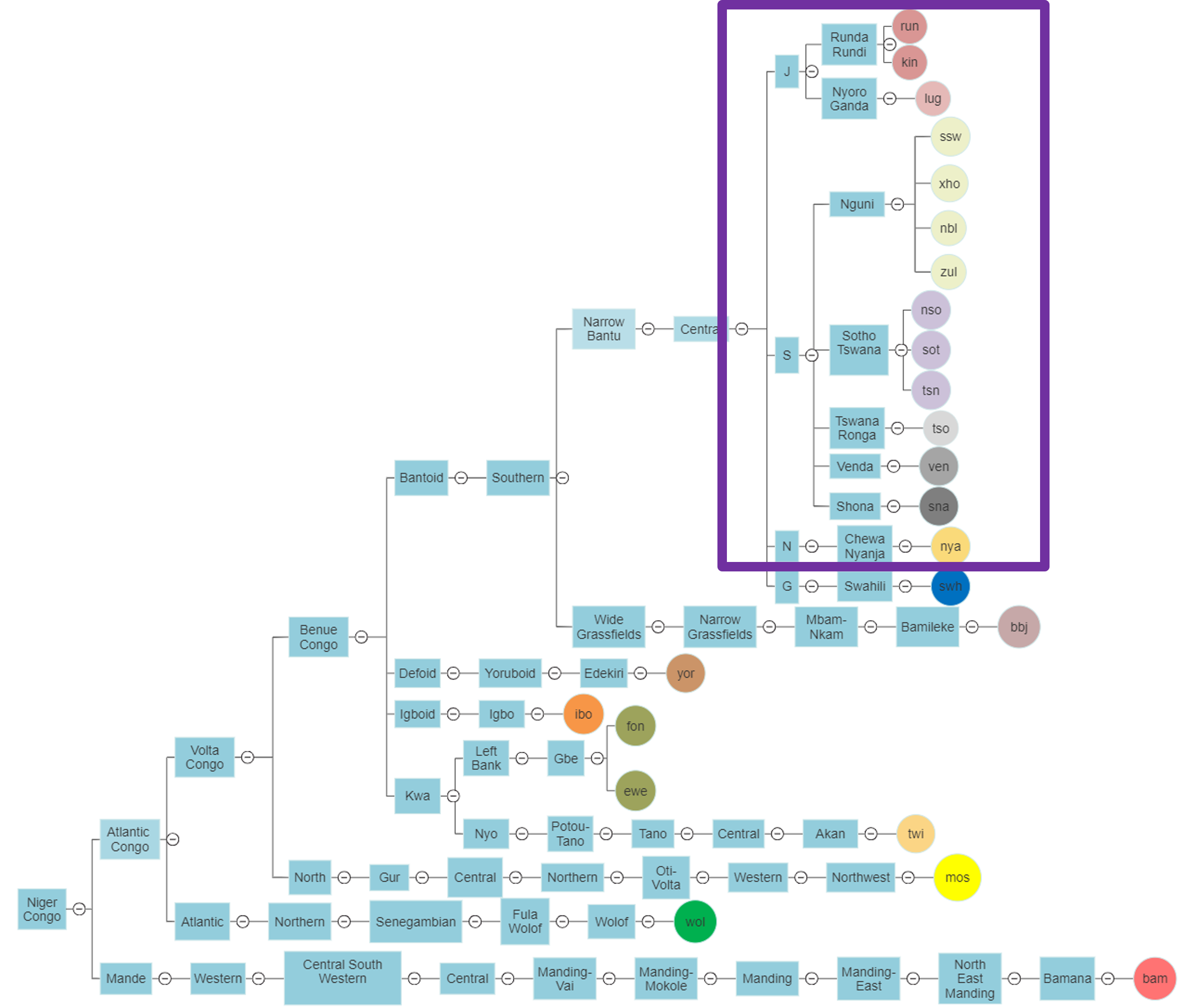}
\caption{\small A genetic classification of Niger-Congo languages in AfroNLU. We highlight in purple the list of languages relevant to our geneaology and language contact analysis. Languages which share stronger mutual intelligibility is represented in similar colours.}\label{fig:geneaology_highlight} 
\end{figure*}
\begin{table*}[h!]
\centering
\resizebox{0.8\textwidth}{!}{%
\begin{tabular}{clcccccc}
\toprule
\textbf{Cluster} &\textbf{Dataset} & \textbf{Lang.} & \textbf{XLMR} & \textbf{mBERT} & \textbf{Afro-XLMR}& \textbf{AfriBERTa} & \textbf{\ourmodel}         \\
\midrule
\multirow{44}{*}{\rotatebox[origin=c]{90}{\parbox[c]{10cm}{\centering \textbf{Named Entity Recognition (NER)}}}} &\multirow{10}{*}{masakaner-v1 }& amh & 73.98\textsuperscript{$\pm$0.64} & \colorbox{red!10}{0.0}\textsuperscript{$\pm$0.0} & \textbf{77.38}\textsuperscript{$\pm$0.47} & 69.61\textsuperscript{$\pm$0.76} & 74.26\textsuperscript{$\pm$0.54} \\
 &  &hau & 91.39\textsuperscript{$\pm$0.24} & \colorbox{red!10}{88.25}\textsuperscript{$\pm$0.42} & 91.92\textsuperscript{$\pm$0.86} & 91.12\textsuperscript{$\pm$0.37} & \textbf{92.03}\textsuperscript{$\pm$0.59} \\
 &  &ibo & \colorbox{red!10}{84.55}\textsuperscript{$\pm$0.15} & \colorbox{red!10}{84.44}\textsuperscript{$\pm$0.97} & 87.51\textsuperscript{$\pm$0.92} & 87.95\textsuperscript{$\pm$0.54} & \textbf{87.82}\textsuperscript{$\pm$0.63} \\
 &  & kin & \colorbox{red!10}{73.54}\textsuperscript{$\pm$0.35} & \colorbox{red!10}{71.02}\textsuperscript{$\pm$1.34} & 78.46\textsuperscript{$\pm$0.34} & \colorbox{red!10}{75.07}\textsuperscript{$\pm$0.51} & \textbf{78.56}\textsuperscript{$\pm$0.34} \\
 &  &lug & \colorbox{red!10}{78.65}\textsuperscript{$\pm$1.25} & \colorbox{red!10}{79.07}\textsuperscript{$\pm$2.01} & \colorbox{red!10}{82.11}\textsuperscript{$\pm$0.99} & \colorbox{red!10}{77.84}\textsuperscript{$\pm$0.4} & \textbf{84.61}\textsuperscript{$\pm$0.4} \\
 &  &luo & \colorbox{red!10}{74.28}\textsuperscript{$\pm$1.87} & \colorbox{red!10}{74.47}\textsuperscript{$\pm$0.08} & \colorbox{red!10}{75.20}\textsuperscript{$\pm$1.23} & \colorbox{red!10}{70.76}\textsuperscript{$\pm$1.57} & \textbf{77.28}\textsuperscript{$\pm$1.61} \\
 &  &pcm & \colorbox{red!10}{88.89}\textsuperscript{$\pm$0.56} & \colorbox{red!10}{88.88}\textsuperscript{$\pm$0.91} & \textbf{90.07}\textsuperscript{$\pm$0.18} & 87.65\textsuperscript{$\pm$0.43} & 89.65\textsuperscript{$\pm$0.63} \\
 &  &swa & 87.68\textsuperscript{$\pm$0.98} & 86.12\textsuperscript{$\pm$0.5} & 87.77\textsuperscript{$\pm$0.1} & 87.72\textsuperscript{$\pm$0.13} & \textbf{88.08}\textsuperscript{$\pm$0.13} \\
 &  &wol & \colorbox{red!10}{63.4}\textsuperscript{$\pm$0.68} & \colorbox{red!10}{64.25}\textsuperscript{$\pm$1.66} & \colorbox{red!10}{\textbf{68.09}}\textsuperscript{$\pm$1.65} & \colorbox{red!10}{60.9}\textsuperscript{$\pm$1.69} & 66.26\textsuperscript{$\pm$1.47} \\
 &  &yor & \colorbox{red!10}{78.97}\textsuperscript{$\pm$0.93} & 79.45\textsuperscript{$\pm$0.36} & \textbf{83.76}\textsuperscript{$\pm$0.34} & 79.89\textsuperscript{$\pm$0.89} & 83.08\textsuperscript{$\pm$1.18}
\\\cmidrule{2-8}
 &\multirow{19}{*}{masakaner-v2 } & bam & \colorbox{red!10}{80.66}\textsuperscript{$\pm$0.99} & \colorbox{red!10}{79.2}\textsuperscript{$\pm$1.43} & \colorbox{red!10}{81.04}\textsuperscript{$\pm$0.31} & \colorbox{red!10}{78.55}\textsuperscript{$\pm$0.42} & \textbf{82.11}\textsuperscript{$\pm$0.53} \\
& & bbj & \colorbox{red!10}{72.82}\textsuperscript{$\pm$1.07} & \colorbox{red!10}{62.44}\textsuperscript{$\pm$0.59} & \colorbox{red!10}{73.31}\textsuperscript{$\pm$0.74} & \colorbox{red!10}{71.97}\textsuperscript{$\pm$1.61} & \textbf{73.66}\textsuperscript{$\pm$0.87} \\
& & ewe & \colorbox{red!10}{88.54}\textsuperscript{$\pm$0.23} & \colorbox{red!10}{84.19}\textsuperscript{$\pm$1.12} & \colorbox{red!10}{89.58}\textsuperscript{$\pm$0.54} & \colorbox{red!10}{86.97}\textsuperscript{$\pm$0.4} & \textbf{89.75}\textsuperscript{$\pm$0.14} \\
& & fon & \colorbox{red!10}{82.34}\textsuperscript{$\pm$0.09} & \colorbox{red!10}{77.87}\textsuperscript{$\pm$0.47} & \colorbox{red!10}{82.62}\textsuperscript{$\pm$0.73} & \colorbox{red!10}{78.66}\textsuperscript{$\pm$0.39} & \textbf{82.86}\textsuperscript{$\pm$0.53} \\
& & hau & 86.09\textsuperscript{$\pm$0.61} & \colorbox{red!10}{82.66}\textsuperscript{$\pm$1.46} & 87.29\textsuperscript{$\pm$0.67} & 86.14\textsuperscript{$\pm$0.38} & \textbf{87.33}\textsuperscript{$\pm$0.62} \\
& & ibo & \colorbox{red!10}{89.67}\textsuperscript{$\pm$0.28} & \colorbox{red!10}{84.04}\textsuperscript{$\pm$1.09} & 91.99\textsuperscript{$\pm$0.11} & 91.56\textsuperscript{$\pm$0.36} & \textbf{92.28}\textsuperscript{$\pm$0.21} \\
& & kin & \colorbox{red!10}{84.04}\textsuperscript{$\pm$0.48} & \colorbox{red!10}{83.53}\textsuperscript{$\pm$0.81} & \textbf{86.51}\textsuperscript{$\pm$0.3} & \colorbox{red!10}{83.22}\textsuperscript{$\pm$0.25} & 86.38\textsuperscript{$\pm$0.35} \\
& & lug & \colorbox{red!10}{86.18}\textsuperscript{$\pm$0.22} & \colorbox{red!10}{85.78}\textsuperscript{$\pm$1.41} & \colorbox{red!10}{88.17}\textsuperscript{$\pm$0.56} & \colorbox{red!10}{85.32}\textsuperscript{$\pm$0.49} & \textbf{89.24}\textsuperscript{$\pm$0.37} \\
& & mos & \colorbox{red!10}{74.55}\textsuperscript{$\pm$0.65} & \colorbox{red!10}{67.75}\textsuperscript{$\pm$1.84} & \colorbox{red!10}{\textbf{75.25}}\textsuperscript{$\pm$0.71} & 69.95\textsuperscript{$\pm$0.89} & 73.74\textsuperscript{$\pm$1.62} \\
& & nya & \colorbox{red!10}{90.23}\textsuperscript{$\pm$0.14} & \colorbox{red!10}{88.6}\textsuperscript{$\pm$0.65} & \textbf{91.84}\textsuperscript{$\pm$0.23} & \colorbox{red!10}{88.83}\textsuperscript{$\pm$0.11} & 91.29\textsuperscript{$\pm$0.19} \\
& & pcm & \colorbox{red!10}{89.11}\textsuperscript{$\pm$0.1} & \colorbox{red!10}{87.90}\textsuperscript{$\pm$1.0} & \textbf{89.27}\textsuperscript{$\pm$0.4} & 87.81\textsuperscript{$\pm$0.45} & 88.77\textsuperscript{$\pm$0.37} \\
& & sna & \colorbox{red!10}{94.15}\textsuperscript{$\pm$0.19} & \colorbox{red!10}{93.06}\textsuperscript{$\pm$0.75} & 95.35\textsuperscript{$\pm$0.16} & \colorbox{red!10}{93.51}\textsuperscript{$\pm$0.32} & \textbf{95.92}\textsuperscript{$\pm$0.2} \\
& & swa & 92.37\textsuperscript{$\pm$0.05} & 91.09\textsuperscript{$\pm$0.33} & \textbf{93.06}\textsuperscript{$\pm$0.14} & 92.43\textsuperscript{$\pm$0.11} & 92.87\textsuperscript{$\pm$0.33} \\
& & tsn & \colorbox{red!10}{85.69}\textsuperscript{$\pm$0.89} & \colorbox{red!10}{85.02}\textsuperscript{$\pm$0.85} & \colorbox{red!10}{88.24}\textsuperscript{$\pm$0.26} & \colorbox{red!10}{83.58}\textsuperscript{$\pm$0.79} & \textbf{88.43}\textsuperscript{$\pm$0.1} \\
& & twi & \colorbox{red!10}{79.60}\textsuperscript{$\pm$1.45} & \colorbox{red!10}{78.05}\textsuperscript{$\pm$2.3} & \colorbox{red!10}{79.94}\textsuperscript{$\pm$1.6} & \colorbox{red!10}{75.35}\textsuperscript{$\pm$0.81} & \textbf{80.25}\textsuperscript{$\pm$1.1} \\
& & wol & \colorbox{red!10}{85.14}\textsuperscript{$\pm$0.34} & \colorbox{red!10}{83.65}\textsuperscript{$\pm$1.11} & \colorbox{red!10}{84.60}\textsuperscript{$\pm$0.4} & \colorbox{red!10}{81.68}\textsuperscript{$\pm$0.38} & \textbf{85.97}\textsuperscript{$\pm$0.43} \\
& & xho & 87.6\textsuperscript{$\pm$0.15} & \colorbox{red!10}{86.24}\textsuperscript{$\pm$1.2} & \textbf{89.59}\textsuperscript{$\pm$0.37} & \colorbox{red!10}{86.18}\textsuperscript{$\pm$0.17} & 88.76\textsuperscript{$\pm$0.76} \\
& & yor & \colorbox{red!10}{86.56}\textsuperscript{$\pm$0.36} & 83.45\textsuperscript{$\pm$1.63} & \textbf{88.91}\textsuperscript{$\pm$0.27} & 87.45\textsuperscript{$\pm$0.17} & 87.99\textsuperscript{$\pm$0.61} \\
& & zul & \colorbox{red!10}{86.32}\textsuperscript{$\pm$0.6} & \colorbox{red!10}{84.16}\textsuperscript{$\pm$1.75} & 89.75\textsuperscript{$\pm$0.16} & \colorbox{red!10}{84.9}\textsuperscript{$\pm$0.27} & \textbf{90.41}\textsuperscript{$\pm$0.24}

\\\cmidrule{2-8}
 &\multirow{11}{*}{nchlt-ner} & afr              & 80.68\textsuperscript{$\pm$0.75}  & 80.08\textsuperscript{$\pm$0.29}   & 80.55\textsuperscript{$\pm$0.11}      & \colorbox{red!10}{74.5}\textsuperscript{$\pm$0.64}        & \textbf{81.57}\textsuperscript{$\pm$0.59} \\
 &  & nbl              & \colorbox{red!10}{74.64}\textsuperscript{$\pm$0.66}  & \colorbox{red!10}{73.48}\textsuperscript{$\pm$0.18}   & 75.26\textsuperscript{$\pm$0.28}      & \colorbox{red!10}{72.28}\textsuperscript{$\pm$0.67}       & \textbf{77.13}\textsuperscript{$\pm$0.67} \\
 &  & nso              & \colorbox{red!10}{77.0}\textsuperscript{$\pm$1.23}   & \colorbox{red!10}{78.75}\textsuperscript{$\pm$0.45}   & 80.13\textsuperscript{$\pm$0.51}      & \colorbox{red!10}{75.45}\textsuperscript{$\pm$1.09}       & \textbf{80.69}\textsuperscript{$\pm$0.64} \\
 &  & sot              & \colorbox{red!10}{54.71}\textsuperscript{$\pm$1.51}  & \colorbox{red!10}{54.68}\textsuperscript{$\pm$0.49}   & 55.57\textsuperscript{$\pm$0.2}       & \colorbox{red!10}{54.09}\textsuperscript{$\pm$0.98}       & \textbf{56.26}\textsuperscript{$\pm$1.52} \\
 &  & ssw              & \colorbox{red!10}{71.75}\textsuperscript{$\pm$0.65}  & \colorbox{red!10}{71.24}\textsuperscript{$\pm$0.75}   & 72.35\textsuperscript{$\pm$1.02}      & \colorbox{red!10}{69.38}\textsuperscript{$\pm$0.58}       & \textbf{73.37}\textsuperscript{$\pm$0.82} \\
 &  & tsn              & \colorbox{red!10}{77.02}\textsuperscript{$\pm$0.22}  & \colorbox{red!10}{76.35}\textsuperscript{$\pm$0.47}   & 77.68\textsuperscript{$\pm$0.96}      & \colorbox{red!10}{73.89}\textsuperscript{$\pm$1.41}       & \textbf{79.05}\textsuperscript{$\pm$0.75} \\
 &  & tso              & \colorbox{red!10}{74.24}\textsuperscript{$\pm$0.08}  & \colorbox{red!10}{72.95}\textsuperscript{$\pm$0.67}   & \colorbox{red!10}{74.85}\textsuperscript{$\pm$0.43}      & \colorbox{red!10}{71.05}\textsuperscript{$\pm$0.9}        & \textbf{75.13}\textsuperscript{$\pm$0.31} \\
 &  & ven              & \colorbox{red!10}{64.06}\textsuperscript{$\pm$0.31}  & \colorbox{red!10}{63.11}\textsuperscript{$\pm$1.27}   & \colorbox{red!10}{64.39}\textsuperscript{$\pm$0.36}      & \colorbox{red!10}{63.24}\textsuperscript{$\pm$1.26}       & \textbf{65.42}\textsuperscript{$\pm$0.76} \\
 &  & xho              & 70.77\textsuperscript{$\pm$2.45}  & \colorbox{red!10}{68.54}\textsuperscript{$\pm$1.44}   & 72.37\textsuperscript{$\pm$0.39}      & \colorbox{red!10}{67.00}\textsuperscript{$\pm$1.27}        & \textbf{72.92}\textsuperscript{$\pm$0.29} \\
 &  & zul              & \colorbox{red!10}{69.44}\textsuperscript{$\pm$0.62}  & \colorbox{red!10}{67.74}\textsuperscript{$\pm$1.46}   & 70.28\textsuperscript{$\pm$0.49}      & \colorbox{red!10}{67.17}\textsuperscript{$\pm$0.15}       & \textbf{71.20}\textsuperscript{$\pm$0.44}  \\
 \cmidrule{2-8}
& \multirow{6}{*}{Wikiann}          & amh              & 57.76\textsuperscript{$\pm$0.45}        & \colorbox{red!10}{33.96}\textsuperscript{$\pm$1.83}         & 64.27\textsuperscript{$\pm$1.91}            & 60.16\textsuperscript{$\pm$2.83}             & \textbf{68.11}\textsuperscript{$\pm$1.75} \\
       &  & ibo              & \colorbox{red!10}{73.6}\textsuperscript{$\pm$1.32}         & \colorbox{red!10}{70.83}\textsuperscript{$\pm$1.86}         & 73.93\textsuperscript{$\pm$1.12}            & \textbf{76.14}\textsuperscript{$\pm$1.42}             & 75.73\textsuperscript{$\pm$2.78} \\
       &  & kin              & \colorbox{red!10}{69.67}\textsuperscript{$\pm$2.07}        & \colorbox{red!10}{77.35}\textsuperscript{$\pm$4.47}         & \textbf{82.24}\textsuperscript{$\pm$2.17}            & \colorbox{red!10}{79.8}\textsuperscript{$\pm$1.06}              & 79.78\textsuperscript{$\pm$1.78} \\
       &  & swh              & 88.09\textsuperscript{$\pm$0.32}        & 88.00\textsuperscript{$\pm$0.28}          & 88.83\textsuperscript{$\pm$0.47}            & 86.13\textsuperscript{$\pm$0.2}              & \textbf{89.16}\textsuperscript{$\pm$0.35} \\
       &  & yor              & \colorbox{red!10}{83.8}\textsuperscript{$\pm$2.06}         & 81.96\textsuperscript{$\pm$0.88}         & \textbf{87.96}\textsuperscript{$\pm$1.24}            & 82.77\textsuperscript{$\pm$0.23}             & 85.00\textsuperscript{$\pm$2.42} \\ 
                                  \midrule
                                  
\multirow{9}{*}{\rotatebox[origin=c]{90}{\centering \textbf{Phrase Chunking}}} & \multirow{9}{*}{phrase-chunk} & afr              & 95.34\textsuperscript{$\pm$0.16}        & 95.68\textsuperscript{$\pm$0.30}          & 95.13\textsuperscript{$\pm$0.06}            & \colorbox{red!10}{90.22}\textsuperscript{$\pm$0.81}             &\textbf{ 96.01}\textsuperscript{$\pm$0.14}\\

       &           & nso              & \colorbox{red!10}{96.57}\textsuperscript{$\pm$0.61}        & \colorbox{red!10}{96.85}\textsuperscript{$\pm$0.55}         & \textbf{98.36}\textsuperscript{$\pm$0.2}             & \colorbox{red!10}{96.47}\textsuperscript{$\pm$0.14}             & 98.28\textsuperscript{$\pm$0.1} \\
      &            & sot              & \colorbox{red!10}{82.93}\textsuperscript{$\pm$0.38}        & \colorbox{red!10}{83.08}\textsuperscript{$\pm$0.78}         & 85.28\textsuperscript{$\pm$0.61}            & \colorbox{red!10}{82.18}\textsuperscript{$\pm$0.93}             & \textbf{85.69}\textsuperscript{$\pm$0.76}\\
     &             & ssw              & \colorbox{red!10}{82.9}\textsuperscript{$\pm$1.03}         & \colorbox{red!10}{81.91}\textsuperscript{$\pm$0.47}         & \textbf{84.73}\textsuperscript{$\pm$0.18}            & \colorbox{red!10}{83.24}\textsuperscript{$\pm$0.11}             & 83.45\textsuperscript{$\pm$0.12}\\
    &              & tsn              & \colorbox{red!10}{92.77}\textsuperscript{$\pm$0.16}        & \colorbox{red!10}{92.64}\textsuperscript{$\pm$0.66}         & 94.11\textsuperscript{$\pm$0.49}            & \colorbox{red!10}{92.71}\textsuperscript{$\pm$0.42}             & \textbf{94.03}\textsuperscript{$\pm$0.19}\\
     &             & tso              & \colorbox{red!10}{86.42}\textsuperscript{$\pm$0.46}        & \colorbox{red!10}{86.90}\textsuperscript{$\pm$0.31}          & \colorbox{red!10}{87.39}\textsuperscript{$\pm$0.18}            & \colorbox{red!10}{86.73}\textsuperscript{$\pm$0.95}             & \textbf{89.32}\textsuperscript{$\pm$0.43}\\
      &            & ven              & \colorbox{red!10}{92.31}\textsuperscript{$\pm$0.45}        & \colorbox{red!10}{90.47}\textsuperscript{$\pm$0.32}         & \colorbox{red!10}{92.42}\textsuperscript{$\pm$0.68}            & \colorbox{red!10}{92.02}\textsuperscript{$\pm$0.33}             & \textbf{92.54}\textsuperscript{$\pm$0.21}\\
      &            & zul              & \colorbox{red!10}{87.30}\textsuperscript{$\pm$0.26}         & \colorbox{red!10}{87.29}\textsuperscript{$\pm$1.04}         & 88.67\textsuperscript{$\pm$0.66}            & \colorbox{red!10}{85.74}\textsuperscript{$\pm$0.55}             & \textbf{90.05}\textsuperscript{$\pm$0.81}\\

\bottomrule
\end{tabular}%
}
\caption{Performance of mPLMs on each language in each task. ($F_1$) score is the evaluation metric. We use \colorbox{red!10}{\textbf{Red}} highlights to indicate languages in zero-shot setting.}
\label{tab:langs_test_results}
\end{table*}

\section{Detailed Geneaology and Language Contact Analysis}\label{app:lang_contact}
In this Section, we use Figures and Tables to provide evidence for the influence of similar languages in zero-shot settings. First, we highlight in purple the similar languages that we perform genealogy analysis on in Figure \ref{fig:geneaology_highlight}. In the figure, the languages with mutual intelligibility are presented in similar coloured circles. To determine the significance of language similarity and language contact in our own zero-shot settings, we measure the Jaccard similarity between the pretraining data for the South African languages in AfroNLU (see Table \ref{tab:sa_jaccard}). To calculate the Jaccard similarities, we removed digits, emojis, and punctuation marks. We do this to ensure that we reduce interference with the similarity scores. We find strong similarities between some of these languages as in the bolded examples in Table \ref{tab:sa_jaccard}. 

We find that although XLM-R, mBERT, and AfriBERTa are not trained on most most of these languages, we record high scores in zero-shot settings see Table~\ref{tab:langs_test_results}). We argue that XLM-R in addition to cross-lingual transfers from other languages acquires representation from afr and xho where xho alone shares more than 0.4 similarity with afr, nbl, nso, and zul. mBERT also learns representation from afr while AfriBERTa learns representations from Gahuza which is a code-mixed variety of kin and run. \ourmodel~however, outperforms other models on these datasets indicating that learning the representation of each language improves performance. 

Next, we finetune a BERT model and compare the performance of BERT with MBERT. We do this because BERT is a monolingual model and does not include any similar language in its representation. In Table \ref{tab:genealogy_analysis}, BERT significantly performs lower than MBERT in all languages in NCHLT-NER. BERT also has lower performance on the phrase-chunk dataset in all languages except on ssw, and ven. 

This analysis is far from being conclusive and future work can further probe the influence of similar languages in more detail. This is necessary to evaluate to what extent similar languages have an influence on performance in zero-shot settings and why in zero shot settings, some monolingual models outperform multilingual ones. For example, in the case of ssw and ven. 
\begin{table*}[h!]
\small
\centering
\resizebox{\textwidth}{!}{%
\begin{tabular}{ll ll ll ll }
\toprule
\textbf{ISO-639-3} & \textbf{Language}                & \textbf{ISO-639-3} & \textbf{Language}             & \textbf{ISO-639-3} & \textbf{Language}                       & \textbf{ISO-639-3} & \textbf{Language}         \\
\midrule
aar  & Afar / Qafar            & bky   & Bokyi                & dow   & Doyayo                         & gol   & Gola             \\
aba   & Abe / Abbey             & bmo   & Bambalang            & dsh   & Daasanach                      & gqr   & Gor              \\
abn   & Abua                    & bmv   & Bum                  & dua   & Douala                         & gso   & Gbaya, Southwest \\
acd   & Gikyode                 & bom   & Berom                & dug   & Chiduruma                      & gud   & Dida, Yocoboue   \\
ach   & Acholi                  & bov   & Tuwuli               & dwr   & Dawro                          & gur   & Farefare         \\
ada   & Dangme                  & box   & Bwamu / Buamu        & dyi   & Sénoufo, Djimini               & guw   & Gun              \\
adh   & Jopadhola / Adhola      & bqc   & Boko                 & dyu   & Jula                           & gux   & Gourmanchema     \\
adj   & Adjukru  / Adioukrou    & bqj   & Bandial              & ebr   & Ebrie                          & guz   & Ekegusii         \\
afr   & Afrikaans               & bsc   & Oniyan               & ebu   & Kiembu / Embu                  & gvl   & Gulay            \\
agq   & Aghem                   & bsp   & Baga Sitemu          & efi   & Efik                           & gwr   & Gwere            \\
aha   & Ahanta                  & bss   & Akoose               & ego   & Eggon                          & gya   & Gbaya, Northwest \\
ajg   & Aja                     & bst   & Basketo              & eka   & Ekajuk                         & hag   & Hanga            \\
akp   & Siwu                    & bud   & Ntcham               & eko   & Koti                           & har   & Harari           \\
alz   & Alur                    & bum   & Bulu                 & eto   & Eton                           & hau   & Hausa            \\
amh   & Amharic                 & bun   & Sherbro              & etu   & Ejagham                        & hay   & Haya             \\
ann   & Obolo                   & bus   & Bokobaru             & etx   & Iten / Eten                    & hbb   & Nya huba         \\
anu   & Anyuak / Anuak          & buy   & Bullom So            & ewe   & Ewe                            & heh   & Hehe             \\
anv   & Denya                   & bwr   & Bura Pabir           & ewo   & Ewondo                         & her   & Herero           \\
asa   & Asu                     & bwu   & Buli                 & fak   & Fang                           & hgm   & Haillom          \\
asg   & Cishingini              & bxk   & Bukusu               & fat   & Fante                          & hna   & Mina             \\
atg   & Ivbie North-Okpela-Arhe & byf   & Bete                 & ffm   & Fulfulde, Maasina              & ibb   & Ibibio           \\
ati   & Attie                   & byv   & Medumba              & fia   & Nobiin                         & ibo   & Igbo             \\
avn   & Avatime                 & bza   & Bandi                & fip   & Fipa                           & idu   & Idoma            \\
avu   & Avokaya                 & bzw   & Basa                 & flr   & Fuliiru                        & igb   & Ebira            \\
azo   & Awing                   & cce   & Chopi                & fon   & Fon                            & ige   & Igede            \\
bam   & Bambara                 & chw   & Chuabo               & fub   & Fulfulde, Adamawa              & igl   & Igala            \\
bav   & Vengo                   & cjk   & Chokwe               & fue   & Fulfulde, Borgu                & ijn   & Kalabari         \\
bba   & Baatonum                & cko   & Anufo                & fuf   & Pular                          & ikk   & Ika              \\
bbj   & Ghomala                & cme   & Cerma                & fuh   & Fulfulde, Western Niger        & ikw   & Ikwere           \\
bbk   & Babanki                 & cop   & Coptic               & ful   & Fulah                          & iqw   & Ikwo             \\
bci   & Baoule                  & cou   & Wamey                & fuq   & Fulfulde Central Eastern Niger & iri   & Rigwe            \\
bcn   & Bali                    & crs   & Seychelles Creole    & fuv   & Fulfude Nigeria                & ish   & Esan             \\
bcw   & Bana                    & csk   & Jola Kasa            & gaa   & Ga                             & iso   & Isoko            \\
bcy   & Bacama                  & cwe   & Kwere                & gax   & Oromo, Borana-Arsi-Guji        & iyx   & yaka             \\
bdh   & Baka                    & daa   & Dangaleat            & gaz   & Oromo, West Central            & izr   & Izere            \\
bds   & Burunge                 & dag   & Dagbani              & gbo   & Grebo, Northern                & izz   & Izii             \\
bem   & Bemba / Chibemba        & dav   & Dawida / Taita       & gbr   & Gbagyi                         & jgo   & Ngomba           \\
beq   & Beembe                  & dga   & Dagaare              & gde   & Gude                           & jib   & Jibu             \\
ber   & Berber                  & dgd   & Dagaari Dioula       & gid   & Gidar                          & jit   & Jita             \\
bex   & Jur Modo                & dgi   & Dagara, Northern     & giz   & South Giziga                   & jmc   & Machame          \\
bez   & Bena                    & dhm   & Dhimba               & gjn   & Gonja                          & kab   & Kabyle           \\
bfa   & Bari                    & dib   & Dinka, South Central & gkn   & Gokana                         & kam   & Kikamba          \\
bfd   & Bafut                   & did   & Didinga              & gkp   & Kpelle, Guinea                 & kbn   & Kare             \\
bfo   & Birifor, Malba          & dig   & Chidigo              & gmv   & Gamo                           & kbo   & Keliko           \\
bib   & Bisa                    & dik   & Dinka, Southwestern  & gna   & Kaansa                         & kbp   & Kabiye           \\
bim   & Bimoba                  & dip   & Dinka, Northeastern  & gnd   & Zulgo-gemzek                   & kby   & Kanuri, Manga    \\
bin   & Edo                     & diu   & Gciriku              & gng   & Ngangam                        & kcg   & Tyap             \\
biv   & Birifor, Southern       & dks   & Dinka, Southeastern  & gof   & Goofa                          & kck   & Kalanga          \\
bjv   & Bedjond                 & dnj   & Dan                  & gog   & Gogo                           & kdc   & Kutu   \\
\bottomrule
\end{tabular}%
}

\caption{Languages covered in \ourmodel~- Part I. }
\label{tab:lang_listI}
\end{table*}

\begin{table*}[]
\small
\centering
\resizebox{\textwidth}{!}{%
\begin{tabular}{ll ll ll ll }
\toprule
\textbf{ISO-639-3} & \textbf{Language}                & \textbf{ISO-639-3} & \textbf{Language}             & \textbf{ISO-639-3} & \textbf{Language}                       & \textbf{ISO-639-3} & \textbf{Language}         \\
\midrule
kde & Makonde               & laj & Lango                       & mfh & Matal                           & ngb & Ngbandi, Northern                     \\
kdh & Tem                   & lam & Lamba                       & mfi & Wandala                         & ngc & Ngombe                                \\
kdi & Kumam                 & lap & Laka                        & mfk & Mofu, North                     & ngl & Lomwe                                 \\
kdj & Ng’akarimojong        & lee & Lyélé                       & mfq & Moba                            & ngn & Bassa                                 \\
kdl & Tsikimba              & lef & Lelemi                      & mfz & Mabaan                          & ngo & Ngoni                                 \\
kdn & Kunda                 & lem & Nomaande                    & mgc & Morokodo                        & ngp & Ngulu                                 \\
kea & Kabuverdianu          & lgg & Lugbara                     & mgh & Makhuwa-Meetto                  & nhr & Naro                                  \\
ken & Kenyang               & lgm & Lega-mwenga                 & mgo & Meta'                           & nhu & Noone                                 \\
khy & Kele / Lokele         & lia & Limba, West-Central         & mgq & Malila                          & nih & Nyiha                                 \\
kia & Kim                   & lik & Lika                        & mgr & Mambwe-Lungu                    & nim & Nilamba / kinilyamba                  \\
kik & Gikuyu / Kikuyu       & lin & Lingala                     & mgw & Matumbi                         & nin & Ninzo                                 \\
kin & Kinyarwanda           & lip & Sekpele                     & mif & Mofu-Gudur                      & niy & Ngiti                                 \\
kiz & Kisi                  & lmd & Lumun                       & mkl & Mokole                          & nka & Nkoya / ShiNkoya                      \\
kki & Kagulu                & lmp & Limbum                      & mlg & Malagasy                        & nko & Nkonya                                \\
kkj & Kako                  & lnl & Banda, South Central        & mlr & Vame                            & nla & Ngombale                              \\
kln & Kalenjin              & log & Logo                        & mmy & Migaama                         & nnb & Nande / Ndandi                        \\
klu & Klao                  & lom & Loma                        & mnf & Mundani                         & nnh & Ngiemboon                             \\
kma & Konni                 & loq & Lobala                      & mnk & Mandinka                        & nnq & Ngindo                                \\
kmb & Kimbundu              & lot & Latuka                      & moa & Mwan                            & nse & Chinsenga                             \\
kmy & Koma                  & loz & Silozi                      & mos & Moore                           & nnw & {\color[HTML]{717171} Nuni, Southern} \\
knf & Mankanya              & lro & Laro                        & moy & Shekkacho                       & nso & Sepedi                                \\
kng & Kongo                 & lsm & Saamya-Gwe / Saamia         & moz & Mukulu                          & ntr & Delo                                  \\
knk & Kuranko               & lth & Thur / Acholi-Labwor        & mpe & Majang                          & nuj & Nyole                                 \\
kno & Kono                  & lto & Tsotso                      & mpg & Marba                           & nus & Nuer                                  \\
koo & Konzo                 & lua & Tshiluba                    & mqb & Mbuko                           & nwb & Nyabwa                                \\
koq & Kota                  & luc & Aringa                      & msc & Maninka, Sankaran               & nxd & Ngando                                \\
kqn & Kikaonde              & lue & Luvale                      & mur & Murle                           & nya & Chichewa                              \\
kqp & Kimré                 & lug & Luganda                     & muy & Muyang                          & nyb & Nyangbo                               \\
kqs & Kisi                  & lun & Lunda                       & mwe & Mwera                           & nyd & Olunyole / Nyore                      \\
kqy & Koorete               & luo & Dholuo / Luo                & mwm & Sar                             & nyf & Giryama                               \\
kri & Krio                  & lwg & Wanga                       & mwn & Cinamwanga                      & nyk & Nyaneka                               \\
krs & Gbaya                 & lwo & Luwo                        & mws & Mwimbi-Muthambi                 & nym & Nyamwezi                              \\
krw & Krahn, Western        & maf & Mafa                        & myb & Mbay                            & nyn & Nyankore / Nyankole                   \\
krx & Karon                 & mas & Maasai                      & myk & Sénoufo, Mamara                 & nyo & Nyoro                                 \\
ksb & Shambala / Kishambala & maw & Mampruli                    & myx & Masaaba                         & nyu & Nyungwe                               \\
ksf & Bafia                 & mbu & Mbula-Bwazza                & mzm & Mumuye                          & nyy & Nyakyusa-Ngonde / Kyangonde           \\
ksp & Kabba                 & mck & Mbunda                      & mzw & Deg                             & nza & Mbembe, Tigon                         \\
ktj & Krumen, Plapo         & mcn & Masana / Massana            & naq & {\color[HTML]{717171} Khoekhoe} & nzi & Nzema                                 \\
ktu & Kikongo               & mcp & Makaa                       & naw & Nawuri                          & odu & Odual                                 \\
kua & Oshiwambo             & mcu & Mambila, Cameroon           & nba & Nyemba                          & ogo & Khana                                 \\
kub & Kutep                 & mda & Mada                        & nbl & IsiNdebele                      & oke & Okpe                                  \\
kuj & Kuria                 & mdm & Mayogo                      & ncu & Chunburung                      & okr & Kirike                                \\
kus & Kusaal                & mdy & Maale                       & ndc & Ndau                            & oku & Oku                                   \\
kvj & Psikye                & men & Mende                       & nde & IsiNdebele                      & orm & Oromo                                 \\
kwn & Kwangali              & meq & Merey                       & ndh & Ndali                           & ozm & Koonzime                              \\
kyf & Kouya                 & mer & Kimiiru                     & ndj & Ndamba                          & pcm & Nigerian Pidgin                       \\
kyq & Kenga                 & mev & Maan / Mann                 & ndo & Ndonga                          & pem & Kipende                               \\
kzr & Karang                & mfe & Morisyen / Mauritian Creole & ndv & Ndut                            & pkb & Kipfokomo / Pokomo                    \\
lai & Lambya                & mfg & Mogofin                     & ndz & Ndogo    \\
\bottomrule
\end{tabular}%
}
\caption{Languages covered in \ourmodel~- Part II }
\label{tab:lang_listII}
\end{table*}

\begin{table*}[]
\small
\centering
\resizebox{0.8\textwidth}{!}{%
\begin{tabular}{ll ll ll}
\toprule
\textbf{ISO-639-3} & \textbf{Language}                & \textbf{ISO-639-3} & \textbf{Language}             & \textbf{ISO-639-3} & \textbf{Language}                                \\
\midrule
pov & Guinea-Bissau Creole      & tcd & Tafi               & won                      & Wongo               \\
poy & Pogolo / Shipogoro-Pogolo & ted & Krumen, Tepo       & xan                      & Xamtanga            \\
rag & Lulogooli                 & tem & Timne              & xed                      & Hdi                 \\
rel & Rendille                  & teo & Teso               & xho                      & Isixhosa            \\
rif & Tarifit                   & tex & Tennet             & xnz                      & Mattokki            \\
rim & Nyaturu                   & tgw & Senoufo, Tagwana   & xog                      & Soga                \\
rnd & Uruund                    & thk & Tharaka            & xon                      & Konkomba            \\
rng & Ronga / ShiRonga          & thv & Tamahaq, Tahaggart & xpe                      & Kpelle              \\
rub & Gungu                     & tir & Tigrinya           & xrb                      & Karaboro, Eastern   \\
run & Rundi / Kirundi           & tiv & Tiv                & xsm                      & Kasem               \\
rwk & Rwa                       & tke & Takwane            & xtc                      & Katcha-Kadugli-Miri \\
sag & Sango                     & tlj & Talinga-Bwisi      & xuo                      & Kuo                 \\
saq & Samburu                   & tll & Otetela            & yal                      & Yalunka             \\
sba & Ngambay                   & tog & Tonga              & yam                      & Yamba               \\
sbd & Samo, Southern            & toh & Gitonga            & yao                      & Yao / Chiyao        \\
sbp & Sangu                     & toi & Chitonga           & yat                      & Yambeta             \\
sbs & Kuhane                    & tpm & Tampulma           & yba                      & Yala                \\
sby & Soli                      & tsc & Tshwa              & ybb                      & Yemba               \\
sef & Sénoufo, Cebaara          & tsn & Setswana           & yom                      & Ibinda              \\
ses & Songhay, Koyraboro Senni  & tso & Tsonga             & yor                      & Yoruba              \\
sev & Sénoufo, Nyarafolo        & tsw & Tsishingini        & yre                      & Yaoure              \\
sfw & Sehwi                     & ttj & Toro / Rutoro      & zaj                      & Zaramo              \\
sgw & Sebat Bet Gurage          & ttq & Tawallammat        & zdj                      & Comorian, Ngazidja  \\
shi & Tachelhit                 & ttr & Nyimatli           & zga                      & Kinga               \\
shj & Shatt                     & tui & Toupouri           & ziw                      & Zigula              \\
shk & Shilluk                   & tul & Kutule             & zne                      & Zande / paZande     \\
sid & Sidama                    & tum & Chitumbuka         & zul                      & Isizulu             \\
sig & Paasaal                   & tuv & Turkana            &  &                     \\
sil & Sisaala, Tumulung         & tvu & Tunen              &  &                     \\
sna & Shona                     & twi & Twi                &  &                     \\
snf & Noon                      & umb & Umbundu            &  &                     \\
sng & Sanga / Kiluba            & urh & Urhobo             &  &                     \\
snw & Selee                     & uth & ut-Hun            &  &                     \\
som & Somali                    & vag & Vagla              &  &                     \\
sop & Kisonge                   & vai & Vai                &  &                     \\
sor & Somrai                    & ven & Tshivenda          &  &                     \\
sot & Sesotho                   & vid & Chividunda         &  &                     \\
soy & Miyobe                    & vif & Vili               &  &                     \\
spp & Senoufo, Supyire          & vmk & Makhuwa-Shirima    &  &                     \\
ssw & Siswati                   & vmw & Macua              &  &                     \\
suk & Sukuma                    & vun & Kivunjo            &  &                     \\
sus & Sosoxui                   & vut & Vute               &  &                     \\
swa & Swahili                   & wal & Wolaytta           &  &                     \\
swc & Swahili Congo             & wbi & Vwanji             &  &                     \\
swh & Swahili                   & wec & Guere              &  &                     \\
swk & Sena, Malawi              & wes & Pidgin, Cameroon   &  &                     \\
sxb & Suba                      & wib & Toussian, Southern &  &                     \\
taq & Tamasheq                  & wmw & Mwani              &  &                     \\
tcc & Datooga                   & wol & Wolof              &  & 
\\
\bottomrule
\end{tabular}%
}

\caption{Languages covered in \ourmodel~- Part III.}
\label{tab:lang_listIII}
\end{table*}
\begin{table}[h]
\centering
\begin{tabular}{ll}\toprule
\multicolumn{1}{c}{\textbf{ISO-639-3}} & \multicolumn{1}{c}{\textbf{\#Tokens}} \\\toprule
swh                                & 2,912,488,735                         \\
afr                                & 1,264,478,436                         \\
som                                & 587,549,878                           \\
swa                                & 499,792,448                           \\
hau                                & 286,806,539                           \\
amh                                & 241,700,000                           \\
mlg                                & 137,852,716                           \\
zne                                & 89,981,183                            \\
sna                                & 75,413,519                            \\
...                                & ...                                   \\
bam                                & 3,262                                 \\
har                                & 3,066                                 \\
dyo                                & 1,797                                 \\
fvr                                & 1,680                                 \\
tbz                                & 1,578                                 \\
ddn                                & 1,372                                 \\
fuc                                & 1,336                                 \\
knc                                & 1,097                                 \\
eot                                & 1,041                                 \\
cgg                                & 845                                  \\ \bottomrule
\end{tabular}
\caption{The sizes of the top $10$ and bottom $10$ languages in \ourmodel~pretraining. }
\label{tab:freq}
\end{table}



\end{document}